\documentclass[plain]{IEEEtran}
\usepackage[english]{babel}
\usepackage{amsmath,verbatim,paralist,url}
\usepackage{color}
\usepackage{cite}
\usepackage{mathdots}
\usepackage{amsfonts}
\usepackage{graphicx}
\usepackage{multirow}
\usepackage[colorinlistoftodos]{todonotes}
\usepackage{mathrsfs}
\usepackage{calc}
\usepackage{empheq}
\usepackage{centernot}
\usepackage{algorithmic}
\usepackage{bbding,graphicx}
\usepgflibrary{patterns}
\usetikzlibrary{patterns}
\usepackage{algorithm}
\usepackage{leftidx}
\usepackage{booktabs}
\usepackage{blindtext}
\usepackage[abs]{overpic}
\usepackage{xcolor,varwidth}
\usepackage{threeparttable}
\usepackage{framed}
\usepackage{qtree}
\usepackage{fancybox}
\usepackage{subcaption}
\usepackage{amssymb}
\usepackage{extarrows}
\usepackage{bbm}
\usetikzlibrary{chains,arrows}
\newcommand{\argmax}{\operatornamewithlimits{argmax}}

\newcommand{\rsone}{\textcolor{black}{\text{equal}}}
\newcommand{\rstwo}{\text{patrol}}

\newcommand{\rsthree}{\textcolor{black}{\text{asym}}}
\newcommand{\unique}{\operatorname{u}}
\usetikzlibrary{calc}

\usepackage{amsthm}
\usepackage{array}

\DeclareMathOperator{\atan2}{atan2}

\newtheorem{thm}{Theorem}

\begin{document}

\title{Syntactic Enhancement to VSIMM for Roadmap Based Anomalous Trajectory Detection: A Natural Language  Processing Approach}

\author{{Vikram~Krishnamurthy,
~\IEEEmembership{Fellow,~IEEE} and Sijia~Gao}
\thanks{Vikram Krishnamurthy and S. Gao are with the School of Electrical Engineering and Cornell Tech, Cornell University, N.Y., USA. Vikram~Krishnamurthy (vikramk@cornell.edu) is the corresponding author. Both authors contributed equally to the technical aspects of the paper.
  This research was funded by an Airforce Office of Scientific Research grant  FA9550-18-1-0007 through the Dynamic Data Driven Application Systems Program.
  
This paper has supplementary downloadable material available at {http://ieeexplore.ieee.org}, provided by the authors. The material includes the Matlab code that generates all the simulation results.  The code is also available from https://vikram.ece.cornell.edu/
}}

\maketitle

\begin{abstract}
\textcolor{black}{Syntactic tracking aims} to classify a target's spatio-temporal trajectory \textcolor{black}{by} using natural language processing models. This paper \textcolor{black}{proposes constrained stochastic context free grammar (CSCFG) models for target trajectories} confined to a roadmap. We present a particle filtering algorithm that \textcolor{black}{exploits the CSCFG model structure to estimate} the target's trajectory. This meta-level algorithm operates in conjunction with a base-level target tracking algorithm. Extensive numerical results using simulated ground moving target indicator (GMTI) radar measurements show \textcolor{black}{useful} improvement in both trajectory classification and target \textcolor{black}{state (both coordinates and velocity)} estimation.
\end{abstract}

\begin{IEEEkeywords}
syntactic tracking, constrained stochastic context free grammar (CSCFG), \textcolor{black}{particle filter}, \textcolor{black}{Earley Stolcke parser}
\end{IEEEkeywords}

\IEEEpeerreviewmaketitle

\section{Introduction}
\label{sec:intro}
\begin{figure}
\centering
\begin{tikzpicture}[scale=1,circle dotted/.style={dash pattern=on .04mm off 1mm,
                              line cap=round}]
\tikzstyle{vertex}=[rectangle,fill=gray!50,minimum size=0.9cm]

\begin{scope}
\node at (0,0)[vertex]{ };
\node at (1,0)[vertex]{ };
\node at (2,0)[vertex]{ };
\draw[pattern=crosshatch] (2.6,-0.4) rectangle (3.4,0.35);
\node at (0,1)[vertex]{ };
\node at (1,1)[vertex]{ };
\node at (2,1)[vertex]{ };
\node at (3,1)[vertex]{ };
\node at (0,2)[vertex]{ };
\draw[pattern=crosshatch] (0.6,1.6) rectangle (1.4,2.35);
\node at (2,2)[vertex]{ };
\node at (3,2)[vertex]{ };
\node at (0,3)[vertex]{ };
\node at (1,3)[vertex]{ };
\node at (2,3)[vertex]{ };
\node at (3,3)[vertex]{ };

\draw [-,dashed](0.5,-0.4) to (0.5,0.4);
\draw [-,dashed](1.5,-0.4) to (1.5,0.4);
\draw [-,dashed](2.5,-0.4) to (2.5,0.4);

\draw [-,dashed](0.5,0.6) to (0.5,1.4);
\draw [-,dashed](1.5,0.6) to (1.5,1.4);
\draw [-,dashed](2.5,0.6) to (2.5,1.4);

\draw [-,dashed](0.5,1.6) to (0.5,2.4);
\draw [-,dashed](1.5,1.6) to (1.5,2.4);
\draw [-,dashed](2.5,1.6) to (2.5,2.4);

\draw [-,dashed](0.5,2.6) to (0.5,3.4);
\draw [-,dashed](1.5,2.6) to (1.5,3.4);
\draw [-,dashed](2.5,2.6) to (2.5,3.4);

\draw [-,dashed](-0.5,2.6) to (-0.5,3.4);
\draw [-,dashed](-0.5,1.6) to (-0.5,2.4);
\draw [-,dashed](-0.5,0.6) to (-0.5,1.4);
\draw [-,dashed](-0.5,-0.4) to (-0.5,0.4);

\draw [-,dashed](3.5,2.6) to (3.5,3.4);
\draw [-,dashed](3.5,1.6) to (3.5,2.4);
\draw [-,dashed](3.5,0.6) to (3.5,1.4);
\draw [-,dashed](3.5,-0.4) to (3.5,0.4);

\draw [-,dashed](-0.4,0.5) to (0.4,0.5);
\draw [-,dashed](0.6,0.5) to (1.4,0.5);
\draw [-,dashed](1.6,0.5) to (2.4,0.5);
\draw [-,dashed](2.6,0.5) to (3.4,0.5);

\draw [-,dashed](-0.4,1.5) to (0.4,1.5);
\draw [-,dashed](0.6,1.5) to (1.4,1.5);
\draw [-,dashed](1.6,1.5) to (2.4,1.5);
\draw [-,dashed](2.6,1.5) to (3.4,1.5);

\draw [-,dashed](-0.4,2.5) to (0.4,2.5);
\draw [-,dashed](0.6,2.5) to (1.4,2.5);
\draw [-,dashed](1.6,2.5) to (2.4,2.5);
\draw [-,dashed](2.6,2.5) to (3.4,2.5);

\draw [-,dashed](-0.4,3.5) to (0.4,3.5);
\draw [-,dashed](0.6,3.5) to (1.4,3.5);
\draw [-,dashed](1.6,3.5) to (2.4,3.5);
\draw [-,dashed](2.6,3.5) to (3.4,3.5);

\draw [-,dashed](-0.4,-0.5) to (0.4,-0.5);
\draw [-,dashed](0.6,-0.5) to (1.4,-0.5);
\draw [-,dashed](1.6,-0.5) to (2.4,-0.5);
\draw [-,dashed](2.6,-0.5) to (3.4,-0.5);

\node [blue] at (-0.5,3.5){$v_{21}$};
\node [blue] at (0.5,3.5){$v_{22}$};
\node [blue] at (1.5,3.5){$v_{23}$};
\node [blue] at (2.5,3.5){$v_{24}$};
\node [blue] at (3.5,3.5){$v_{25}$};

\node [blue] at (-0.5,2.5){$v_{16}$};
\node [blue] at (0.5,2.5){$v_{17}$};
\node [blue] at (1.5,2.5){$v_{18}$};
\node [blue] at (2.5,2.5){$v_{19}$};
\node [blue] at (3.5,2.5){$v_{20}$};

\node [blue] at (-0.5,1.5){$v_{11}$};
\node [blue] at (0.5,1.5){$v_{12}$};
\node [blue] at (1.5,1.5){$v_{13}$};
\node [blue] at (2.5,1.5){$v_{14}$};
\node [blue] at (3.5,1.5){$v_{15}$};

\node [blue] at (-0.5,0.5){$v_{6}$};
\node [blue] at (0.5,0.5){$v_{7}$};
\node [blue] at (1.5,0.5){$v_{8}$};
\node [blue] at (2.5,0.5){$v_{9}$};
\node [blue] at (3.5,0.5){$v_{10}$};

\node [blue] at (-0.5,-0.5){$v_{1}$};
\node [blue] at (0.5,-0.5){$v_{2}$};
\node [blue] at (1.5,-0.5){$v_{3}$};
\node [blue] at (2.5,-0.5){$v_{4}$};
\node [blue] at (3.5,-0.5){$v_{5}$};
\end{scope}

\begin{scope}
\draw[red,line width = 0.5mm] (0.55,1.55) -- (0.55,2.45);
\draw[red,line width = 0.5mm] (0.55,2.45) -- (1.45,2.45);
\draw[red,line width = 0.5mm,] (1.45,2.45) -- (1.45,1.55);
\draw[red,line width = 0.5mm] (1.45,1.55) -- (0.45,1.55);
\draw[red,line width = 0.5mm] (0.45,1.55) -- (0.45,0.45);
\draw[red,line width = 0.5mm] (0.45,0.45) -- (1.45,0.45);

\draw[red,line width = 0.5mm] (1.45,0.45) -- (3.45,0.45);
\draw[red,line width = 0.5mm] (3.45,0.45) -- (3.45,-0.45);
\draw[red,line width = 0.5mm] (3.45,-0.45) -- (2.55,-0.45);
\draw[red,line width = 0.5mm] (2.55,-0.45) -- (2.55,0.45);

\draw[green,line width = 0.5mm] (2.45,0.4) -- (3.4,0.4);
\draw[green,line width = 0.5mm] (3.4,0.4) -- (3.4,-0.4);
\draw[green,line width = 0.5mm] (3.4,-0.4) -- (2.6,-0.4);
\draw[green,line width = 0.5mm] (2.6,-0.4) -- (2.6,1.55);
\draw[green,line width = 0.5mm] (2.6,1.55) -- (1.45,1.55);
\draw[green,line width = 0.5mm] (1.45,1.6) -- (0.6,1.6);
\draw[green,line width = 0.5mm] (0.6,1.6) -- (0.6,2.4);
\draw[green,line width = 0.5mm] (0.6,2.4) -- (1.4,2.4);
\draw[green,line width = 0.5mm] (1.4,2.4) -- (1.4,1.4);
\end{scope}
\begin{scope}[xshift=1.9in]
\draw[->](0,0) to (0.3,0);
\node at (0.65,0){East};
\draw[->](0,0) to (-0.3,0);
\node at (-0.65,0){West};
\draw[->](0,0) to (0,0.3);
\node at (0,0.48){North};
\draw[->](0,0) to (0,-0.3);
\node at (0,-0.45){South};
\end{scope}
\end{tikzpicture}
\caption{Example of a roadmap. $V=\{v_1, v_2,\ldots,v_{25}\}$ denotes the set of road intersections. $E=\{e_{ij}| v_i,v_j\in V\}$ denotes the set of directed roads where $e_{ij}$ denotes the road  from $v_i$ to $v_j$. Grey squares denote blocks. The red and green lines are examples of roadmap based anomalous trajectories discussed in this paper. We construct generative models for these trajectories using constrained stochastic context free grammars (CSCFG).}
\label{fig:roadmap}
\end{figure}
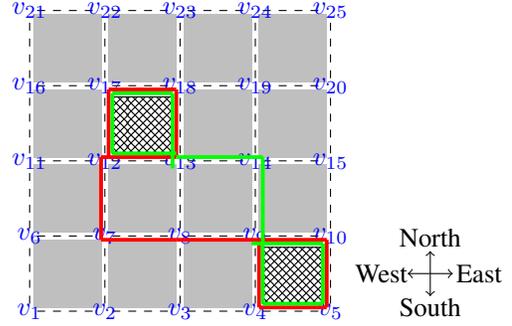

\textcolor{black}{Consider a moving} target confined to the road network illustrated in Fig.\,{\ref{fig:roadmap}}. Assume that the target is being tracked by a ground moving target indicator (GMTI) radar system. At each discrete time~$k$, let $\mathbf{x}_{k}$ denote target's kinematic  state vector comprising position and velocity of the target as it moves in a two dimensional space. Let $\mathbf{z}_k$ denote the noisy measurement of $\mathbf{x}_k$ obtained from a GMTI radar. \textcolor{black}{How can the constraints of a digital roadmap be exploited for tracking a target confined to a road network?}
Classical (base-level) \textcolor{black}{target} tracking algorithms  have been well studied in the literature\cite{baselevel2,baselevel3,baselevel4,VSIMM1,VSIMM2}. These include the variable structure interacting multiple model (VSIMM) tracker\cite{VSIMM1}\cite{VSIMM2}. Let $d_{k}$ and $l_k$ denote, respectively, the direction of motion and location (road or intersection names) of the target. In a VSIMM tracker, the direction sequence $d_{1:k}=(d_1,\ldots,d_k)$ is modeled as a Markov chain with state space dependent on the location sequence $l_{1:k}=(l_1,\ldots,l_k)$. These direction and location sequences are chosen so that the target is confined to roads and intersections in a roadmap. Also {\cite{jump_Markov} uses a jump Markov process to model vehicular mobility on an urban roadmap.}

At a higher level of abstraction (lower degree of spatial resolution and slower time scale), a moving target confined to a roadmap can be characterized by the string of symbols denoting the roads it traverses. For example, \textcolor{black}{the trajectory (red and green dotted lines) illustrated in  Fig.\,{\ref{fig:roadmap}} describes a target performing anomalous (suspicious) patrol activities \textcolor{black}{on two crosshatched blocks}: circling \textcolor{black}{the crosshatched blocks} (red line) followed by circling the same blocks in the reverse order (green line).} Given noisy radar observations, how can an automated system\footnote{Traditionally, given track information, a radar (human) operator examines target trajectories to determine anomalous behavior. This paper develops natural language models
and meta-level signal processing algorithms for estimating anomalous trajectories. Such ``middleware" forms the interface between  the physical signal processing layer and the radar operator; see also~\cite{ahmad2016bayesian} for alternative models for intent using bridging distributions.} detect such an anomalous trajectory      {\cite{anomalous_1,anomalous_2,anomalous_3,edit_distance,masterthesis}}? The main idea in this paper is to model such anomalous trajectories as symbols from a natural language generated by a constrained stochastic context free grammar (\textcolor{black}{CSCFG}). Put simply, the target trajectory speaks a language about its intent.

This paper provides a \textcolor{black}{method} \textcolor{black}{for maximum likelihood classification of anomalous trajectories} via a ``syntactic'' enhancement to the baseline VSIMM tracker. The syntactic enhancement operates at a higher (meta) level and models anomalous trajectories. \textcolor{black}{Our aim is to compute the maximum likelihood estimate of the target's trajectory:}
\begin{equation}
\textcolor{black}{\argmax_{G\in\mathbf{G}}p(G|\mathbf{z}_{1:k})}, \quad k=1,2,\ldots \label{eq:aim}
\end{equation}
Here, $\mathbf{z}_{1:k}=(\mathbf{z_1},\mathbf{z}_2,\ldots,\mathbf{z}_k)$ \textcolor{black}{denotes} the noisy observation sequence recorded by a GMTI radar. $G$ denotes a \textcolor{black}{class of trajectories (such as all strings that result in trajectories are rectangles) which is a finite set of strings}. $\mathbf{G}$ denotes \textcolor{black}{the} set of different \textcolor{black}{classes of trajectories that a radar operator is interested in}. The posterior probability mass function $p(G|\mathbf{z}_{1:k})$ in (\ref{eq:aim}) takes into account the roadmap statistics (e.g., turning ratio from one road to another) obtained from traffic data.

\subsection{\textcolor{black}{CSCFG as a Generative Model}}
The syntactic tracker we propose in this paper consists of two parts: a meta-level tracker and a base-level tracker. The meta-level tracker uses tracklets generated by the base-level tracker to model higher level trajectories. The architecture of a syntactic tracker is illustrated in Fig.\,{\ref{fig:syntactic tracker}}. 

\textcolor{black}{Our key idea} is that for each \textcolor{black}{trajectory class} $G\in\mathbf{G}$ in \eqref{eq:aim}, we propose a generative model using a \emph{constrained stochastic context free grammar} (CSCFG)\textcolor{black}{\cite{constrained_grammar}\cite{constrained_grammar_parameter_estimation}} on a weighted, directed graph. By a generative model for trajectory class $G$, we mean the following necessary and sufficient condition:
\begin{equation}
\label{eq:generative model}
\begin{split}
&\text{1) Each string generated by the CSCFG with specified}\\
&\text{grammatical rules belongs to class } G.\\
&\text{2) Each string in class } G \text{ can be generated by the CSCFG}.
\end{split}
\end{equation}
Therefore, the class $G$ of anomalous trajectories is equivalent to the corresponding CSCFG grammatical rules. So to classify a trajectory, we only need to classify the grammar within the family of CSCFGs. This 
\textcolor{black}{is} done in Sec.\ref{sec:algorithm} via a particle filter algorithm that combines the functionalities of CSCFG and VSIMM.

\begin{figure}
\centering 
\includegraphics[width=0.35\textwidth]{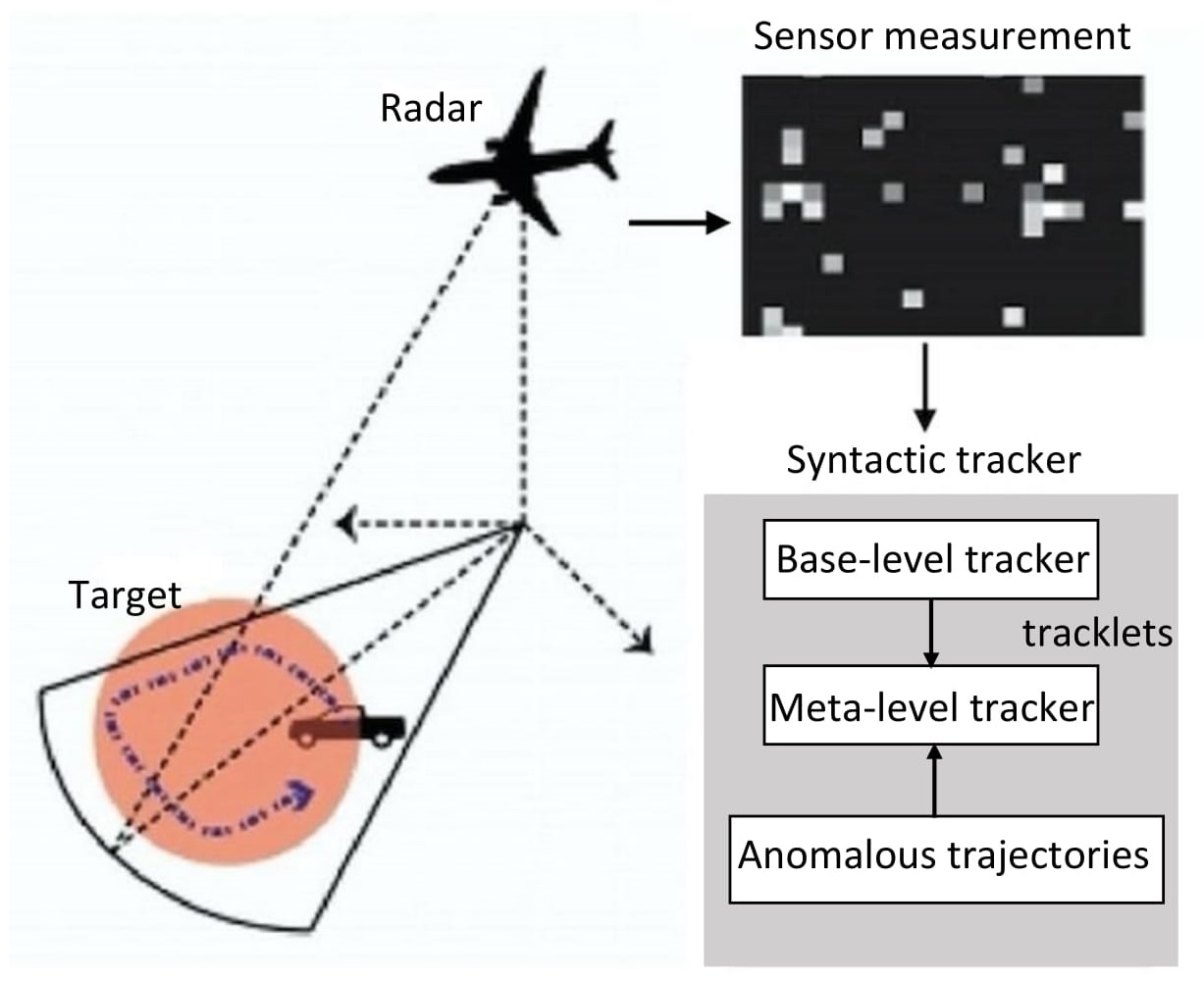}
 \caption{The architecture of a syntactic tracker. Anomalous trajectories are pre-defined by the radar operator and indicate the intent of a target.}
 \label{fig:syntactic tracker}
\end{figure}
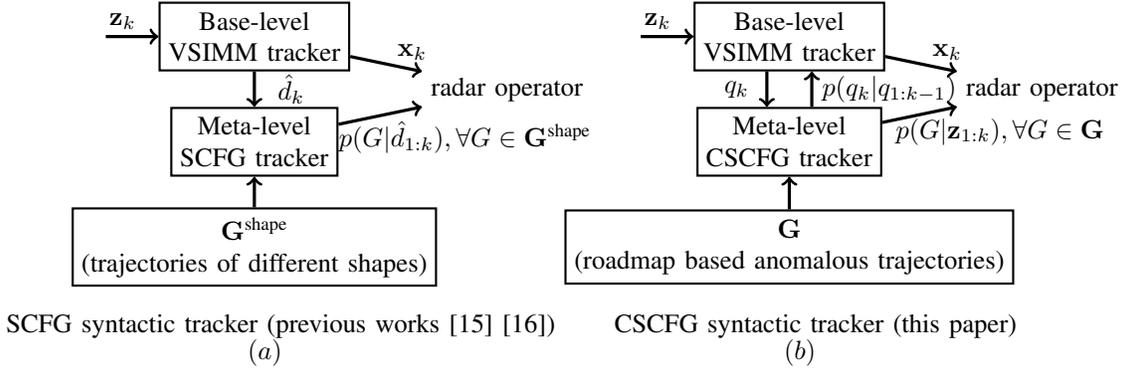
\begin{figure*} 
\centering
\begin{tikzpicture}[scale=0.7]
\begin{scope}
\node (input) at (-3,0){};
\node at (-2.5,0.3){$\mathbf{z}_{k}$};
\node[draw=black,rectangle,thick,align=center,fill=white] (a) at (0,0)  {Base-level\\VSIMM tracker};
\draw[very thick,->] (input) edge (a);
\node[draw=black,rectangle,thick,align=center,fill=white] (b) at (0,-2)  {Meta-level\\SCFG tracker};
\node at (3,-0.3){$\mathbf{x}_{k}$};
\node[draw=black,rectangle,thick,align=center,fill=white] (c) at (0,-4)  { $\mathbf{G}^{\text{shape}}$\\(trajectories of different shapes)};
\node at (0.7,-1) {$\hat d_k$};
\draw[very thick,->] (a) edge (b);
\draw[very thick,->] (c) edge (b);
\node at (4,-1.9){$p(G|\hat d_{1:k}),\forall G\in\mathbf{G}^{\text{shape}}$};
\node (operator) at (4.8,-1){\textcolor{black}{radar operator}};
\draw[very thick,->] (a) edge (operator);
\draw[very thick,->] (b) edge (operator);
\node at(0.5,-5.5){SCFG syntactic tracker (previous works \cite{SCFG_intent_inference}\cite{SCFG_syntactic_tracker})};
\node at(0.2,-6){$(a)$};
\end{scope} 

\begin{scope}[xshift=4in]
\node (input) at (-3,0){};
\node at (-2.5,0.3){$\mathbf{z}_{k}$};
\node[draw=black,rectangle,thick,align=center,fill=white] (a) at (0,0)  {Base-level\\VSIMM tracker};
\draw[very thick,->] (input) edge (a);
\node[draw=black,rectangle,thick,align=center,fill=white] (b) at (0,-2)  {Meta-level\\CSCFG tracker};
\node[draw=black,rectangle,thick,align=center,fill=white] (c) at (0,-4)  {$\mathbf{G}$\\(roadmap based anomalous trajectories)};
\node at (-1,-1) {$\textcolor{black}{q_k}$};
\node at (1.9,-1) {\textcolor{black}{$p(q_k|q_{1:k-1})$}};
\draw[very thick,->,transform canvas={xshift=0.85em}] (b) edge (a);
\draw[very thick,->,transform canvas={xshift=-0.85em}] (a) edge (b);
\draw[very thick,->] (c) edge (b);
\node (operator) at (4.8,-1){\textcolor{black}{radar operator}};
\draw[very thick,->] (a) edge (operator);
\draw[very thick,->] (b) edge (operator);
\node at (3,-0.3){$\mathbf{x}_{k}$};
\node at (4,-1.8){$p(G|\textcolor{black}{\mathbf{z}_{1:k}}),\forall G\in\mathbf{G}$};
\node at(0.5,-5.5){CSCFG syntactic tracker (this paper)};
\node at(0.2,-6){$(b)$};
\end{scope} 
\end{tikzpicture}
\caption{Architectures of $(a)$ the SCFG syntactic tracker and $(b)$ the CSCFG syntactic tracker \textcolor{black}{to assist radar operator in detecting anomalous trajectories.} $\mathbf{z}_{k}$ denotes the noisy observation recorded by a GMTI radar at time $k$ and $\mathbf{x}_k$ denotes the target's state vector. $\mathbf{G}^{\text{shape}}$ and $\mathbf{G}$ \textcolor{black}{denote two different classes of} trajectories. $\hat d_{1:k}=(\hat d_1,\hat d_2,\ldots,\hat d_{k})$ denotes the estimated (soft or hard) moving direction sequence. \textcolor{black}{$q_{1:k-1}$ defined in \eqref{eq:qk old} is \textcolor{black}{the} sequence of roads the target travels \textcolor{black}{until time $k$}}. The posterior probability \textcolor{black}{$p(G|\textcolor{black}{\mathbf{z}_{1:k}})$} of anomalous trajectories is fed to the radar operator.} 
\label{fig:architecture}
\end{figure*}

\subsection{\textcolor{black}{Organization and Main Results}}
\label{subsec:organization}
\textcolor{black}{The organization and }main results of this paper are:

1. Sec.\,{\ref{sec:architecture and CSCFG}} describes syntactic tracker architectures and gives more insight into \textcolor{black}{the natural language models used in this paper. Specifically, }the differences between CSCFGs, SCFGs and template matching \textcolor{black}{are discussed.}

2. Sec.\,{\ref{sec:3 level model}} describes the construction of a natural language driven model for the roadmap based syntactic tracking problem in \eqref{eq:aim}. The main idea is that we facilitate CSCFGs as generative models for target trajectories on a directed, weighted graph formulated from the roadmap. This CSCFG model operates in conjunction with the baseline VSIMM which has measurements from a GMTI radar\footnote{\textcolor{black}{The GMTI radar assumption is only to make the problem concrete. The methods proposed in this paper  apply to other types of radar/sensor models}.}.

3. Sec.\,{\ref{sec:level2}} details the construction of CSCFG generative models for the roadmap confined target trajectories. Several
important examples of anomalous trajectories are modeled.

4. \textcolor{black}{Sec.\,{\ref{sec:algorithm}} \textcolor{black}{presents} a particle \textcolor{black}{filtering algorithm} to classify the target trajectory given noisy radar observations. The novelty of \textcolor{black}{this algorithm} is that it exploits a modified Earley Stolcke parser \textcolor{black}{(arising in natural language parsing of CSCFGs) to compute the one step prediction} and likelihood of the target trajectory.}

5. Sec.\,{\ref{sec:simulation}} presents numerical studies \textcolor{black}{of the} CSCFG syntactic tracker. Compared \textcolor{black}{to} the baseline VSIMM tracker, the CSCFG syntactic \textcolor{black}{tracker enables} anomalous trajectory detection and has smaller state estimate error.

\section{\textcolor{black}{Syntactic Tracker Architectures and CSCFGs}}
\label{sec:architecture and CSCFG}
\subsection{Syntactic Tracker Architectures}
\label{subsec:architecture}
To give \textcolor{black}{context} into the main ideas of this paper, we \textcolor{black}{describe two} architectures for syntactic trackers, namely (i) CSCFG syntactic tracker proposed in this paper and (ii) stochastic context free grammar (SCFG) syntactic tracker proposed in earlier works \cite{SCFG_intent_inference}\cite{SCFG_syntactic_tracker}.

The SCFG syntactic tracker in previous works \cite{SCFG_metalevel_modeling,SCFG_metalevel_estimate,balajia2012consistency} was used to classify shapes (lines, arcs or m-rectangles) of trajectories. In such a tracker, the direction sequence $d_{1:k}$ is modeled via a stochastic context free grammar (SCFG) at the meta-level. The architecture of the SCFG syntactic tracker is illustrated in Fig.\,{\ref{fig:architecture}}(a); see also
 \cite{blasch2012high} for an excellent review.

The key new idea in this paper is to combine a Markov chain and a SCFG which results in a \emph{constrained stochastic context free grammar} (CSCFG). A CSCFG forms a generative model for the  distinct roads that the target traverses, namely 
\begin{equation}
\label{eq:qk old}
\unique(q_{1:k})\text{ where }q_{1:k}=(q_1,q_2,\ldots,q_k)
\end{equation}
\textcolor{black}{In \eqref{eq:qk old}, $q_k$ denotes the road the target moves on at time $k$. $\unique(.)$ denotes a function on a string that removes repetitions in the string, e.g., $\unique(``122343")=``12343"$.} The trajectory (red and green lines) in Fig.\,{\ref{fig:roadmap}} can be described by the following directed road sequence:
\begin{equation}
\label{eq:CSCFG example}
\begin{split}
&\underbrace{e_{12,17}e_{17,18}e_{18,13}e_{13,12}}_{\text{patrol block 10}}\underbrace{\ldots}_{\text{move from block 10 to 4}} \underbrace{e_{9,10}e_{10,5}e_{5,4}e_{4,9}}_{\text{patrol block 4}}\\
&\underbrace{e_{9,10}e_{10,5}e_{5,4}e_{4,9}}_{\text{patrol block 4}}\underbrace{\ldots}_{\text{move from block 4 to 10}} \underbrace{e_{13,12}e_{12,17}e_{17,18}e_{18,13}}_{\text{patrol block 10}}
\end{split}
\end{equation}
\textcolor{black}{The transition from one road to another is modeled as a Markov chain based} on the traffic statistics. \textcolor{black}{The blocks that the target traverses are modeled via a SCFG.} Hence, the road sequence in \eqref{eq:qk old} is a combination of a Markov chain and a SCFG which is equivalent to a CSCFG: see \cite{constrained_grammar}\cite{constrained_grammar_parameter_estimation}. The structure of the CSCFG syntactic tracker is illustrated in Fig.\,{\ref{fig:architecture}}(b). 

In contrast, it is important to note that the SCFG is not a generative model for the directed road sequence in \eqref{eq:CSCFG example}. Directed road sequences modeled via a SCFG include physically impossible trajectories of the target \textcolor{black}{such as, for example}
\begin{equation}
e_{6,11}\,e_{12,17}\,e_{18,19}\,e_{20,15}\,e_{14,9}\label{eq:unrealize}
\end{equation}
The sequence in \eqref{eq:unrealize} \textcolor{black}{describes} a target moving from $e_{6,11}$ to $e_{12,17}$ which is physically impossible because $e_{6,11}$ and $e_{12,17}$ are not connected via an intersection on the roadmap in Fig.\,{\ref{fig:roadmap}}. To summarize, a CSCFG is essential as a generative model for spatial trajectories that are physically realizable. \textcolor{black}{The syntactic tracker architecture can be viewed as ``middleware'' which interfaces the physical signal processing layer with the radar operator (human decision maker).}

\subsection{\textcolor{black}{Context.} CSCFG vs SCFG vs Markov Chain vs Template Matching\textcolor{black}{: A} Toy Example}
\label{subsec:compare MC, SCFG and CSCFG}
\textcolor{black}{To give further insight into the key ideas of this paper,  we give a toy example to illustrate the difference between trajectories generated by a Markov chain, SCFG and CSCFG. The  Chomsky hierarchy of natural languages \cite{pumping_lemma} says} 
\begin{equation}
\label{eq:chomsky}
\text{Markov Chain} \subset   \text{ SCFG } \subset \text{  CSCFG }   
\end{equation}
\textcolor{black}{implying that CSCFGs generate trajectories with much more complex dependency structures than a Markov chain can.}

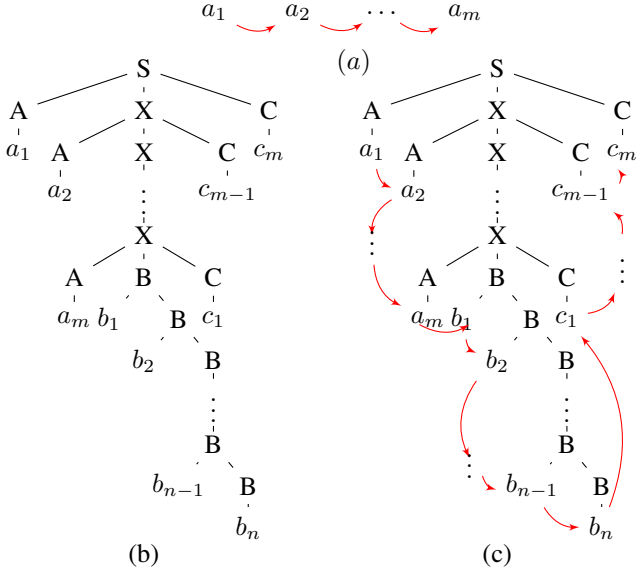
\begin{figure}
\centering
\begin{tikzpicture}[scale=0.37]
\tikzstyle{level 1}=[sibling distance=45mm] 
\tikzstyle{level 2}=[sibling distance=30mm]
\tikzstyle{level 3}=[sibling distance=25mm]
\tikzset{edge/.style = {red,->,> = latex'}}
\begin{scope}[xshift=-1in]
\node (a_1)at (0,0){$a_1$};
\node (a_2)at (3,0){$a_2$};
\node (a_medium) at (6,0){$\ldots$};
\node (a_m) at (9,0){$a_m$};

\draw [edge] (a_1) to [bend right] (a_2);
\draw [edge] (a_2) to [bend right] (a_medium);
\draw [edge] (a_medium) to [bend right] (a_m);
\node at (5,-1.8){$(a)$};
\end{scope}
\begin{scope}[xshift=-2in,yshift=-0.8in]
\node {S}
    child{node{A}
      child{node {$a_1$}}
    }
    child{node{X}
      child{node{A}
        child{node{$a_2$}}
      }
      child{node{X}
        child{node{$\vdots$}
          child{node{X}
            child{node {A}
              child{node {$a_m$}}
            }
            child{node{B}
               child{node{$b_1$}}
               child{node{B}
                 child{node{$b_2$}}
                 child{node{B}
                   child{node{$\vdots$}
                     child{node{B}
                       child{node{$b_{n-1}$}}
                       child{node{B}
                         child{node{$b_n$}}
                       }
                     }
                     }
                 }
                 }
               }
             child{node{C}
              child{node {$c_1$}}
            }
          }
        }
      }
      child{node{C}
        child{node {$c_{m-1}$}}
      }
    }
    child{node {C}
      child{node {$c_m$}}
    };
\node at (0,-17.5){(b)};
  
\end{scope}

\begin{scope}[xshift=3in,yshift=-0.8in]
\node {S}
    child{node{A}
      child{node (a_1){$a_1$}}
    }
    child{node{X}
      child{node{A}
        child{node (a_2) {$a_2$}}
      }
      child{node{X}
        child{node{$\vdots$}
          child{node{X}
            child{node {A}
              child{node (a_m){$a_m$}}
            }
            child{node{B}
               child{node (b_1){$b_1$}}
               child{node{B}
                 child{node (b_2) {$b_2$}}
                 child{node{B}
                   child{node{$\vdots$}
                     child{node{B}
                       child{node (b_{n-1}){$b_{n-1}$}}
                       child{node{B}
                         child{node (b_n){$b_n$}}
                       }
                     }
                     }
                 }
                 }
               }
            child{node{C}
              child{node (c_1){$c_1$}}
            }
          }
        }
      }
      child{node{C}
        child{node (c_{m-1}){$c_{m-1}$}}
      }
    }
    child{node {C}
      child{node (c_m){$c_m$}}
    };

\draw [edge] (a_1) to [bend right] (a_2);
\draw [edge] (a_2) to [bend right] (-4.5,-6);
\node (a_medium) at (-4.5,-6){$\vdots$};
\draw [edge] (a_medium) to [bend right] (a_m);
\draw [edge] (-2.8,-9.2) to [bend right] (-1,-9.2);
\draw [edge] (b_1) to [bend right] (b_2);
\draw [edge] (b_2) to [bend right] (-1,-14);
\node (b_medium) at (-1,-14){$\vdots$};
\draw [edge] (b_medium) to [bend right] (b_{n-1});
\draw [edge] (b_{n-1}) to [bend right] (b_n);
\draw [edge] (b_n) to [bend right] (c_1);
\node (c_medium) at (4.5,-7){$\vdots$};
\draw [edge] (c_1) to [bend right] (c_medium);
\draw [edge] (c_medium) to [bend right] (c_{m-1});
\draw [edge] (c_{m-1}) to [bend right] (c_m);
\node at (0,-17.5){(c)};
\end{scope}
\end{tikzpicture}
\caption{\textcolor{black}{Comparison of dependency structures of $(a)$ Markov chain, $(b)$ SCFG model and $(c)$ CSCFG model}. A CSCFG model is a combination of a Markov chain and a SCFG model. Black edges represent tree dependencies and red arrows indicate serial dependencies. $X$, $A$, $B$ and $C$ denote nonterminals (hidden states). Tree dependencies are described in Sec.\,{\ref{subsec:compare MC, SCFG and CSCFG}}.}
\label{fig:dependency}
\end{figure}

{\em Markov Chain (serial dependency)}. \textcolor{black}{Consider} a first-order Markov chain with trajectory $a_1,\ldots,a_m$ for some fixed time $m$. The dependency
structure \textcolor{black}{is a chain graph as shown in} Fig.\,{\ref{fig:dependency}}(a).

{\em SCFG Arc Trajectory ({tree dependency})}. An arc is a simple example of trajectory with a SCFG generative model.
\begin{compactenum}
\item  Generate positive random integers $m,n$.
\item Then generate the following three iid finite state  sequences $a_1,\ldots, a_m$, $b_1,\ldots,b_n$ and $ c_1,\ldots, c_m$ with specified probabilities. Concatenate these into a single string.
\end{compactenum}
\noindent The directed road sequence modeled in \eqref{eq:unrealize}, with $a_{1:m}=e_{6,11}e_{12,17}$, $b_{1:n}=e_{18,19}$, $c_{1:m}=e_{20,15}e_{14,9}$ is an example of a SCFG generated arc trajectory: \textcolor{black}{ the number of road segments directed northwards that the target traverses equals the number of road segments that the target traverses southwards.} The dependency
structure of a \textcolor{black}{SCFG \textcolor{black}{has a tree type graphical representation} as shown} in Fig.\,{\ref{fig:dependency}}(b). It can be proved via a pumping lemma \cite{pumping_lemma} that a  Markov chain is not a generative model \textcolor{black}{for an arc trajectory since the arbitrary integer $m$ needs to be remembered.}

{\em CSCFG ({serial-tree dependency})}. A CSCFG trajectory (used in this paper) allows more general dependencies than a SCFG as follows:
\begin{compactenum}
\item Generate positive random integers $m$, $n$.
\item Then generate the following three Markovian finite state  sequences $a_1,\ldots,a_m$, $b_1,\ldots,b_n$ and $c_1,\ldots, c_m$ with specified transition probabilities. Concatenate these into a single string.
\end{compactenum}
$e_{6,11}e_{11,16}e_{16,17}e_{17,12}e_{12,7}$ is an example of an arc trajectory (number of roads directed to north equals that directed to south) generated by the CSCFG with $a_{1:m}= e_{6,11}e_{11,16}$, $b_{1:n}=e_{16,17}$, $c_{1:m}=e_{17,12}e_{12,7}$. The Markovian property \textcolor{black}{constrains} the road directions from a vertex to ensure that physically unrealizable trajectories such as \eqref{eq:unrealize} do not occur. The dependency
structure of a CSCFG is a tree-chain graph as shown in Fig.\,{\ref{fig:dependency}}(c). The main point is that a CSCFG model facilitates both serial and tree dependencies. 

{\em \textcolor{black}{Template Matching vs CSCFG}}. 
Apart from being  generative models for several classes of  complex spatial trajectories, CSCFGs also offer the advantage of having computationally efficient Bayesian estimation algorithms compared to  classical template matching. For example, consider a target trajectory represented by the string $\textcolor{black}{M}=a_{1:m}b_{1:n}c_{1:p}$ of length $N=m+n+p$ where $m$, $n$, $p$ are unknown nonnegative integers. Suppose $a_{1:m}$, $b_{1:n}$, $c_{1:p}$ are finite state Markov sequences with state spaces $\{a^1,a^2\}$, $\{b^1,b^2\}$, $\{c^1,c^2\}$. How can we detect whether the string \textcolor{black}{$M$} contains same number of alphabets $a^1$ and $c^1$? A naive template matching approach requires an exponential number $\sum_{m=1}^{N}\sum_{p=1}^{N-m}\sum_{k=1}^{\min(m,p)}\dbinom{m}{k}\dbinom{p}{k}2^{N-m-p}$ of templates. \textcolor{black}{To consider arbitrary length dependencies}, a Markov chain needs an exponential number $\sum_{k=1}^{N}6^{k}$ of states. By comparison, a CSCFG only requires polynomial $\mathcal{O}(N^3)$ computational cost. Of course, in this paper, we consider the further modeling complexity that the string $M$ itself is observed in noise due to errors in the classical target tracking algorithm-\textcolor{black}{nevertheless the computational cost is $\mathcal{O}(N^3)$.}
 
\section{{Roadmap based Syntactic Tracking}: A 3-level Model}
\label{sec:3 level model}
In this section, we construct a model for the roadmap based syntactic tracking problem. Our model operates at three levels of abstraction. At the highest level \textcolor{black}{of abstraction}, we have the roadmap which is modeled as a directed, weighted graph. At the second level, we model the target's trajectory \textcolor{black}{constrained to a roadmap} as a \textcolor{black}{string generated from a} CSCFG. Finally at the lowest level (physical sensor layer), the directed road sequence \textcolor{black}{$\unique(q_{1:k})$} defined in \eqref{eq:qk old} drives a baseline VSIMM state space model which has measurements from a GMTI radar. 

\subsection{Level 1: Roadmap as a Directed, Weighted Graph}
\label{subsec:level1}
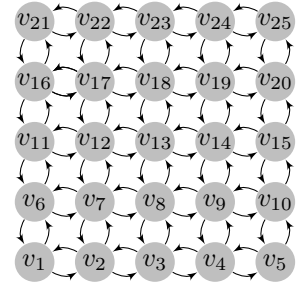
\begin{figure}
\centering
\begin{tikzpicture}[scale=0.2]
\begin{scope}[shift={(0,0)},rotate=0]
\tikzstyle{vertex}=[circle,fill=black!25,minimum size=15pt,inner sep=0pt]
\tikzset{edge/.style = {->,> = latex'}}
\node[vertex] (v21) at  (0,0) {$v_{21}$};
\node[vertex] (v22) at  (4,0) {$v_{22}$};
\node[vertex] (v23) at (8,0) {$v_{23}$};
\node[vertex] (v24) at (12,0) {$v_{24}$};
\node[vertex] (v25) at (16,0) {$v_{25}$};

\node[vertex] (v16) at  (0,-4) {$v_{16}$};
\node[vertex] (v17) at  (4,-4) {$v_{17}$};
\node[vertex] (v18) at (8,-4) {$v_{18}$};
\node[vertex] (v19) at (12,-4) {$v_{19}$};
\node[vertex] (v20) at (16,-4) {$v_{20}$};

\node[vertex] (v11) at  (0,-8) {$v_{11}$};
\node[vertex] (v12) at  (4,-8) {$v_{12}$};
\node[vertex] (v13) at (8,-8) {$v_{13}$};
\node[vertex] (v14) at (12,-8) {$v_{14}$};
\node[vertex] (v15) at (16,-8) {$v_{15}$};

\node[vertex] (v6) at  (0,-12) {$v_{6}$};
\node[vertex] (v7) at  (4,-12) {$v_{7}$};
\node[vertex] (v8) at (8,-12) {$v_{8}$};
\node[vertex] (v9) at (12,-12) {$v_{9}$};
\node[vertex] (v10) at (16,-12) {$v_{10}$};

\node[vertex] (v1) at  (0,-16) {$v_{1}$};
\node[vertex] (v2) at  (4,-16) {$v_{2}$};
\node[vertex] (v3) at (8,-16) {$v_{3}$};
\node[vertex] (v4) at (12,-16) {$v_{4}$};
\node[vertex] (v5) at (16,-16) {$v_{5}$};
\draw[edge] (v1) to[bend right] (v2);
\draw[edge] (v2) to[bend right] (v1);

\draw[edge] (v2) to[bend right] (v3);
\draw[edge] (v3) to[bend right] (v2);

\draw[edge] (v3) to[bend right] (v4);
\draw[edge] (v4) to[bend right] (v3);

\draw[edge] (v4) to[bend right] (v5);
\draw[edge] (v5) to[bend right] (v4);

\draw[edge] (v6) to[bend right] (v7);
\draw[edge] (v7) to[bend right] (v6);

\draw[edge] (v7) to [bend right] (v8);
\draw[edge] (v8) to [bend right] (v7);

\draw[edge] (v8) to [bend right] (v9);
\draw[edge] (v9) to [bend right] (v8);

\draw[edge] (v9) to [bend right] (v10);
\draw[edge] (v10) to [bend right] (v9);

\draw[edge] (v11) to [bend right] (v12);
\draw[edge] (v12) to [bend right] (v11);

\draw[edge] (v12) to [bend right] (v13);
\draw[edge] (v13) to [bend right] (v12);

\draw[edge] (v13) to [bend right] (v14);
\draw[edge] (v14) to [bend right] (v13);

\draw[edge] (v14) to [bend right] (v15);
\draw[edge] (v15) to [bend right] (v14);

\draw[edge] (v2) to [bend right] (v7);
\draw[edge] (v7) to [bend right] (v2);

\draw[edge] (v3) to [bend right] (v8);
\draw[edge] (v8) to [bend right] (v3);

\draw[edge] (v4) to [bend right] (v9);
\draw[edge] (v9) to [bend right] (v4);

\draw[edge] (v7) to [bend right] (v12);
\draw[edge] (v12) to [bend right] (v7);

\draw[edge] (v8) to [bend right] (v13);
\draw[edge] (v13) to [bend right] (v8);

\draw[edge] (v9) to [bend right] (v14);
\draw[edge] (v14) to [bend right] (v9);

\draw[edge] (v16) to [bend right] (v17);
\draw[edge] (v17) to [bend right] (v16);

\draw[edge] (v17) to [bend right] (v18);
\draw[edge] (v18) to [bend right] (v17);

\draw[edge] (v18) to [bend right] (v19);
\draw[edge] (v19) to [bend right] (v18);

\draw[edge] (v19) to [bend right] (v20);
\draw[edge] (v20) to [bend right] (v19);

\draw[edge] (v21) to [bend right] (v22);
\draw[edge] (v22) to [bend right] (v21);

\draw[edge] (v22) to [bend right] (v23);
\draw[edge] (v23) to [bend right] (v22);

\draw[edge] (v23) to [bend right] (v24);
\draw[edge] (v24) to [bend right] (v23);

\draw[edge] (v24) to [bend right] (v25);
\draw[edge] (v25) to [bend right] (v24);

\draw[edge] (v1) to [bend right](v6);
\draw[edge] (v6) to [bend right] (v1);
\draw[edge] (v5) to [bend right] (v10);
\draw[edge] (v10) to [bend right] (v5);
\draw[edge] (v6) to [bend right](v11);
\draw[edge] (v11) to [bend right] (v6);
\draw[edge] (v16) to [bend right] (v17);
\draw[edge] (v17) to [bend right] (v16);
\draw[edge] (v10) to [bend right] (v15);
\draw[edge] (v15) to [bend right] (v10);
\draw[edge] (v11) to [bend right](v16);
\draw[edge] (v16) to [bend right] (v11);
\draw[edge] (v12) to [bend right] (v17);
\draw[edge] (v17) to [bend right] (v12);
\draw[edge] (v13) to [bend right] (v18);
\draw[edge] (v18) to [bend right] (v13);
\draw[edge] (v14) to [bend right] (v19);
\draw[edge] (v19) to [bend right] (v14);
\draw[edge] (v15) to [bend right] (v20);
\draw[edge] (v20) to [bend right] (v15);
\draw[edge] (v16) to [bend right](v21);
\draw[edge] (v21) to [bend right] (v16);
\draw[edge] (v17) to [bend right] (v22);
\draw[edge] (v22) to [bend right] (v17);
\draw[edge] (v18) to [bend right] (v23);
\draw[edge] (v23) to [bend right] (v18);
\draw[edge] (v19) to [bend right] (v24);
\draw[edge] (v24) to [bend right] (v19);
\draw[edge] (v20) to [bend right] (v25);
\draw[edge] (v25) to [bend right] (v20);
\end{scope}
\end{tikzpicture}
\caption{Formulation of the roadmap in Fig.\,{\ref{fig:roadmap}} as a directed, weighted graph $\mathcal{G}=\{V,E,\theta\}$. \textcolor{black}{The vertices }$v_1,\ldots,v_{25}$ denote road intersections and edges denote directed roads. The weight of \textcolor{black}{each} edge denotes the direction (north, south, east or west) of \textcolor{black}{the} edge.}
\label{fig:roadmap graph}
\end{figure}

\textcolor{black}{Here}, we model the roadmap as a directed, weighted graph $\mathcal{G}$ with vertices $V$, directed edges $E$ and weights $\theta$
\begin{equation}
\mathcal{G}=\{V,E,\theta\}\label{eq:graph}
\end{equation}
The set of vertices $V=\{v_1,v_2,\ldots,v_n\}$
denotes intersections. The set of directed edges $E= \{e_{ij}| v_i,v_j\in V\}$ denotes directed roads. The set of weights $\theta=\{\theta(e_{ij})\in\{\text{north, south, east, west}\}, \forall e_{ij}\in E\}$ denotes directions of roads with respect to a reference coordinate. The directed, weighted graph for the roadmap in Fig.\,{\ref{fig:roadmap}} is presented in Fig.\,{\ref{fig:roadmap graph}}.
Denote the vertex (road intersection)
\begin{equation}
v_i=\operatorname{from}(e_{ij})\text{  and }v_j=\operatorname{into}(e_{ij})
\label{eq:from and into}
\end{equation}
Note that the road network directed graph model described above does not require the road map to be a rectangular (Manhattan) grid.

\subsection{Level 2: Trajectories and CSCFG Modeling on Directed, Weighted Graph}
\label{subsec:level2}
The second level of our 3-level model is a CSCFG that \textcolor{black}{serves as a generative model for the} target's trajectories. These trajectories determine the target's directed road sequence:
\begin{equation}
\label{eq:qk new}
\begin{split}
&\unique(q_{1:N}),\quad \text{where } q_{1:N}=(q_1,q_2,\ldots,q_N)\text{ with}\\
&q_k\in E, k=1,2,\ldots,N \text{ and}\\
&q_{k-1}=q_k\text{ or }\operatorname{into}(q_{k-1})=\operatorname{from}(q_{k}),k=2,3,\ldots,\textcolor{black}{N}
\end{split}
\end{equation}
\textcolor{black}{Here, $\unique(.)$, $E$ are defined in \eqref{eq:qk old} and \eqref{eq:graph}. $\operatorname{into}(.)$ and $\operatorname{from}(.)$ are defined in \eqref{eq:from and into}}.
Recall $q_k$ denotes the road the target moves on at time $k$.

A CSCFG is a 5-tuple of the form\footnote{For the reader's convenience, in Appendix~\ref{appendix:background}, we give a short description of a SCFG and additional examples. Mathematically speaking, CSCFGs belong to the class of multi-type Galton Watson branching random processes.} $\text{CSCFG}=\{\mathcal{N},\mathcal{T},S,\mathcal{R},\mathcal{P}\}$ where $\mathcal{N}$ is a finite set of nonterminals ({hidden states}) and $\mathcal{T}$ is a finite set of terminals ({observations}) such that $\mathcal{N}\cap\mathcal{T}=\emptyset$. $S\in\mathcal{N}$ is chosen to be the start symbol. $\mathcal{R}$ is a set of production rules of the form
\begin{equation}
\nonumber
\begin{split}
&X\rightarrow \lambda|a\text{ where }X\in\mathcal{N}\cup S, a\in\mathcal{T}\\
&\lambda:\textcolor{black}{\text{a string of nonterminals and terminals}}
\end{split}
\end{equation}
\textcolor{black}{which indicates the nonterminal $X$ can be replaced with $\lambda$ if the previous terminal is $a$}. $\mathcal{P}: \mathcal{R}\rightarrow [0,1]$ is a probability function over production rules in $\mathcal{R}$ such~that $\sum_{n_{Xa}}P(X\rightarrow \lambda|a)=1$. $n_{Xa}$ denotes the number of production rules in $\mathcal{R}$ associated with the nonterminal $X$ and terminal $a$. Starting from $S$, \textcolor{black}{repetitively} replace the leftmost nonterminal (such deviations can be represented as a parse tree \cite{earley_stolcke_parser}) according to the production rules in $\mathcal{R}$ and probabilities in $\mathcal{P}$ until the resulting string only comprises terminals.

A stochastic  parsing algorithm in natural language processing computes the one-step conditional  probabilities
\begin{equation}
p(a_{k+1}|a_{1:k},\text{CSCFG}),k=1,2,\ldots,N\nonumber
\end{equation}
and the prefix conditional   probabilities
\begin{equation}
p(a_{1:k}|\text{CSCFG}),k=1,2,\ldots,N\nonumber
\end{equation}
Here, $a_{1:k}=(a_1,a_2,\ldots,a_k)$ denotes a string of terminals.
  The parsing algorithm for CSCFGs is a generalization of the well known  forward filtering algorithm for HMMs. It operates over a parse tree via a top-down and then bottom-up manner. Appendix A gives a short description of SCFGs and CSCFGs.

\subsection{Level 3: VSIMM Base-Level Model}
\label{subsec:level3}
Here, we describe the third and final component of our 3-level roadmap constrained target model. The model is almost identical to the classical VSIMM except that the variable \textcolor{black}{$q_{k}$} below couples the model with the target's trajectory (modeled as a CSCFG in Level 2). We construct a VSIMM for the baseline target's kinematics\footnote{Our setup assumes a single
target with no missing measurements or  data association errors. Actually, missing measurements are easily handled at both the syntactic and base-level trackers. Data association is handled by the baseline tracking algorithm and not the meta-level tracker.}
which are measured by a GMTI radar system.

The target's kinematic state evolves as
\begin{equation}
\label{eq:state equ}
\begin{split}
\mathbf{x}_{k}&=\mathcal{F}(\mathbf{x}_{k-1},d_{k})+\tilde{\mathbf{w}}_{k}(d_{k})\\
d_{k}&=\theta(q_k),\,\mathcal{B}(\mathbf{x}_k)\in\{q_k,\operatorname{from}(q_k)\}
\end{split}
\end{equation}
Here, $\mathbf{x}_{k} = [x_{k},y_{k},\dot x_{k},\dot y_{k}]'$ is the 4-dimensional state vector of the target at time~$k$ that comprises position and velocity components in the $x$ and $y$ directions. $d_k$ denotes the target's moving direction. $q_{k}$, $\theta(.)$, $\operatorname{from}(.)$ are defined in \eqref{eq:qk new}, \eqref{eq:graph} and \textcolor{black}{\eqref{eq:from and into}}. $\mathcal{B}(.)$ denotes the target's meta-level location (road or intersection name at time $k$), that is $\mathcal{B}(.)$ maps the target's state vector to an edge (road) or a vertex (road intersection) in the graph $\mathcal{G}$ defined in \eqref{eq:graph}. $\mathcal{F}$ in \eqref{eq:state equ} is a nonlinear function and models the target's state process:
\begin{equation}
\label{eq:mathcal F}
\mathcal{F}(\mathbf{x}_{k-1},d_{k})=\\
\begin{bmatrix}
x_{k-1}+T\dot{x}_{k-1}\\
y_{k-1}+T\dot{y}_{k-1}\\
\sqrt{\dot{x}_{k-1}^2+\dot{y}_{k-1}^2}\cos(d_{k})\\
\sqrt{\dot{x}_{k-1}^2+\dot{y}_{k-1}^2}\sin(d_{k})\\
\end{bmatrix}
\end{equation}
Here, $T$ is the \textcolor{black}{sampling} interval between GMTI measurements. The state noise $\tilde{\mathbf{w}}_{k}(d_k)$ in \eqref{eq:state equ} is a zero-mean \textcolor{black}{iid} Gaussian process with covariance matrix $\tilde{Q}_{k}(d_{k})$ computed as
\begin{equation}
\label{eq:Q}
\begin{split}
&\tilde{Q}_{k}(d_{k})=AQ_{k}(d_{k})A'\\
A=\begin{bmatrix} 
\frac{T^2}{2}&0\\
0&\frac{T^2}{2}\\
T&0\\
0&T
\end{bmatrix},\quad
&Q_{k}(d_{k})=\rho_{d_k}
\begin{bmatrix}
\sigma_{o}^2&0\\0&\sigma_{a}^2\end{bmatrix} \rho_{d_{k}}' \quad\text{with}\\ 
\rho_{d_{k}}&=
\begin{bmatrix} 
\sin(d_{k})&\cos(d_{k})\\
-\cos(d_{k})& \sin(d_{k})
\end{bmatrix}
\end{split}
\end{equation}
Here, $'$ denotes transpose\textcolor{black}{,} $\sigma_a^2$ is the variance
along the direction of motion indicated by $d_k$ and $\sigma_o^2$ is the variance along the direction of motion orthogonal to~$d_k$.

The observation equation \textcolor{black}{specifies} the GMTI radar \textcolor{black}{measurements}:
\begin{equation}
\label{eq:obs equ}
\begin{split}
\mathbf{z}_{k}&=\mathcal{H}(\mathbf{x}_{k},\mathbf{c}_{k})+\mathbf{v}_{k}\quad\text{where}\\
\mathcal{H}(\mathbf{x}_{k},\mathbf{c}_{k})&=
\begin{bmatrix}
\sqrt{(x_{k}-x_{k}^c)^2+(y_{k}-y_{k}^c)^2+(z_{k}^c)^2}\\\\
\frac{(\dot x_{k}-\dot x_{k}^c)(x_{k}-x_{k}^c)+(\dot y_{k}-\dot y_{k}^c)(y_{k}-y_{k}^c)}{\sqrt{(x_{k}-x_{k}^c)^2+(y_{k}-y_{k}^c)^2+(z_{k}^c)^2}}\\\\
\frac{180}{\pi}\atan2 (y_{k}-y_{k}^c,x_{k}-x_{k}^c)
\end{bmatrix}
\end{split}
\end{equation}
Here, $\mathbf{z}_{k}=[r_{k},\dot r_{k}, a_{k}]^\prime $ denotes the 3-dimensional noisy observation vector recorded by a GMTI radar at time~$k$. $r_{k}$, $\dot r_{k}$, $a_{k}$ denote, respectively, the range, range rate and azimuth (in degrees, $(-180^\circ,180^\circ ] $). $\mathbf{c}_{k}=[x_{k}^c,y_{k}^c,\dot x_{k}^c,\dot y_{k}^c]'$ is the 4-dimensional state vector for the phase center of the GMTI radar's antenna on the aircraft it is mounted on. The vector \textcolor{black}{$\mathbf{c}_k$} comprises position and velocity components in the $x$ and $y$ directions. We assume that $z_{k}^c $ is the (constant) altitude of the aircraft and the (constant) altitude of the target is zero. $\atan2$ denotes the four-quadrant inverse tangent (in radians). 

\textcolor{black}{The observation noise} $\mathbf{v}_{k}$ in \eqref{eq:obs equ} is assumed to be a zero-mean white Gaussian process with covariance matrix
\begin{equation}
R=\begin{bmatrix}
\sigma_{r_{k}}^2&0&0\\0&\sigma_{\dot r_k}^2&0\\0&0&\sigma_{\textcolor{black}{\theta_k}}^2\label{eq:R}
\end{bmatrix}
\end{equation}
where $\sigma_{r_{k}}$, $\sigma_{\dot r_k}$ and $\sigma_{\textcolor{black}{\theta_k}}$ are standard deviations for range, range rate and azimuth, respectively. Note that $R$ is a diagonal matrix reflecting the assumption that the errors in the range, range rate and azimuth are uncorrelated.

To summarize, the VSIMM model above is a CSCFG modulated   stochastic state space model; where the CSCFG is a generative model for the complex spatial trajectory of the target.
  In comparison, classical VSIMM deals with a Markov modulated model.

\section{Trajectories and CSCFG modeling}
\label{sec:level2}
Given the 3-level model \textcolor{black}{of} Sec.\,{\ref{sec:3 level model}}, we now elaborate on Level 2 \textcolor{black}{of the model, namely spatial }trajectories and CSCFG generative models described in Sec.\,{\ref{subsec:level2}}. In particular, we discuss three  important anomalous trajectories: (i) {Equal effort search} (ii) {Asymmetric search} and (iii) Patrol. These trajectory classes exhibit \textcolor{black}{a tree-chain dependency} (recall Sec.\,{\ref{subsec:compare MC, SCFG and CSCFG}}) and CSCFG forms generative models for these classes. That is, the necessary and sufficient condition \eqref{eq:generative model} holds. Markov chains and SCFGs are not generative models for these examples. 

\subsection{\textcolor{black}{Example 1: Equal Effort Search Trajectory Class}}
\label{subsec:rsone}
{How can an intent  inference algorithm determine if a target has spent equal effort patrolling two distinct sites? The aim is to characterize targets whose intent is to spend  equal amount of resources patrolling  two distinct sites. As shown below, the resulting equal effort search trajectory  class is an  CSCFG.}

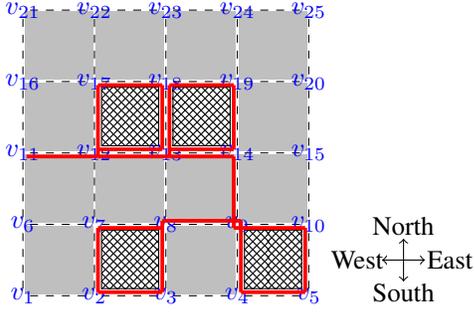
\begin{figure}
\centering
\begin{tikzpicture}[scale=0.95,circle dotted/.style={dash pattern=on .04mm off 1mm,
                              line cap=round}]
\tikzstyle{vertex}=[rectangle,fill=gray!50,minimum size=0.9cm]
\begin{scope}
\node at (0,0)[vertex]{ };
\draw[pattern=crosshatch] (0.6,-0.4) rectangle (1.4,0.42);
\node at (2,0)[vertex]{ };
\draw[pattern=crosshatch] (2.6,-0.4) rectangle (3.4,0.42);
\node at (0,1)[vertex]{ };
\node at (1,1)[vertex]{ };
\node at (2,1)[vertex]{ };
\node at (3,1)[vertex]{ };
\node at (0,2)[vertex]{ };
\draw[pattern=crosshatch] (0.6,1.6) rectangle (1.4,2.42);
\draw[pattern=crosshatch] (1.6,1.6) rectangle (2.4,2.42);
\node at (3,2)[vertex]{ };
\node at (0,3)[vertex]{ };
\node at (1,3)[vertex]{ };
\node at (2,3)[vertex]{ };
\node at (3,3)[vertex]{ };

\draw [-,dashed](0.5,-0.4) to (0.5,0.4);
\draw [-,dashed](1.5,-0.4) to (1.5,0.4);
\draw [-,dashed](2.5,-0.4) to (2.5,0.4);

\draw [-,dashed](0.5,0.6) to (0.5,1.4);
\draw [-,dashed](1.5,0.6) to (1.5,1.4);
\draw [-,dashed](2.5,0.6) to (2.5,1.4);

\draw [-,dashed](0.5,1.6) to (0.5,2.4);
\draw [-,dashed](1.5,1.6) to (1.5,2.4);
\draw [-,dashed](2.5,1.6) to (2.5,2.4);

\draw [-,dashed](0.5,2.6) to (0.5,3.4);
\draw [-,dashed](1.5,2.6) to (1.5,3.4);
\draw [-,dashed](2.5,2.6) to (2.5,3.4);

\draw [-,dashed](-0.5,2.6) to (-0.5,3.4);
\draw [-,dashed](-0.5,1.6) to (-0.5,2.4);
\draw [-,dashed](-0.5,0.6) to (-0.5,1.4);
\draw [-,dashed](-0.5,-0.4) to (-0.5,0.4);

\draw [-,dashed](3.5,2.6) to (3.5,3.4);
\draw [-,dashed](3.5,1.6) to (3.5,2.4);
\draw [-,dashed](3.5,0.6) to (3.5,1.4);
\draw [-,dashed](3.5,-0.4) to (3.5,0.4);

\draw [-,dashed](-0.4,0.5) to (0.4,0.5);
\draw [-,dashed](0.6,0.5) to (1.4,0.5);
\draw [-,dashed](1.6,0.5) to (2.4,0.5);
\draw [-,dashed](2.6,0.5) to (3.4,0.5);

\draw [-,dashed](-0.4,1.5) to (0.4,1.5);
\draw [-,dashed](0.6,1.5) to (1.4,1.5);
\draw [-,dashed](1.6,1.5) to (2.4,1.5);
\draw [-,dashed](2.6,1.5) to (3.4,1.5);

\draw [-,dashed](-0.4,2.5) to (0.4,2.5);
\draw [-,dashed](0.6,2.5) to (1.4,2.5);
\draw [-,dashed](1.6,2.5) to (2.4,2.5);
\draw [-,dashed](2.6,2.5) to (3.4,2.5);

\draw [-,dashed](-0.4,3.5) to (0.4,3.5);
\draw [-,dashed](0.6,3.5) to (1.4,3.5);
\draw [-,dashed](1.6,3.5) to (2.4,3.5);
\draw [-,dashed](2.6,3.5) to (3.4,3.5);

\draw [-,dashed](-0.4,-0.5) to (0.4,-0.5);
\draw [-,dashed](0.6,-0.5) to (1.4,-0.5);
\draw [-,dashed](1.6,-0.5) to (2.4,-0.5);
\draw [-,dashed](2.6,-0.5) to (3.4,-0.5);

\node [blue] at (-0.5,3.5){$v_{21}$};
\node [blue] at (0.5,3.5){$v_{22}$};
\node [blue] at (1.5,3.5){$v_{23}$};
\node [blue] at (2.5,3.5){$v_{24}$};
\node [blue] at (3.5,3.5){$v_{25}$};

\node [blue] at (-0.5,2.5){$v_{16}$};
\node [blue] at (0.5,2.5){$v_{17}$};
\node [blue] at (1.5,2.5){$v_{18}$};
\node [blue] at (2.5,2.5){$v_{19}$};
\node [blue] at (3.5,2.5){$v_{20}$};

\node [blue] at (-0.5,1.5){$v_{11}$};
\node [blue] at (0.5,1.5){$v_{12}$};
\node [blue] at (1.5,1.5){$v_{13}$};
\node [blue] at (2.5,1.5){$v_{14}$};
\node [blue] at (3.5,1.5){$v_{15}$};

\node [blue] at (-0.5,0.5){$v_{6}$};
\node [blue] at (0.5,0.5){$v_{7}$};
\node [blue] at (1.5,0.5){$v_{8}$};
\node [blue] at (2.5,0.5){$v_{9}$};
\node [blue] at (3.5,0.5){$v_{10}$};

\node [blue] at (-0.5,-0.5){$v_{1}$};
\node [blue] at (0.5,-0.5){$v_{2}$};
\node [blue] at (1.5,-0.5){$v_{3}$};
\node [blue] at (2.5,-0.5){$v_{4}$};
\node [blue] at (3.5,-0.5){$v_{5}$};
\end{scope}

\begin{scope}
\draw[red,line width = 0.5mm] (-0.45,1.45) -- (0.55,1.45);
\draw[red,line width = 0.5mm] (0.55,1.45) -- (0.55,2.45);
\draw[red,line width = 0.5mm] (0.55,2.45) -- (1.45,2.45);
\draw[red,line width = 0.5mm] (1.45,2.45) -- (1.45,1.55);
\draw[red,line width = 0.5mm] (1.45,1.55) -- (0.55,1.55);
\draw[red,line width = 0.5mm] (0.55,1.45) -- (1.55,1.45);
\draw[red,line width = 0.5mm] (1.55,1.45) -- (1.55,2.45);
\draw[red,line width = 0.5mm] (1.55,2.45) -- (2.45,2.45);
\draw[red,line width = 0.5mm] (2.45,2.45) -- (2.45,1.55);
\draw[red,line width = 0.5mm] (2.45,1.55) -- (1.55,1.55);
\draw[red,line width = 0.5mm] (1.55,1.45) -- (2.45,1.45);
\draw[red,line width = 0.5mm] (2.45,1.45) -- (2.45,0.45);
\draw[red,line width = 0.5mm] (2.45,0.45) -- (3.45,0.45);
\draw[red,line width = 0.5mm] (3.45,0.45) -- (3.45,-0.45);
\draw[red,line width = 0.5mm] (3.45,-0.45) -- (2.55,-0.45);
\draw[red,line width = 0.5mm] (2.55,-0.45) -- (2.55,0.55);
\draw[red,line width = 0.5mm] (2.55,0.55) -- (1.45,0.55);
\draw[red,line width = 0.5mm] (1.45,0.55) -- (1.45,-0.45);
\draw[red,line width = 0.5mm] (1.45,-0.45) -- (0.55,-0.45);
\draw[red,line width = 0.5mm] (0.55,-0.45) -- (0.55,0.45);
\draw[red,line width = 0.5mm] (0.55,0.45) -- (1.45,0.45);
\end{scope}
\begin{scope}[xshift=1.9in]
\draw[->](0,0) to (0.3,0);
\node at (0.65,0){East};
\draw[->](0,0) to (-0.3,0);
\node at (-0.65,0){West};
\draw[->](0,0) to (0,0.3);
\node at (0,0.48){North};
\draw[->](0,0) to (0,-0.3);
\node at (0,-0.45){South};
\end{scope}
\end{tikzpicture}
\caption{Example of an \textcolor{black}{equal effort search} trajectory class. The target selects two block rows and searches equal number of blocks located at each of them.} 
\label{fig:rsone}
\end{figure}

In the \textcolor{black}{equal effort search} trajectory class $G^{\rsone}$, a \textcolor{black}{target} (searcher) spends equal effort \textcolor{black}{searching two rows of blocks (blocks are defined in Fig.\,{\ref{fig:roadmap}})} The trajectory for searching each block is characterized by a string of 4 roads that surround the block. For example, searching the northwest crosshatched block in Fig.\,{\ref{fig:rsone}} is described by any string from the set $\{e_{12,17}\,e_{17,18}\,e_{18,13}\,e_{13,12},\,e_{17,18}\,e_{18,13}\,e_{13,12}\,e_{12,17},\\
e_{18,13}\,e_{13,12}\,e_{12,17}\,e_{17,18},\,e_{13,12}\,e_{12,17}\,e_{17,18}\,e_{18,13}\}$. More generally, $G^{\rsone}$ is defined by the class of strings (spatial trajectories)
\begin{equation}
\label{eq:rsone}
\begin{split}
&G^{\rsone}=a_{1}a_{2}\ldots a_{N}b_{1}b_2\ldots b_m c_{1}c_{2}\ldots c_{N}\\
\begin{tikzpicture}[overlay,remember picture,out=315,in=225,distance=0.4cm]
\draw[->,red,shorten >=3pt,shorten <=3pt] (1.3,0.5) to (1.8,0.5);
\draw[->,red,shorten >=3pt,shorten <=3pt] (1.7,0.5) to (2.2,0.5);
\draw[->,red,shorten >=3pt,shorten <=3pt] (2.4,0.5) to (2.9,0.5);
\draw[->,red,shorten >=3pt,shorten <=3pt] (2.7,0.5) to (3.2,0.5);
\draw[->,red,shorten >=3pt,shorten <=3pt] (3.0,0.5) to (3.5,0.5);
\draw[->,red,shorten >=3pt,shorten <=3pt] (3.3,0.5) to (3.8,0.5);
\draw[->,red,shorten >=3pt,shorten <=3pt] (3.9,0.5) to (4.4,0.5);
\draw[->,red,shorten >=3pt,shorten <=3pt] (4.3,0.5) to (4.8,0.5);
\draw[->,red,shorten >=3pt,shorten <=3pt] (4.7,0.5) to (5.2,0.5);
\draw[->,red,shorten >=3pt,shorten <=3pt] (5.4,0.5) to (5.9,0.5);
\end{tikzpicture}
&a\,b: \operatorname{into}(a)=\operatorname{from}(b)\\
\begin{tikzpicture}[overlay,remember picture,out=315,in=225,distance=0.4cm]
\draw[->,red,shorten >=3pt,shorten <=3pt] (0,0.5) to (0.5,0.5);
\end{tikzpicture}
&a_k=\text{search block } \textcolor{black}{(i_k,j), i_1<i_2<\ldots<i_N}\\
&c_k=\text{search block } \textcolor{black}{(m_k,n), m_1>m_2>\ldots>m_N}\\
&j\neq n\\
\end{split}
\end{equation}
Here, block ($i,j$) represents the block located at $i$th column, $j$th row, e.g.\,, block (2,3) indicates the northwest crosshatched block in Fig.\,{\ref{fig:rsone}}. $b_1 b_2 \ldots b_m$ denotes a string of edges directing from block \textcolor{black}{$(i_N,j)$ to block $(m_1,n)$}. $\operatorname{into}(.)$ and $\operatorname{from}(.)$ are defined in \eqref{eq:from and into}. 

\textcolor{black}{The key two points to note in \eqref{eq:rsone}} are: (i) the length of the strings $a_1a_2\ldots a_N$ and \textcolor{black}{$c_1c_2\ldots c_N$} are equal (tree dependency) (ii) current edge directs from the vertex that its previous edge directs into (red \textcolor{black}{arrows}, chain dependency). Therefore, the class of spatial trajectories $G^{\rsone}$ exhibits \textcolor{black}{a tree-chain dependency}. The CSCFG production rules that yield a generative model for $G^{\text{\rsone}}$ are given in Appendix~\ref{appendix:rule and consistency}.

\subsection{\textcolor{black}{Example 2: Asymmetric Search Trajectory Class}}
\label{subsec:rsthree}
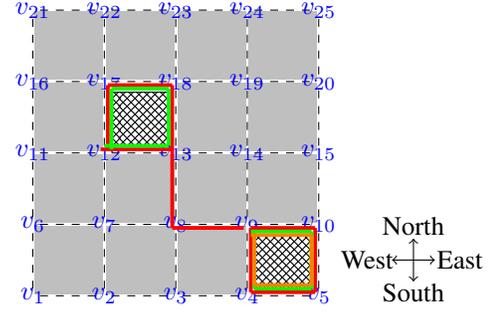
\begin{figure}
\centering
\begin{tikzpicture}[scale=0.95,circle dotted/.style={dash pattern=on .04mm off 1mm,line cap=round}]
\tikzstyle{vertex}=[rectangle,fill=gray!50,minimum size=0.9cm]
\begin{scope}
\node at (0,0)[vertex]{ };
\node at (1,0)[vertex]{ };
\node at (2,0)[vertex]{ };
\draw[pattern=crosshatch] (2.6,-0.4) rectangle (3.4,0.35);
\node at (0,1)[vertex]{ };
\node at (1,1)[vertex]{ };
\node at (2,1)[vertex]{ };
\node at (3,1)[vertex]{ };
\node at (0,2)[vertex]{ };
\draw[pattern=crosshatch] (0.6,1.6) rectangle (1.4,2.35);
\node at (2,2)[vertex]{ };
\node at (3,2)[vertex]{ };
\node at (0,3)[vertex]{ };
\node at (1,3)[vertex]{ };
\node at (2,3)[vertex]{ };
\node at (3,3)[vertex]{ };
\draw [-,dashed](0.5,-0.4) to (0.5,0.4);
\draw [-,dashed](1.5,-0.4) to (1.5,0.4);
\draw [-,dashed](2.5,-0.4) to (2.5,0.4);
\draw [-,dashed](0.5,0.6) to (0.5,1.4);
\draw [-,dashed](1.5,0.6) to (1.5,1.4);
\draw [-,dashed](2.5,0.6) to (2.5,1.4);
\draw [-,dashed](0.5,1.6) to (0.5,2.4);
\draw [-,dashed](1.5,1.6) to (1.5,2.4);
\draw [-,dashed](2.5,1.6) to (2.5,2.4);
\draw [-,dashed](0.5,2.6) to (0.5,3.4);
\draw [-,dashed](1.5,2.6) to (1.5,3.4);
\draw [-,dashed](2.5,2.6) to (2.5,3.4);
\draw [-,dashed](-0.5,2.6) to (-0.5,3.4);
\draw [-,dashed](-0.5,1.6) to (-0.5,2.4);
\draw [-,dashed](-0.5,0.6) to (-0.5,1.4);
\draw [-,dashed](-0.5,-0.4) to (-0.5,0.4);
\draw [-,dashed](3.5,2.6) to (3.5,3.4);
\draw [-,dashed](3.5,1.6) to (3.5,2.4);
\draw [-,dashed](3.5,0.6) to (3.5,1.4);
\draw [-,dashed](3.5,-0.4) to (3.5,0.4);
\draw [-,dashed](-0.4,0.5) to (0.4,0.5);
\draw [-,dashed](0.6,0.5) to (1.4,0.5);
\draw [-,dashed](1.6,0.5) to (2.4,0.5);
\draw [-,dashed](2.6,0.5) to (3.4,0.5);
\draw [-,dashed](-0.4,1.5) to (0.4,1.5);
\draw [-,dashed](0.6,1.5) to (1.4,1.5);
\draw [-,dashed](1.6,1.5) to (2.4,1.5);
\draw [-,dashed](2.6,1.5) to (3.4,1.5);
\draw [-,dashed](-0.4,2.5) to (0.4,2.5);
\draw [-,dashed](0.6,2.5) to (1.4,2.5);
\draw [-,dashed](1.6,2.5) to (2.4,2.5);
\draw [-,dashed](2.6,2.5) to (3.4,2.5);
\draw [-,dashed](-0.4,3.5) to (0.4,3.5);
\draw [-,dashed](0.6,3.5) to (1.4,3.5);
\draw [-,dashed](1.6,3.5) to (2.4,3.5);
\draw [-,dashed](2.6,3.5) to (3.4,3.5);
\draw [-,dashed](-0.4,-0.5) to (0.4,-0.5);
\draw [-,dashed](0.6,-0.5) to (1.4,-0.5);
\draw [-,dashed](1.6,-0.5) to (2.4,-0.5);
\draw [-,dashed](2.6,-0.5) to (3.4,-0.5);
\node [blue] at (-0.5,3.5){$v_{21}$};
\node [blue] at (0.5,3.5){$v_{22}$};
\node [blue] at (1.5,3.5){$v_{23}$};
\node [blue] at (2.5,3.5){$v_{24}$};
\node [blue] at (3.5,3.5){$v_{25}$};
\node [blue] at (-0.5,2.5){$v_{16}$};
\node [blue] at (0.5,2.5){$v_{17}$};
\node [blue] at (1.5,2.5){$v_{18}$};
\node [blue] at (2.5,2.5){$v_{19}$};
\node [blue] at (3.5,2.5){$v_{20}$};
\node [blue] at (-0.5,1.5){$v_{11}$};
\node [blue] at (0.5,1.5){$v_{12}$};
\node [blue] at (1.5,1.5){$v_{13}$};
\node [blue] at (2.5,1.5){$v_{14}$};
\node [blue] at (3.5,1.5){$v_{15}$};
\node [blue] at (-0.5,0.5){$v_{6}$};
\node [blue] at (0.5,0.5){$v_{7}$};
\node [blue] at (1.5,0.5){$v_{8}$};
\node [blue] at (2.5,0.5){$v_{9}$};
\node [blue] at (3.5,0.5){$v_{10}$};
\node [blue] at (-0.5,-0.5){$v_{1}$};
\node [blue] at (0.5,-0.5){$v_{2}$};
\node [blue] at (1.5,-0.5){$v_{3}$};
\node [blue] at (2.5,-0.5){$v_{4}$};
\node [blue] at (3.5,-0.5){$v_{5}$};
\end{scope}
\begin{scope}
\draw[red,line width = 0.5mm] (0.55,1.55) -- (0.55,2.45);
\draw[red,line width = 0.5mm] (0.55,2.45) -- (1.45,2.45);
\draw[red,line width = 0.5mm] (1.45,2.45) -- (1.45,1.55);
\draw[red,line width = 0.5mm] (1.45,1.55) -- (0.45,1.55);
\draw[green,line width = 0.5mm] (0.6,1.55) -- (0.6,2.4);
\draw[green,line width = 0.5mm] (0.6,2.4) -- (1.4,2.4);
\draw[green,line width = 0.5mm] (1.4,2.4) -- (1.4,1.6);
\draw[green,line width = 0.5mm] (1.4,1.6) -- (0.5,1.6);
\draw[red,line width = 0.5mm] (1.45,1.55) -- (1.45,0.45);
\draw[red,line width = 0.5mm] (1.45,0.45) -- (2.45,0.45);
\draw[red,line width = 0.5mm] (2.55,0.45) -- (3.45,0.45);
\draw[red,line width = 0.5mm] (3.45,0.45) -- (3.45,-0.45);
\draw[red,line width = 0.5mm] (3.45,-0.45) -- (2.55,-0.45);
\draw[red,line width = 0.5mm] (2.55,-0.45) -- (2.55,0.45);
\draw[green,line width = 0.5mm] (2.55,0.4) -- (3.4,0.4);
\draw[green,line width = 0.5mm] (3.4,0.4) -- (3.4,-0.4);
\draw[green,line width = 0.5mm] (3.4,-0.4) -- (2.6,-0.4);
\draw[green,line width = 0.5mm] (2.6,-0.4) -- (2.6,0.4);
\draw[orange,line width = 0.5mm] (2.55,0.35) -- (3.4,0.35);
\draw[orange,line width = 0.5mm] (3.4,0.35) -- (3.4,-0.35);
\draw[orange,line width = 0.5mm] (3.4,-0.35) -- (2.6,-0.35);
\draw[orange,line width = 0.5mm] (2.6,-0.35) -- (2.6,0.35);
\end{scope}
\begin{scope}[xshift=1.9in]
\draw[->](0,0) to (0.3,0);
\node at (0.65,0){East};
\draw[->](0,0) to (-0.3,0);
\node at (-0.65,0){West};
\draw[->](0,0) to (0,0.3);
\node at (0,0.48){North};
\draw[->](0,0) to (0,-0.3);
\node at (0,-0.45){South};
\end{scope}
\end{tikzpicture}
\caption{Example of an 
\textcolor{black}{asymmetric effort search} trajectory class. The target makes \textcolor{black}{two round trips of the west crosshatched block and three round trips of the east crosshatched block}.}
\label{fig:rsthree}
\end{figure}

{Here we characterize a target that searches one site more times than another site.  Moreover, we incorporate  the precedence constraint that the target
  first searches the {low} priority region and then the }{high} priority region.  Such asymmetric search trajectories\footnote{Although tangential to our tracking problem,  there is a significant body of research \cite{wolfe2004attributes,asymmetric1,asymmetric2} dealing with
    why human perception is  asymmetric;  \cite{levin1996classifying} shows that  white subjects
were able to detect the presence of a face of another
race among white faces faster than they were to detect a
white face among cross-race faces. Also, humans require significantly more time to search for  some
    features compared to others. In terms of intent, these  translate to the driver of a vehicle searching one site  with more effort than another site.}
    with  precedence constraints reflect that the target poses a higher threat to the 
 second site, and that search of the first site, causally affects search of the second site. The corresponding   \textcolor{black}{asymmetric  search} trajectory class $G^{\text{\rsthree}}$ is as follows:  a target makes $N$ round \textcolor{black}{trips of} block \textcolor{black}{$(i,j)$} followed by $N+\Delta$ round \textcolor{black}{trips of} another block \textcolor{black}{$(m,n)$}. Both $N$ and $\Delta$ are \textcolor{black}{random} positive integers. An example of \textcolor{black}{the trajectory} $G^{\text{\rsthree}}$ is illustrated in \textcolor{black}{Fig.\,{\ref{fig:rsthree}}}. In general, $G^{\text{\rsthree}}$ is defined by the class of strings 
\begin{equation}
\label{eq:rsthree}
\begin{split}
&G^{\text{\rsthree}}=a_1a_2\ldots a_N b_1b_2 \ldots b_m c_1 c_2 \ldots c_{N+\Delta}\\
\begin{tikzpicture}[overlay,remember picture,out=315,in=225,distance=0.4cm]
\draw[->,red,shorten >=3pt,shorten <=3pt] (1.3,0.5) to (1.8,0.5);
\draw[->,red,shorten >=3pt,shorten <=3pt] (1.7,0.5) to (2.2,0.5);
\draw[->,red,shorten >=3pt,shorten <=3pt] (2.4,0.5) to (2.9,0.5);
\draw[->,red,shorten >=3pt,shorten <=3pt] (2.7,0.5) to (3.2,0.5);
\draw[->,red,shorten >=3pt,shorten <=3pt] (3.0,0.5) to (3.5,0.5);
\draw[->,red,shorten >=3pt,shorten <=3pt] (3.3,0.5) to (3.8,0.5);
\draw[->,red,shorten >=3pt,shorten <=3pt] (4.0,0.5) to (4.5,0.5);
\draw[->,red,shorten >=3pt,shorten <=3pt] (4.3,0.5) to (4.8,0.5);
\draw[->,red,shorten >=3pt,shorten <=3pt] (4.7,0.5) to (5.2,0.5);
\draw[->,red,shorten >=3pt,shorten <=3pt] (5.5,0.5) to (6.0,0.5);
\end{tikzpicture}
&a\,b: \operatorname{into}(a)=\operatorname{from}(b)\\
\begin{tikzpicture}[overlay,remember picture,out=315,in=225,distance=0.4cm]
\draw[->,red,shorten >=3pt,shorten <=3pt] (0,0.5) to (0.5,0.5);
\end{tikzpicture}
&a_k=\text{search block } (i,j),k=1,2,\ldots,N\\
&c_k=\text{search block } \textcolor{black}{(m,n)},k=1,2,\ldots,N+\Delta\\
&(i,j)\neq (m,n)
\end{split}
\end{equation}
In \eqref{eq:rsthree}, $b_1 b_2 \ldots b_m$ denotes a string of edges directing from block $(i,j)$ to block \textcolor{black}{$(m,n)$}. In $G^{\text{\rsthree}}$: (i) the length of substring containing $c$ is longer than that containing $a$ (tree dependency) (ii) current edge directs from the vertex that its previous edge directs into (red \textcolor{black}{arrows}, chain dependency). Therefore, the class of spatial trajectories $G^{\text{\rsthree}}$ exhibits \textcolor{black}{a tree-chain dependency}. The CSCFG production rules that serve as a generative model for $G^{\text{\rsthree}}$ are provided in Appendix~\ref{appendix:rule and consistency}. 

\subsection{\textcolor{black}{Example 3: Patrol Trajectory Class}}
\label{subsec:rstwo}
The patrol trajectory class $G^{\text{
\rstwo}}$ specifies \textcolor{black}{a target patrolling several blocks followed by} patrolling the same blocks in the reverse order. $G^{\text{\rstwo}}$ signifies a routine patrol behavior: in the morning, the target moves to some other edge while patrolling specific blocks in a pre-designed order; in the afternoon, it \textcolor{black}{patrols} the same blocks in the reverse order and \textcolor{black}{returns} to its origin. An example of \textcolor{black}{the trajectory class} $G^{\text{\rstwo}}$ is illustrated by the red and green lines in Fig.\,{\ref{fig:roadmap}}. \textcolor{black}{In general,} $G^{\text{\rstwo}}$ is defined by \textcolor{black}{the class of strings (spatial trajectories)}
\begin{equation}
\label{eq:rstwo}
\begin{split}
&G^{\text{\rstwo}}=a_{1}a_{2}\ldots a_{N} c_{N} c_{N-1}\ldots c_{1} (N\geq 2)\text{ with}\\
\begin{tikzpicture}[overlay,remember picture,out=315,in=225,distance=0.4cm]
\draw[->,red,shorten >=3pt,shorten <=3pt] (1.4,0.5) to (1.9,0.5);
\draw[->,red,shorten >=3pt,shorten <=3pt] (1.7,0.5) to (2.2,0.5);
\draw[->,red,shorten >=3pt,shorten <=3pt] (2.4,0.5) to (2.9,0.5);
\draw[->,red,shorten >=3pt,shorten <=3pt] (2.8,0.5) to (3.3,0.5);
\draw[->,red,shorten >=3pt,shorten <=3pt] (3.2,0.5) to (3.7,0.5);
\draw[->,red,shorten >=3pt,shorten <=3pt] (3.9,0.5) to (4.4,0.5);
\draw[->,red,shorten >=3pt,shorten <=3pt] (4.6,0.5) to (5.0,0.5);
\end{tikzpicture}
&a\,b: \operatorname{into}(a)=\operatorname{from}(b)\\
\begin{tikzpicture}[overlay,remember picture,out=315,in=225,distance=0.4cm]
\draw[->,red,shorten >=3pt,shorten <=3pt] (0,0.5) to (0.5,0.5);
\end{tikzpicture}
&a_k,c_k=\textcolor{black}{\text{search block }} (i_k,j_k),k=1,2,\ldots,N\\
&(i_{k-1},j_{k-1})\neq(i_k,j_k),k=2,3,\ldots,N
\end{split}
\end{equation}
\textcolor{black}{In \eqref{eq:rstwo}, $G^{\text{\rstwo}}$ is a palindrome (tree dependency) with chain constraints (red arrows)}. Therefore, the class of spatial trajectories $G^{\text{\rstwo}}$ exhibits \textcolor{black}{a tree-chain dependency}. The CSCFG production rules that constitute a generative model for $G^{\text{\rstwo}}$ are provided in Appendix~\ref{appendix:rule and consistency}.

\subsection{\textcolor{black}{Anomalous Trajectories Involving Interacting Targets}}
Apart from the 3 \textcolor{black}{examples} above, the CSCFG formalism also serves as a generative model for several other \textcolor{black}{important} classes of anomalous \textcolor{black}{spatial} trajectories. Here we briefly outline two examples that consist of two \textcolor{black}{interacting} targets (To save space the production rules that generate these trajectories \textcolor{black}{and numerical studies} are omitted).
\subsubsection*{\textcolor{black}{Example 4. Accompany Rectangle}}
\textcolor{black}{The} accompany rectangle trajectory class $G^{\text{ar}}$ \textcolor{black}{consists of two interacting targets. The trajectory of } target \textcolor{black}{$A$} is a rectangle which indicates surveillance activities. Target \textcolor{black}{$B$} accompanies \textcolor{black}{$A$} and moves within one block away from \textcolor{black}{$A$} to do auxiliary work, e.g., clear safety threats around \textcolor{black}{$A$}. An example of $G^{\text{ar}}$ is illustrated in Fig.\,{\ref{fig:other trajectory}}. $G^{\text{ar}}$ is defined by the class of spatial trajectories
\begin{equation}
\nonumber
\begin{split}
&G^{\textnormal{ar}}=q_{1:M}, q_k=(q_k^A;q_k^B),k=1,2,\ldots,M\\
&\text{Target }A: q_{1:M}^A=q_1^A,q_2^A,\ldots,q_M^A=a_{1:m}b_{1:n_1}c_{1:m}d_{1:n_2}\text{  where}\\
&\theta(a_k)=\text{north},\theta(c_k)=\text{south},k=1,2,\ldots,m\\
&\theta(b_k)=\text{east},k=1,2,\ldots,n_1\\
&\theta(d_k)=\text{west},k=1,2,\ldots,n_2\\
&\operatorname{from}(a_1)=\operatorname{into}(d_{n_2})\\
&\text{Target }B: q_{1:M}^B=(q_1^B,q_2^B,\ldots,q_M^B)\text{ where }\\
&q_k^B \textcolor{black}{\text{ is within 1 block away from }} q_k^A, k=1,2,\ldots,M
\end{split}
\end{equation}
{Here $q_k^A$ denotes the road target $A$ is on at time $k$}
and  $\theta(.)$ is defined in \eqref{eq:graph}. For target $A$, the length of $a_{1:m}$ and $c_{1:m}$ (\textcolor{black}{$m$ is a random positive integer}) are equal which shows the tree dependency. \textcolor{black}{The trajectory of target $B$ is a Markov chain}. Hence, \textcolor{black}{the class of spatial trajectories} $G^{\text{ar}}$ exhibits \textcolor{black}{a tree-chain dependency} and a CSCFG is a generative model. 

\subsubsection*{\textcolor{black}{Example 5. Following Palindrome}}
\begin{figure}
\centering
\begin{tikzpicture}[scale=0.95,circle dotted/.style={dash pattern=on .05mm off 1mm,
                              line cap=round}]
\tikzstyle{vertex}=[rectangle,fill=gray!50,minimum size=0.9cm]
\begin{scope}
\node at (0,0)[vertex]{ };
\node at (1,0)[vertex]{ };
\node at (2,0)[vertex]{ };
\node at (3,0)[vertex]{ };
\node at (0,1)[vertex]{ };
\node at (1,1)[vertex]{ };
\node at (2,1)[vertex]{ };
\node at (3,1)[vertex]{ };
\node at (0,2)[vertex]{ };
\node at (1,2)[vertex]{ };
\node at (2,2)[vertex]{ };
\node at (3,2)[vertex]{ };
\node at (0,3)[vertex]{ };
\node at (1,3)[vertex]{ };
\node at (2,3)[vertex]{ };
\node at (3,3)[vertex]{ };

\draw [-,dashed](0.5,-0.4) to (0.5,0.4);
\draw [-,dashed](1.5,-0.4) to (1.5,0.4);
\draw [-,dashed](2.5,-0.4) to (2.5,0.4);

\draw [-,dashed](0.5,0.6) to (0.5,1.4);
\draw [-,dashed](1.5,0.6) to (1.5,1.4);
\draw [-,dashed](2.5,0.6) to (2.5,1.4);

\draw [-,dashed](0.5,1.6) to (0.5,2.4);
\draw [-,dashed](1.5,1.6) to (1.5,2.4);
\draw [-,dashed](2.5,1.6) to (2.5,2.4);

\draw [-,dashed](0.5,2.6) to (0.5,3.4);
\draw [-,dashed](1.5,2.6) to (1.5,3.4);
\draw [-,dashed](2.5,2.6) to (2.5,3.4);

\draw [-,dashed](-0.5,2.6) to (-0.5,3.4);
\draw [-,dashed](-0.5,1.6) to (-0.5,2.4);
\draw [-,dashed](-0.5,0.6) to (-0.5,1.4);
\draw [-,dashed](-0.5,-0.4) to (-0.5,0.4);

\draw [-,dashed](3.5,2.6) to (3.5,3.4);
\draw [-,dashed](3.5,1.6) to (3.5,2.4);
\draw [-,dashed](3.5,0.6) to (3.5,1.4);
\draw [-,dashed](3.5,-0.4) to (3.5,0.4);

\draw [-,dashed](-0.4,0.5) to (0.4,0.5);
\draw [-,dashed](0.6,0.5) to (1.4,0.5);
\draw [-,dashed](1.6,0.5) to (2.4,0.5);
\draw [-,dashed](2.6,0.5) to (3.4,0.5);

\draw [-,dashed](-0.4,1.5) to (0.4,1.5);
\draw [-,dashed](0.6,1.5) to (1.4,1.5);
\draw [-,dashed](1.6,1.5) to (2.4,1.5);
\draw [-,dashed](2.6,1.5) to (3.4,1.5);

\draw [-,dashed](-0.4,2.5) to (0.4,2.5);
\draw [-,dashed](0.6,2.5) to (1.4,2.5);
\draw [-,dashed](1.6,2.5) to (2.4,2.5);
\draw [-,dashed](2.6,2.5) to (3.4,2.5);

\draw [-,dashed](-0.4,3.5) to (0.4,3.5);
\draw [-,dashed](0.6,3.5) to (1.4,3.5);
\draw [-,dashed](1.6,3.5) to (2.4,3.5);
\draw [-,dashed](2.6,3.5) to (3.4,3.5);

\draw [-,dashed](-0.4,-0.5) to (0.4,-0.5);
\draw [-,dashed](0.6,-0.5) to (1.4,-0.5);
\draw [-,dashed](1.6,-0.5) to (2.4,-0.5);
\draw [-,dashed](2.6,-0.5) to (3.4,-0.5);


\node [blue] at (-0.5,3.5){$v_{21}$};
\node [blue] at (0.5,3.5){$v_{22}$};
\node [blue] at (1.5,3.5){$v_{23}$};
\node [blue] at (2.5,3.5){$v_{24}$};
\node [blue] at (3.5,3.5){$v_{25}$};

\node [blue] at (-0.5,2.5){$v_{16}$};
\node [blue] at (0.5,2.5){$v_{17}$};
\node [blue] at (1.5,2.5){$v_{18}$};
\node [blue] at (2.5,2.5){$v_{19}$};
\node [blue] at (3.5,2.5){$v_{20}$};

\node [blue] at (-0.5,1.5){$v_{11}$};
\node [blue] at (0.5,1.5){$v_{12}$};
\node [blue] at (1.5,1.5){$v_{13}$};
\node [blue] at (2.5,1.5){$v_{14}$};
\node [blue] at (3.5,1.5){$v_{15}$};

\node [blue] at (-0.5,0.5){$v_{6}$};
\node [blue] at (0.5,0.5){$v_{7}$};
\node [blue] at (1.5,0.5){$v_{8}$};
\node [blue] at (2.5,0.5){$v_{9}$};
\node [blue] at (3.5,0.5){$v_{10}$};

\node [blue] at (-0.5,-0.5){$v_{1}$};
\node [blue] at (0.5,-0.5){$v_{2}$};
\node [blue] at (1.5,-0.5){$v_{3}$};
\node [blue] at (2.5,-0.5){$v_{4}$};
\node [blue] at (3.5,-0.5){$v_{5}$};
\end{scope}

\draw[red,line width = 0.5mm] (0.55,0.45) -- (0.55,2.45);
\draw[red,line width = 0.5mm] (0.55,2.45) -- (2.45,2.45);
\draw[red,line width = 0.5mm] (2.45,2.45) -- (2.45,0.55);
\draw[red,line width = 0.5mm] (2.45,0.55) -- (0.55,0.55);

\draw[green,line width = 0.5mm] (0.55,0.55) -- (-0.45,0.55);
\draw[green,line width = 0.5mm] (-0.45,0.55) -- (-0.45,1.45);
\draw[green,line width = 0.5mm] (-0.45,1.45) -- (0.55,1.45);
\draw[green,line width = 0.5mm] (0.58,1.45) -- (0.58,3.45);
\draw[green,line width = 0.5mm] (0.58,3.45) -- (1.45,3.45);
\draw[green,line width = 0.5mm] (1.45,3.45) -- (1.45,2.55);
\draw[green,line width = 0.5mm] (1.55,2.55) -- (1.55,3.45);
\draw[green,line width = 0.5mm] (1.55,3.45) -- (2.45,3.45);
\draw[green,line width = 0.5mm] (2.45,3.45) -- (2.45,2.45);
\draw[green,line width = 0.5mm] (2.45,2.45) -- (3.45,2.45);
\draw[green,line width = 0.5mm] (3.45,2.45) -- (3.45,1.55);
\draw[green,line width = 0.5mm] (3.45,1.55) -- (1.45,1.55);
\draw[green,line width = 0.5mm] (1.45,1.55) -- (1.45,0.45);
\draw[orange,line width = 0.5mm] (-0.45,1.45) -- (-0.45,3.45);
\draw[orange,line width = 0.5mm] (-0.45,3.45) -- (0.45,3.45);
\draw[orange,line width = 0.5mm] (0.45,3.55) -- (-0.55,3.55);
\draw[orange,line width = 0.5mm] (-0.55,3.55) -- (-0.55,1.45);

\draw[black,densely dotted,line width=0.6mm] (1.55,-0.45) -- (2.45,-0.45);
\draw[black,loosely dotted,line width=0.6mm] (0.45,-0.45) -- (1.55,-0.45);
\draw[black,loosely dotted,line width=0.6mm] (2.45,-0.45) -- (3.45,-0.45);

\draw [fill=orange,orange] (1.7,-0.4) rectangle (2.3,0); 
\node at (2,-0.2) {\text{embassy}};

\begin{scope}[xshift=1.8in]
\draw[->](0,0) to (0.3,0);
\node at (0.65,0){East};
\draw[->](0,0) to (-0.3,0);
\node at (-0.65,0){West};
\draw[->](0,0) to (0,0.3);
\node at (0,0.48){North};
\draw[->](0,0) to (0,-0.3);
\node at (0,-0.45){South};
\end{scope}
\end{tikzpicture}
\caption{Red (target \textcolor{black}{$A$}) and green (target \textcolor{black}{$B$}) lines illustrate an example of the accompany rectangle trajectory class $G^{\text{ar}}$. Orange lines depict an example of a palindrome trajectory.}
\label{fig:other trajectory}
\end{figure}
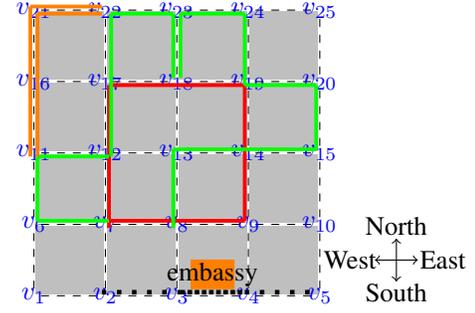

In the palindrome anomalous trajectory class $G^{\text{fp}}$, target $A$ starts from road $e_{ij}$, moves to road $e_{mn}$ and then retraces its path to $e_{ji}$. The trajectory of target $A$ is a palindrome illustrated in Fig.\,{\ref{fig:other trajectory}}. Target $A$ can be a \textcolor{black}{person} that goes from home to \textcolor{black}{office} and then retraces his path to back home. Target $B$ intentionally follows target $A$ by moving on either the same road as $A$ or the previous road $A$ travelled (so $B$ can constantly monitor $A$). $G^{\text{fp}}$ is defined by \textcolor{black}{the class of spatial trajectories}
\begin{equation}
\begin{split}
&G^{\textnormal{fp}}=q_{1:M}, q_k=(q_k^A;q_k^B),k=1,2,\ldots,M\\
&\text{Target }A: q_{1:M}^A=q_1^A,q_2^A,\ldots,q_M^A=a_{1:m}c_{m:1}\text{  where}\\
&\operatorname{from}(a_k)=\operatorname{into}(c_k), \operatorname{into}(a_k)=\operatorname{from}(c_k),k=1,2,\ldots,M\\
&\text{Target }B: q_{1:M}^B=(q_1^B,q_2^B,\ldots,q_M^B)\text{ where }\\
&q_k^B=q_k^A \text{ or } q_k^B=q_{k-1}^A, k=1,2,\ldots,M\nonumber
\end{split}
\end{equation}
Here, $m$ is a random positive integer. The trajectory of target $A$ is a palindrome which represents the tree dependency\cite{SCFG_metalevel_modeling}. The trajectory of target $B$ is a Markov chain which indicates the chain dependency. Therefore, the class  $G^{\text{fp}}$ which combines the trajectories of targets $A$ and $B$ exhibits \textcolor{black}{a tree-chain dependency} and a CSCFG forms a generative model.

\subsection{Summary}  We summarize the main ideas in this section with the following  theorem and then describe some extensions.
\begin{thm} For the spatial trajectories in Examples 1 to 5: \begin{compactenum} \item A CSCFG constitutes a generative model, i.e., the necessary and sufficient condition (\ref{eq:generative model})
holds.  Markov chains are not generative models and naive template matching (Sec.\ref{subsec:compare MC, SCFG and CSCFG})  requires an exponential number of templates. 
\item The CSCFG models are consistent: they generate finite length strings with probability one (proof in Appendix~\ref{appendix:rule and consistency}). \end{compactenum}
\end{thm}

{\textbf{Velocity Tracklets}.} A target's velocity profile can  also indicate anomalous activities. In Fig.\,{\ref{fig:other trajectory}}, the target moves slowly on the road (densely dotted black dots) close to the embassy (orange filled rectangle) and then  speeds up (loosely dotted black dots) after it turns into another road. This velocity profile can indicate suspicious surveillance of the embassy. To incorporate the target's velocity into syntactic tracking, we  simply augment the position tracklet  (road name $e_{ij}$) with
the quantized speed, e.g.,: $[e_{ij},\text{slow}]$ or $[e_{ij},\text{fast}]$  which denotes the target moving on road $e_{ij}$ at a low or high speed. The base-level tracker estimates the target's  speed on an individual road, maps it to the augmented tracklet  and then the meta-level tracker uses these tracklets for inferring intent.

{\textbf{Other Examples of Anomalous Trajectories.}} In general anomalous  trajectories are defined by radar operators and can be codified into  production rules. \cite{SCFG_metalevel_modeling} has several
additional examples of  trajectories that have SCFG generative models  including move-stop-move, two targets meeting at a specific region and then departing, trajectories that are constrained to visit specific regions such as checkpoints, etc.
\section{CSCFG-driven Particle Filter Tracker}
\label{sec:algorithm}
Thus far, we have discussed a 3-level model for the roadmap syntactic tracking problem. Sec.\,{\ref{sec:level2} gave several examples of anomalous trajectories.} In this section, given GMTI radar observations $\mathbf{z}_{1:k}$ of the target, we exploit the 3-level model to develop a syntactic enhanced VSIMM algorithm for real time estimation of the posterior of the target trajectory as\footnote{{We constrain $p(G|q_{1:k})=p(G|\unique(q_{1:k}))$ by setting $p(G|q_{1:k})$ to zero if $q_{1:k}$ is not a feasible tracklet (road name) sequence, e.g., the probability of jumping to a non adjacent road is zero.}}
\begin{equation}
\label{eq:aim again}
p(G|\mathbf{z}_{1:k})=\underset{q_{1:k}}{\sum}p(G|\unique(q_{1:k}))p(q_{1:k}|\mathbf{z}_{1:k}),\forall G\in\mathbf{G}\\
\end{equation}
Recall \eqref{eq:qk new} that  $\unique(q_{1:k})$ is the sequence of distinct roads traversed from time 1 to $k$.  \textcolor{black}{$\mathbf{G}$ denotes the set of trajectory models discussed in Sec.\,{\ref{sec:level2}}}. Evaluating $p(G|\mathbf{z}_{1:k})$ in \eqref{eq:aim again} requires  computing the posterior distributions $p(G|\unique(q_{1:k}))$ and $p(q_{1:k}|\mathbf{z}_{1:k})$. These are discussed below.
\subsection{Context}
\label{subsec:3 steps}
Given the architecture of CSCFG syntactic tracker in Fig.\,{\ref{fig:architecture}}(b) and \eqref{eq:aim again}, the computation of the posterior
$p(G|\mathbf{z}_{1:k})$ of trajectory class $G$ \textcolor{black}{proceeds in} three \textbf{steps}: 

(1) The meta-level tracker computes the one step prediction $p(q_k|q_{1:k-1})$ and feeds it to the base-level tracker. These probabilities
depend on the road traffic statistics and are computed  in (\ref{eq:sample q}) below (via the Earley Stolcke parser).

(2) \textcolor{black}{The base-level tracker receives the one step prediction, computes the posterior $p(q_{1:k}|\mathbf{z}_{1:k}$) in \eqref{eq:aim again} given radar observations $\mathbf{z}_{1:k}$ and returns $q_k$ to the meta-level tracker.} 

(3) \textcolor{black}{The meta-level tracker receives $q_k$ and computes the posterior} $p(G|\unique(q_{1:k}))$ \textcolor{black}{in \eqref{eq:aim again}} as
\begin{equation}
\label{eq:G|q_{1:k}}
p(G|\unique (q_{1:k}))=\frac{p(\unique (q_{1:k})|G)p(G)}{\underset{G\in\mathbf{G}}{\sum}p(\unique (q_{1:k})|G)p(G)}
\end{equation}
Here, $p(G),\forall G\in\mathbf{G}$ is the pre-defined prior distribution of different classes of trajectories. {$p(\unique(q_{1:k})|G)$ denotes the 
likelihood of the road sequence given the grammatical model.}

The particle filtering algorithm described in Sec.\,{\ref{subsec:algorithm}} is based on the \textcolor{black}{above 3 steps} to estimate the posterior distribution $p(G|\mathbf{z}_{1:k})$ in \eqref{eq:aim again}. The novelty of this algorithm is that it exploits the CSCFG natural language model for the target trajectory class as follows: a modified Earley Stolcke parser\footnote{The Earley Stolcke parser is used for forward parsing of SCFGs\cite{earley_stolcke_parser}. In Appendix~\ref{appendix:parsing} we show how it can be modified to parse CSCFGs. \textcolor{black}{The novelty of this modified parser is that it adds serial constraints to the top-down control structure in the Earley Stolcke parser while maintaining  polynomial computation cost.}} for CSCFGs (see Appendix~\ref{appendix:parsing}) is used to compute the one step prediction $p(q_k|q_{1:k-1})$ in \textbf{step} (1)  and target trajectory likelihood $p(\unique(q_{1:k})|G)$ in \textbf{step} (3).

\subsection{CSCFG-driven Particle Filter Algorithm}
\label{subsec:algorithm}
For the 3-level model proposed in Sec.\,{\ref{sec:3 level model}} and Sec.\,{\ref{sec:level2}}, direct computation of the posterior distribution $p(q_{1:k}|\mathbf{z}_{1:k})$ is intractable.\footnote{Note that \eqref{eq:state equ} is a CSCFG-driven  dynamical system. Even for simpler case of a jump Markov linear system, exact computation of the posterior is intractable since the posterior at time $k$ is a mixture distribution with an exponential number of components in $k$. Also see \cite{djuric2010} for adaptive systems of particle filters.}
Therefore, given the noisy observation sequence $\mathbf{z}_{1:k}$, we estimate the posterior distribution
\begin{equation}
p(q_{1:k},\mathbf{x}_{1:k},m_{1:k+1}|\mathbf{z}_{1:k})\nonumber
\end{equation}
Here, \textcolor{black}{$m_{1:k+1}=(m_1,m_2,\ldots,m_{k+1})$ and $m_k$ denotes the length of \textcolor{black}{the directed road sequence} $\unique(q_{1:k})$ in \eqref{eq:qk new}. \textcolor{black}{That is}, $m_k$ is the number of distinct roads the target has traversed until time $k$. Recall $q_k$ is the road the target is on at time $k$.} 

\textcolor{black}{The} particle filter algorithm \text{below uses} the  approximation 
\begin{equation}
\begin{split}
&p(q_{1:k},\mathbf{x}_{1:k},m_{1:k+1}|\mathbf{z}_{1:k})\\
&\approx\displaystyle{\sum_{i=1}^{N_p}}\mathbbm{1}(q_{1:k}^i=q_{1:k},\mathbf{x}_{1:k}^i=\mathbf{x}_{1:k},m_{1:k+1}^i=m_{1:k+1})w_{k}^i\nonumber
\end{split}
\end{equation}
where $N_p$ is the number of particles. Then the real time {posterior probability mass function} of trajectory $G$ given radar observations in \eqref{eq:aim again} is computed as
\begin{equation}
\label{eq:particle and G}
{p(G|\mathbf{z}_{1:k})}=\sum_{i=1}^{N_p}p(G|\unique(q_{1:k}^i))w_{k}^i
\end{equation}
The particle weights $w_k^{i}$ are computed recursively as
\begin{flalign*}
& w_k^{i}=\frac{p(q_{1:k}^i,\mathbf{x}_{1:{k}}^i,m_{1:k+1}^i|\mathbf{z}_{1:k})}{\pi(q_{1:k}^i,\mathbf{x}_{1:k}^i,m_{1:k+1}^i|\mathbf{z}_{1:k})}\nonumber\\
&\propto\frac{p(q_k^i,\mathbf{x}_{k}^i,m_{k+1}^i|q_{1:k-1}^i,\mathbf{x}_{1:k-1}^i,m_{1:k}^i)p(\mathbf{z}_{k}|\mathbf{x}_{k}^i)}{\pi(q_k^i,\mathbf{x}_{k}^i,m_{k+1}^i|q_{1:k-1}^i,\mathbf{x}_{1:k-1}^i,m_{1:k}^i,\mathbf{z}_{1:k})}w_{k-1}^i
\end{flalign*}

\textcolor{black}{We} choose the bootstrap proposal distribution 
\begin{align}
&\pi(q_k^i,\mathbf{x}_{k}^i,m_{k+1}^i|q_{1:k-1}^i,\mathbf{x}_{1:k-1}^i,m_{1:k}^i,\mathbf{z}_{1:k})\nonumber\\
&=p(q_k^i,\mathbf{x}_{k}^i,m_{k+1}^i|q_{1:k-1}^i,\mathbf{x}_{1:k-1}^i,m_{1:k}^i)\nonumber\\
&=p(q_k^i|q_{1:k-1}^i)p(\mathbf{x}_{k}^i|\mathbf{x}_{k-1}^i,q_k^i)p(m_{k+1}^i|m_{k}^i,\mathbf{x}_{k}^i,q_k^i)\label{eq:proposal}
\end{align}
The first term in \eqref{eq:proposal} is computed as\footnote{{The particles  $q_k^i$, $i=1, \ldots, N_p$ are sampled to ensure that they are consistent with the specific CSCFG trajectory model.}}
\begin{equation}
\label{eq:sample q}
\begin{split}
&p(q_k^i|q_{1:k-1}^i)=\\
&\begin{cases}
\sum\limits_{\forall G\in\mathbf{G}}p(q_k^i| \unique(q_{1:k-1}^i),G)p(G|\unique(q_{1:k-1}^i)) & \text{if }\textcolor{black}{m_{k}^i=m_{k-1}^i+1}\\
\mathbbm{1}(q_k^i=q_{k-1}^i) & \text{if }m_k^i=m_{k-1}^i\\
\end{cases}
\end{split}
\end{equation}
In \eqref{eq:sample q}, $p(q_k^i| \unique(q_{1:k-1}^i),G)$ is computed by the modified Earley Stolcke parser for \textcolor{black}{CSCFGs as described in Appendix~\ref{appendix:parsing}; see \eqref{eq:one step predictor}}. \textcolor{black}{The event} $m_k^i=m_{k-1}^i+1$ indicates that the target enters an intersection and needs to decide \textcolor{black}{the next} road to move onto. $p(G|\unique (q_{1:k-1}^i))$ is the posterior distribution of trajectory in \eqref{eq:G|q_{1:k}}. 

\textcolor{black}{Using the dynamics} in $\eqref{eq:state equ}$, the second term in (\ref{eq:proposal}) is
\begin{equation}
\label{eq:sample x}
p(\mathbf{x}_{k}^i|\mathbf{x}_{k-1}^i,q_k^i)=\mathcal{N}(\mathcal{F}(\mathbf{x}_{k-1}^i,d_{k}^i), \tilde Q_{k}(d_{k}^i))
\end{equation}
Here, $d_{k}^i$ is specified in \eqref{eq:state equ}. $\mathcal{F}(.)$ and $\tilde{Q}_{k}(d_{k}^{i})$ are defined in \eqref{eq:mathcal F}, \eqref{eq:Q}. The third term in \eqref{eq:proposal} is \textcolor{black}{updated} as
\begin{equation}
\label{eq:mk}
\begin{split}
&m_{k+1}^i=\\
&\begin{cases}
m_k^i+1 & \mathcal{B}(F\mathbf{x}_k^i)=\operatorname{into}(q_k^i)\\
m_k^i   & \mathcal{B}(F\mathbf{x}_k^i)=q_k^i
\end{cases}\text{ with }F=
\begin{bmatrix}
1&0&T&0\\
0&1&0&T\\
0&0&1&0\\
0&0&0&1
\end{bmatrix}
\end{split}
\end{equation}
Here, \textcolor{black}{$\mathcal{B}(.)$ is defined in \eqref{eq:state equ} and $T$ is the sampling interval in \eqref{eq:mathcal F}}. \eqref{eq:mk} says that $m_k^i$ is incremented by~1 if $F\mathbf{x}_k^i$ is located at the intersection $\operatorname{into}(q_k^i)$.
Recursive computation of the particle weights $w_k^i$ before normalization is
\begin{equation}
\label{eq:weight}
w_k^i=w_{k-1}^i p(\mathbf{z}_{k}|\mathbf{x}_{k}^i)=w_{k-1}^i\mathcal{N}(\mathcal{H}(\mathbf{x}_{k}^i,\mathbf{c}_k),R) 
\end{equation}
where $\mathcal{H}(.)$, $\mathbf{c}_{k}$ are defined in \eqref{eq:obs equ} and $R$ is defined in \eqref{eq:R}.

With the above definitions, the CSCFG-driven particle filtering algorithm is presented in Algorithm~\ref{algorithm:PF}. The output of Algorithm~\ref{algorithm:PF} is the real time \textcolor{black}{estimate of the posterior} of each trajectory model given radar observations, namely $p(G|\mathbf{z}_{1:k}),\forall G\in\mathbf{G}$ in \eqref{eq:aim again}. Algorithm\,{\ref{algorithm:PF}} also computes the target's state estimate (denoted as $\hat {\mathbf{x}}_{k}$) to compare with the VSIMM tracker in Sec.\,{\ref{sec:simulation}.

\begin{algorithm}
\caption{Syntactic Target Tracking Algorithm using CSCFG-Driven Particle Filter}
\begin{algorithmic}
\label{algorithm:PF}
\STATE{CSCFG-Particle Filter\\
\textbf{Input} $\{\mathbf{x}_{1:k-1}^i,q_{1:k-1}^i, m_{1:k}^i, w_{k-1}^i, \mathbf{z}_{k}\}$}\\
\FOR{$i=1$ \TO $N_{p}$}
\STATE{Sample $q_k^i$ using \eqref{eq:sample q}\\
$q_{1:k}^i=(q_{1:k-1}^i,q_{k}^i)$\\
Sample $\mathbf{x}_{k}^i$ using \eqref{eq:sample x}\\
Compute $m_{k+1}^i$ using \eqref{eq:mk}\\
Compute $w_{k}^i$ using \eqref{eq:weight}}
\ENDFOR\\
Normalize $w_k^i=w_k^i/\sum_{i=1}^{N_p}w_k^i,\forall i=1,2,\cdots,N_{p}$\\
\IF{$1/\sum_{i=1}^{N_p}(w_k^{i})^2<\textcolor{black}{\text{threshold}}$}\STATE{RESAMPLE}\ENDIF\\
\textcolor{black}{Target's state estimate $\hat {\mathbf{x}}_{k}=\sum_{i=1}^{N_p}\mathbf{x}_{k}^iw_k^i$}\\
Compute \textcolor{black}{the real time trajectory posterior} $p(G|\mathbf{z}_{1:k})$ using~\eqref{eq:particle and G}\\
\RETURN {$\{\mathbf{x}_{1:k}^i, q_{1:k}^i, m_{1:k+1}^i, w_{k}^i, p(G|\mathbf{z}_{1:k}),\forall G\in~\mathbf{G}$\}}
\end{algorithmic}
\end{algorithm}

\section{Numerical Examples}
\label{sec:simulation}
\textcolor{black}{This section presents} three numerical examples \textcolor{black}{for} detecting roadmap based anomalous trajectories of a target. \textcolor{black}{The CSCFG} syntactic tracker (\textcolor{black}{Algorithm~\ref{algorithm:PF}}) proposed in Sec.\,{\ref{sec:algorithm}} provides real time classification of the trajectory of a target as it moves on a roadmap given GMTI radar observations. \textcolor{black}{Recall that Algorithm~\ref{algorithm:PF} combines the Earley Stolcke parser (\textcolor{black}{detailed in Appendix~\ref{appendix:parsing}}) with a base-level particle filter to compute }the posterior probabilities of trajectory models.

\subsection{\textcolor{black}{Simulation Setup}}
The prior probabilities of the anomalous trajectories of interest are assumed to be uniformly distributed. The initialization of the $i$th particle $s_1^i=(x_1^i;q_1^i)$ is \textcolor{black}{specified as}
\begin{equation}
\nonumber
x_1^i\sim\mathcal{N}(\textcolor{black}{\mathbf{x}_1},\text{diag}(25,25,5,5)),\,
q_1^i=\mathcal{B}(\mathbf{x}_1^i)
\end{equation}
Here, \textcolor{black}{$\mathbf{x}_1$} is the target's true state vector at the initial time and \textcolor{black}{$\mathcal{B}(.)$ defined in \eqref{eq:state equ} denotes the target's meta-level position (road or intersection name).}

The empirical performance of the CSCFG syntactic tracker is evaluated by simulating measurements from a GMTI radar \textcolor{black}{for} a target moving on a 6$\times$6 (number of blocks) roadmap; see Fig.\,{\ref{fig:simu,roadmap}}. The length and width of each directed road \textcolor{black}{is} assumed to be \textcolor{black}{100~m and 5~m}, respectively. The aircraft (\textcolor{black}{on which a GMTI radar is mounted)} starts from 
 (-500 m, -500 m), and then moves along the perimeter of the roadmap with (constant) altitude 3000~m and constant speed 100 m/s; see Fig.\,{\ref{fig:simu,roadmap}}. Table~\ref{table:parameter} specifies the kinematic parameters used in the particle filter. 
\begin{table}[ht]
\centering
\caption{Parameters used in numerical examples} 
\label{table:parameter} 
\begin{tabular}{c c} 
\hline\hline 
$T$ & 0.2s\\ 
number of particles& 1000 \\
$\sigma_{r_k}$ & $5\sim 9$ m \\
$\sigma_{\dot r_k}$ & $0.3\sim 0.7 $ m/s \\ 
$\sigma_{\textcolor{black}{\theta_k}}$ & $0.5^\circ\sim 0.9^\circ$ \\ 
$\sigma_{a}$ & 0.5 m/$\text{s}^2$\\
$\sigma_{o}$  &0.05 m/$\text{s}^2$\\
\hline 
\end{tabular}
\begin{tablenotes}
\item $\sigma_{r_k}$, $\sigma_{\dot r_k}$ and $\sigma_{\textcolor{black}{\theta_k}}$ are the standard deviations of the range, range rate and azimuth in \eqref{eq:R}. $T$ is the sampling interval in \eqref{eq:mathcal F}. $\sigma_{a}$ and $\sigma_{o}$ are the standard deviations along and orthogonal to the direction of motion in \eqref{eq:Q}. 
\end{tablenotes}
\end{table}

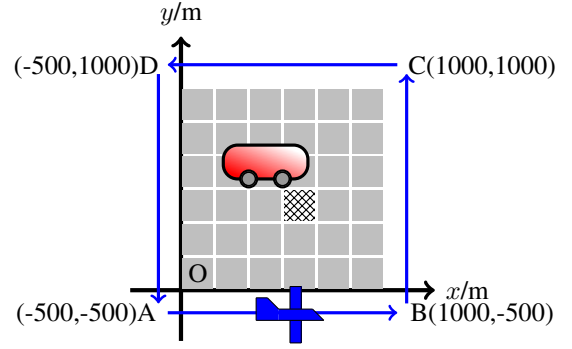
\begin{figure}
\centering
\begin{tikzpicture}[scale=1,circle dotted/.style={dash pattern=on .04mm off 1mm,
                              line cap=round}]
\tikzstyle{vertex}=[rectangle,fill=gray!50,minimum size=0.4cm]
\tikzstyle{crosshatch_vertex}=[rectangle,fill=gray!50,minimum size=0.4cm,pattern=crosshatch]
\begin{scope}[scale=0.75]
\begin{scope}[scale=0.6]

\node at (8.4,-1.2){B(1000,-500)};

\node at (8.4,6.1){C(1000,1000)};

\node at (-3.3,6.1){(-500,1000)D};

\node at (-3.3,-1.2){(-500,-500)A};
\node at (0,0){O};
\draw[->,ultra thick] (-2,-0.5)--(7,-0.5)node[right]{$x$/m};
\draw[->,ultra thick] (-0.5,-2)--(-0.5,7) node[above]{$y$/m};

\node at (0,0)[vertex]{ };
\node at (1,0)[vertex]{ };
\node at (2,0)[vertex]{ };
\node at (3,0)[vertex]{ };
\node at (4,0)[vertex]{ };
\node at (5,0)[vertex]{ };

\node at (0,1)[vertex]{ };
\node at (1,1)[vertex]{ };
\node at (2,1)[vertex]{ };
\node at (3,1)[vertex]{ };
\node at (4,1)[vertex]{ };
\node at (5,1)[vertex]{ };

\node at (0,2)[vertex]{ };
\node at (1,2)[vertex]{ };
\node at (2,2)[vertex]{ };
\node at (3,2)[crosshatch_vertex]{ };
\node at (4,2)[vertex]{ };
\node at (5,2)[vertex]{ };

\node at (0,3)[vertex]{ };
\node at (1,3)[vertex]{ };
\node at (2,3)[vertex]{ };
\node at (3,3)[vertex]{ };
\node at (4,3)[vertex]{ };
\node at (5,3)[vertex]{ };

\node at (0,4)[vertex]{ };
\node at (1,4)[vertex]{ };
\node at (2,4)[vertex]{ };
\node at (3,4)[vertex]{ };
\node at (4,4)[vertex]{ };
\node at (5,4)[vertex]{ };

\node at (0,5)[vertex]{ };
\node at (1,5)[vertex]{ };
\node at (2,5)[vertex]{ };
\node at (3,5)[vertex]{ };
\node at (4,5)[vertex]{ };
\node at (5,5)[vertex]{ };
\node at (0,0){O};
\end{scope}

\begin{scope}
\node (a) at (-0.7,-0.7){};
\node (b) at (3.7,-0.7){};
\node (c) at (3.7,3.7){};
\node (d) at (-0.7,3.7){};
\draw[very thick,->,blue] (a) edge (b);
\draw[very thick,->,blue] (b) edge (c);
\draw[very thick,->,blue] (c) edge (d);
\draw[very thick,->,blue] (d) edge (a);
\end{scope}
\begin{scope}[yshift=-0.25in,xshift=0.8in,scale=0.2]
\fill[blue] (-2,0) rectangle (-1,2);
\fill[blue] (-2,-1) rectangle (-1,-3);
\fill[blue] (-3,0)--(0,0)--(1,-1)--(-3,-1);
\fill[blue] (-5,1)--(-4,1)--(-3,0)--(-3,-1)--(-5,-1);
\draw (0,0) --(1,-1);
\draw (0,0) --(-3,0);
\draw (-3,0) --(-4,1);
\draw (-5,1) --(-4,1);
\draw (-5,-1) --(-5,1);
\draw (1,-1) --(-5,-1);
\draw (-2,0) --(-2,2);
\draw (-1,0) --(-1,2);
\draw (-2,-1) --(-2,-3);
\draw (-1,-1) --(-1,-3);
\draw(-2,2)--(-1,2);
\draw(-2,-3)--(-1,-3);
\end{scope}
\begin{scope}[yshift=0.6in,scale=0.3]
\shade[top color=red, bottom color=white, shading angle={135}][draw=black,fill=red!20,rounded corners=1.2ex,very thick] (1.5,.5) rectangle (6.5,2.5);
\draw[draw=black,fill=gray!50,thick] (3,.5) circle (.5);
\draw[draw=black,fill=gray!50,thick] (5,.5) circle (.5);    
\draw[draw=black,fill=gray!80,semithick] (3,.5) circle (.4);
\draw[draw=black,fill=gray!80,semithick] (5,.5) circle (.4);
\end{scope}
\end{scope}
\end{tikzpicture}
\caption{\textcolor{black}{Trajectory of the aircraft that deploys the GMTI radar}. The aircraft starts from \textcolor{black}{A (-500 m, -500 m)} and moves along the perimeter of the 6$\times$ 6 roadmap. O denotes the origin point. \textcolor{black}{Grey squares denote blocks. The crosshatched block is denoted as ``block $(4,3)$" since it is located at the 4th column, 3rd row in the roadmap.}}
\label{fig:simu,roadmap}
\end{figure}

Our results below involve 20 independent simulation trials for each trajectory model \textcolor{black}{over a sweep of increasing observation noise variances}. In each trial: (i) The target's trajectory that follows either $G^{\rsone}$, $G^{\rstwo}$ or $G^{\rsthree}$ discussed in Sec.\,{\ref{sec:level2}} is simulated. The target's speed is initialized to 10 m/s. (ii) A GMTI measurement sequence is simulated. (iii) The CSCFG syntactic tracker is run to compute the real time trajectory posterior {$p(G|\mathbf{z}_{1:k})$} and the state estimate {$\hat x_k$ in Algorithm~\ref{algorithm:PF}}. Finally, 
these state estimates are compared to that of the classical VSIMM tracker.

The performance of our CSCFG syntactic tracker (\textcolor{black}{Algorithm~\ref{algorithm:PF}}) is evaluated using \textcolor{black}{two criteria}:

1) Normalized detection delay defined as
\begin{equation}
\label{eq:normalized detection delay}
\frac{\text{time instant when anomalous trajectory is detected}}
{\text{total time instants of GMTI observations}}
\end{equation}
We consider an anomalous trajectory to be detected when the posterior probability of the trajectory exceeds 0.9 for a period.

2) State estimate improvement which is computed as
\begin{equation}
\label{eq:state estimate improvement}
\begin{split}
&\text{improvement}=\frac{\text{error}(\text{VSIMM})-\text{error}(\text{CSCFG})}{\text{error}(\text{VSIMM})}\%\\
&\text{where }\text{error}=\text{average}((\hat{\mathbf{x}}_{k}^i-\mathbf{x}_k^i)'(\hat{\mathbf{x}}_{k}^i-\mathbf{x}_k^i))\\
&\textcolor{black}{\text{for } } i=1,2,\ldots,M \text{ and } k=1,2,\ldots,N_i
\end{split}
\end{equation}
Here, $\hat{\mathbf{x}}_k^i$, $\mathbf{x}_k^i$ denote the estimated and true state vector of the target in the $i$th trial at time $k$. \textcolor{black}{In \eqref{eq:state estimate improvement}, $M=(20\times 3)$ denotes the total number of simulation trials over all trajectory models. $N_i$ is the total time instants of the $i$th simulation trial}. 

\subsection{\textcolor{black}{Equal Effort Search Trajectory Model}}
\label{subsec:simu,rsone}
The  trajectory model $G^{\rsone}$ was discussed in Sec.\,{\ref{subsec:rsone}} and Fig.\,{\ref{fig:rsone}}. We generated 20 independent simulation trials where the target follows the $G^{\rsone}$ trajectory model. \textcolor{black}{In each trial, we select two rows in the roadmap randomly and \textcolor{black}{choose $N$ to be either $2$, $3$ or $4$ with probability $1/3$. Recall $N$ is the number of blocks to be searched in each row by the target in \eqref{eq:rsone}}. Then we compute the normalized detection \textcolor{black}{delay \eqref{eq:normalized detection delay}}.}

{\em \textcolor{black}{Results.}} $G^{\rsone}$ has the largest average (over 20 simulation trials) normalized detection delay (=80$\%$) compared with $G^{\rstwo}(=43\%)$ and $G^{\rsthree}(=42\%)$; see Fig.\,{\ref{fig:simu_detection}}(a). This is because it takes   {more time to distinguish between $G^{\rsone}$ and $G^{\rstwo}$}. Specifically, the difference between   {the posteriors} $p(G^{\rsone}|\mathbf{z}_{1:k})$ and $p(G^{\rstwo}|\mathbf{z}_{1:k})$   {becomes} sufficiently large only after   {the target searches 4 blocks}. 

Fig.\,{\ref{fig:simu_detection}}(b) illustrates   {how the posteriors of the three trajectory classes, namely, $p(G^{\rsone}|\mathbf{z}_{1:k})$, $p(G^{\rstwo}|\mathbf{z}_{1:k})$ and $p(G^{\rsthree}|\mathbf{z}_{1:k})$ evolve vs time $k$}.   {The posteriors are computed by the CSCFG syntactic tracker (Algorithm~\ref{algorithm:PF})}.   {Fig.\,{\ref{fig:simu_detection}}(b) shows that the posterior $p(G^{\rsthree}|\mathbf{z}_{1:k})$ becomes negligible when the target does not perform consecutive searches of block $B$}. The difference between the posteriors $p(G^{\rsone}|\mathbf{z}_{1:k})$ and $p(G^{\rstwo}|\mathbf{z}_{1:k})$   {remains} sufficiently large after blocks $A$, $B$, $C$, $D$ are searched.

\subsection{\textcolor{black}{Asymmetric Effort Search Trajectory Model}}
\label{subsec:simu,rsthree}
The asymmetric effort search trajectory model is described in Sec.\,{\ref{subsec:rsthree}} and Fig.\,{\ref{fig:rsthree}}. We generated 20 independent simulation trials where the target follows the $G^{\rsthree}$ trajectory model. In each trial, \textcolor{black}{we choose $N$ to be either 1, 2 or 3 with probability 1/3 and $\Delta$ to be either 1 or 2 with probability 1/2. $N$, $\Delta$ are the number of searches of block $A$ and number of additional searches of block $B$ in \eqref{eq:rsthree}}. Then we compute the normalized detection \textcolor{black}{delay \eqref{eq:normalized detection delay}}. 

{\em \textcolor{black}{Results.}} The average (over 20 simulation trials) normalized detection delay of $G^{\rsthree}$ is $42\%$; see Fig.\,{\ref{fig:simu_detection}}(a).   {We present} simulation results based on} $N (=1,2,3)$ and $\Delta (=1,2)$.

1) {$N=1; \Delta=1$; normalized detection delay=100\%}\\
Fig.\,{\ref{fig:simu_detection}}(c) shows an example of $G^{\rsthree}$ detection where the target's trajectory consists of   {$N=1$} search of block $A$ followed by   {$N+\Delta=2$} searches of block $B$.   {$p(G^{\rstwo}|\mathbf{z}_{1:k})$ drops to zero because the target stops its movement (no GMTI radar input) rather than searching block $A$.}

2) {$N=2,3$; $\Delta=1,2$; normalized detection delay$\approx 10\%$}\\
Fig.\,{\ref{fig:simu_detection}}(d)   {displays the posterior probability of trajectory class} $G^{\rsthree}$ where the target makes successive searches ($\geq 2$) of block $A$.   {Based on the posterior probabilities}, $G^{\rsthree}$ is detected during the target's 2nd search of block $A$.   The value of the posteriors $p(G^{\rsone}|\mathbf{z}_{1:k})$ and $p(G^{\rstwo}|\mathbf{z}_{1:k})$ drop   rapidly with $k$ because consecutive searches of block $A$ are not permitted in   the trajectory classes $G^{\rsone}$ and $G^{\rstwo}$.

\subsection{\textcolor{black}{Patrol Trajectory Model}}
\label{subsec:simu,rstwo}
The patrol trajectory model was presented in Sec.\,{\ref{subsec:rstwo}} and Fig.\,{\ref{fig:roadmap}}. We generated 20 independent simulation trials where the target follows the $G^{\rstwo}$ trajectory model. \textcolor{black}{In each trial, \textcolor{black}{we choose $N$ to be either 2, 3 or 4 with probability 1/3 where $N$ is the number of different blocks searched by the target in \eqref{eq:rstwo}}. Then we compute the normalized detection \textcolor{black}{delay \eqref{eq:normalized detection delay}}.}

\begin{figure*}
\centering
\begin{subfigure}{0.47\textwidth}
 \includegraphics[width=\textwidth]{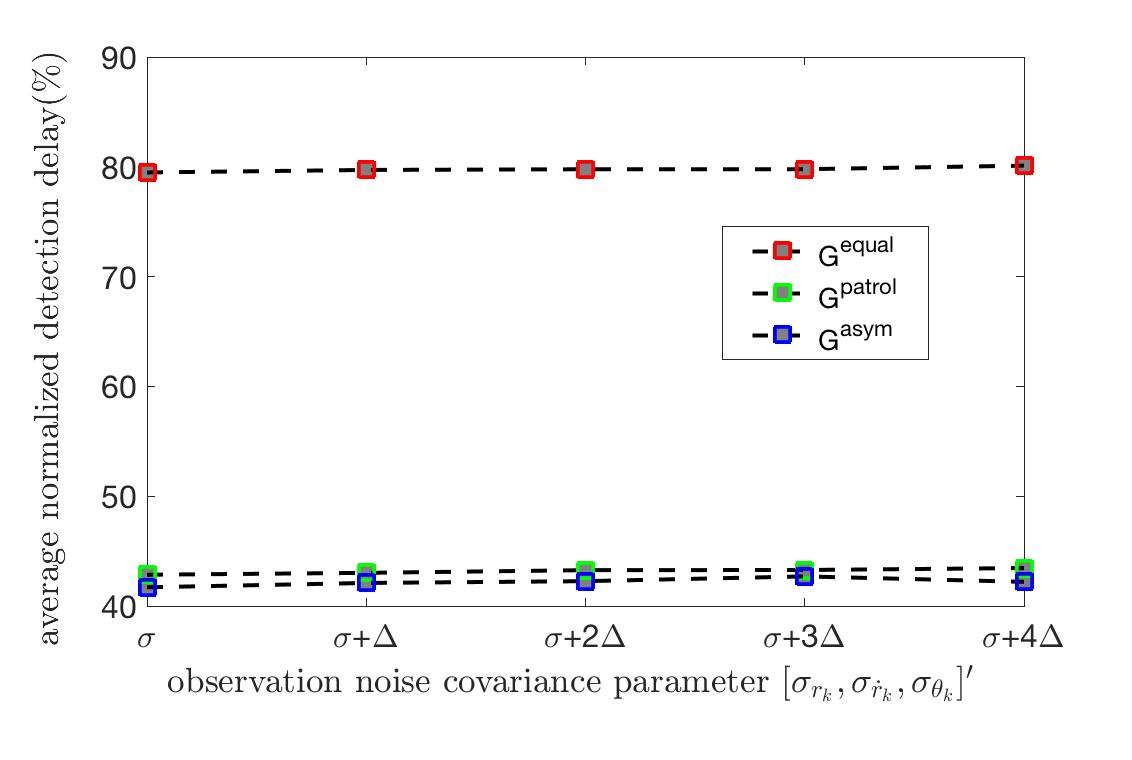}
 
 \caption{}
 
\end{subfigure}%
~
\begin{subfigure}{0.47\textwidth}
 \includegraphics[width=\textwidth]{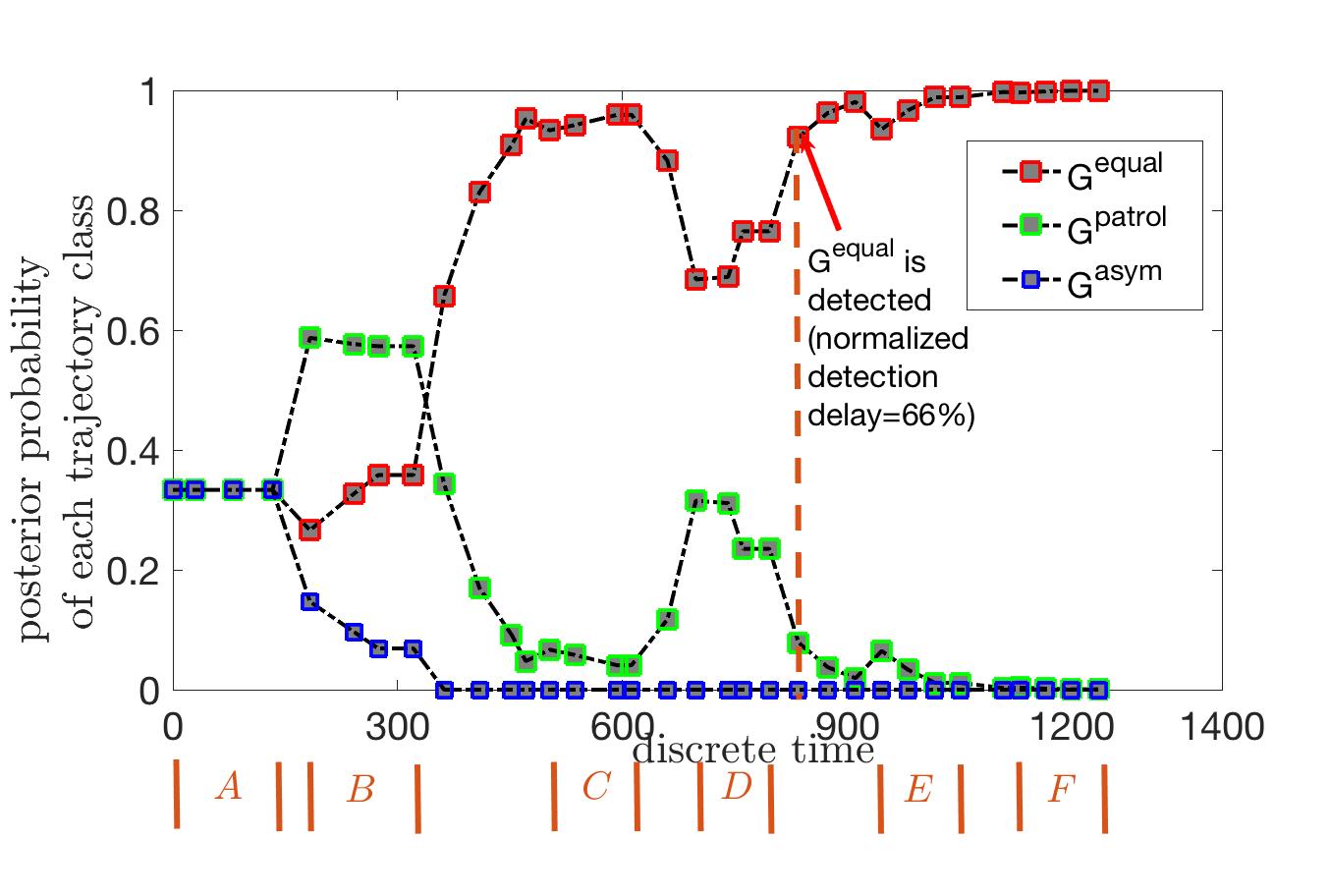}
\footnotesize{$A$:block(1,5) $B$:block(2,5) $C$:block(6,5) $D$:block(6,3) $E$:block(2,3) $F$:block(1,3)}
 \caption{}
\end{subfigure}
\begin{subfigure}{0.47\textwidth}
 \includegraphics[width=\textwidth]{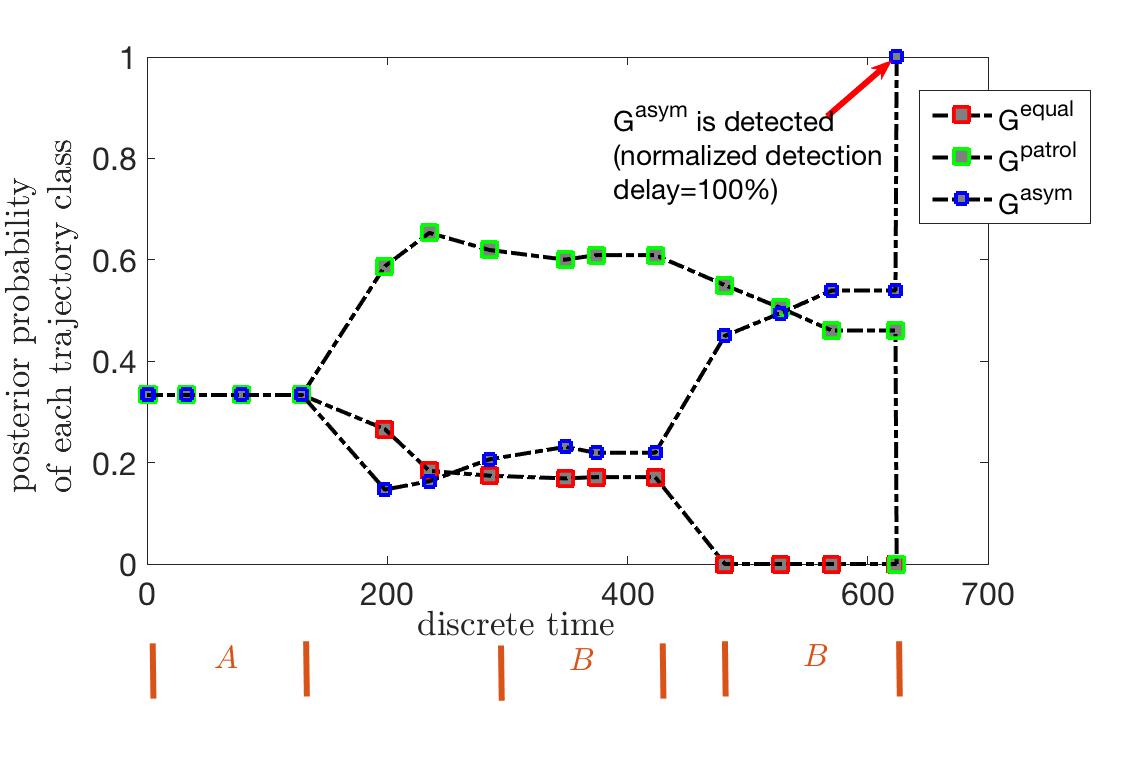}
 \footnotesize{A:block(3,3) B:block(1,1)}
 \caption{}
 
\end{subfigure}%
~
\begin{subfigure}{0.47\textwidth}
 \includegraphics[width=\textwidth]{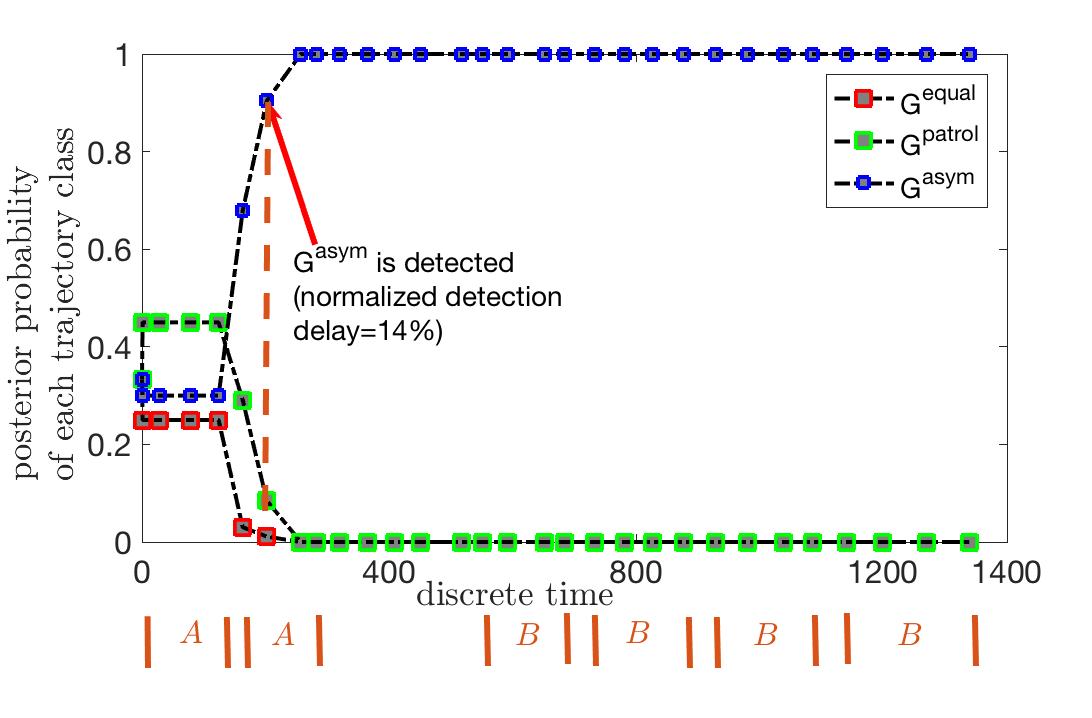}
 \footnotesize{$A$:block(3,6) $B$:block(6,3)}
 \caption{}

\end{subfigure}
\begin{subfigure}{0.47\textwidth}
 \includegraphics[width=\textwidth]{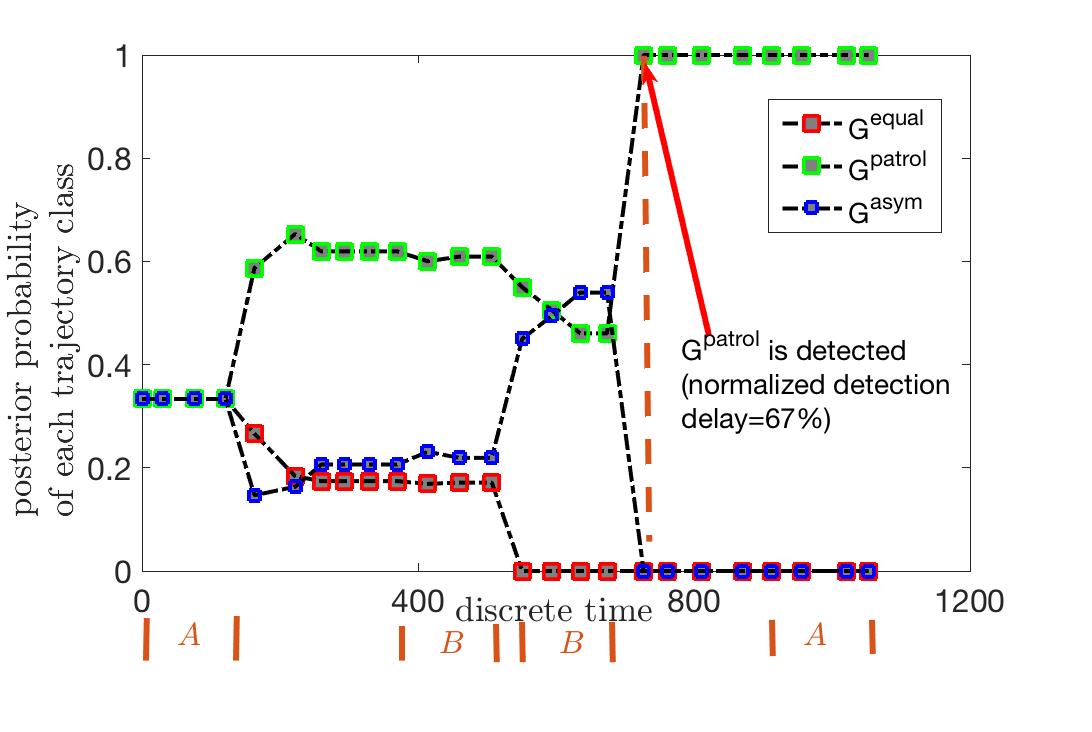}
 \footnotesize{$A$:block(3,5) $B$:block(1,1)}
 \caption{}

\end{subfigure}%
~
\begin{subfigure}{0.47\textwidth}
 \includegraphics[width=\textwidth]{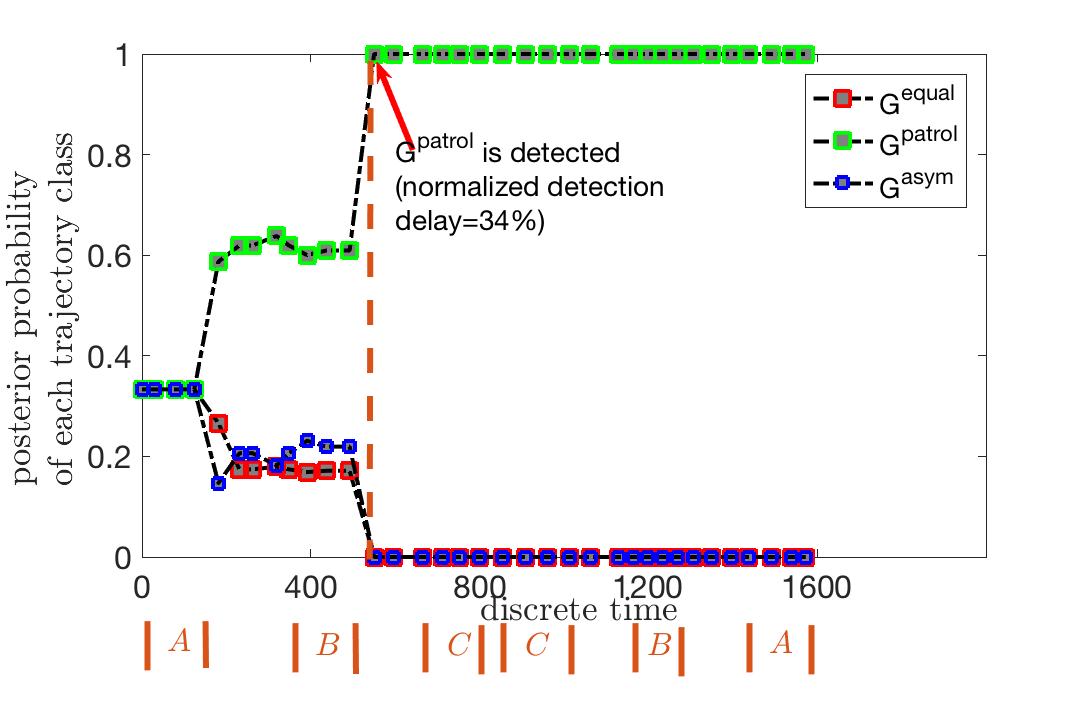}
 \footnotesize{$A$:block(2,2) $B$:block(3,6) $C$:block(3,3)}
 \caption{}
 
\end{subfigure}
\caption{(a) \textcolor{black}{Average normalized detection delay computed by averaging the normalized detection delay in \eqref{eq:normalized detection delay} over 20 simulation trials. $\sigma_{r_k}$, $\sigma_{\dot r_k}$ and $\sigma_{\theta_k}$ are defined in \eqref{eq:R}. $\sigma=[5,0.3,0.5]'$ and $\Delta=[1,0.1,0.1]'$}. (b)-(f) Examples of the posterior probability (vertical axis) of each trajectory   {class} computed by the CSCFG syntactic tracker when tracking a target   {with (b) $G^{\text{\rsone}}$ trajectory model, (c)(d) $G^{\text{\rsthree}}$ trajectory model and (e)(f) $G^{\text{\rstwo}}$ trajectory model}. ``$A$: block($i,j$)" indicates $A$ is the block located at the $i$th column, $j$th row in the roadmap; see Fig.\,{\ref{fig:simu,roadmap}}. ``$|A|$" in brown represents a search of block $A$ (a string of roads that surround block $A$); see Sec.\,{\ref{subsec:rsone}}. The discrete time interval is in units of 0.2 seconds.}
\label{fig:simu_detection}
\end{figure*}

{\em \textcolor{black}{Results}.} The average (over 20 simulation trials) normalized detection delay of $G^{\rstwo}$ is $43\%$; see Fig.\,{\ref{fig:simu_detection}}(a). 

1) $N=2$; normalized detection delay$\approx 70\%$

Fig.\,{\ref{fig:simu_detection}}(e) illustrates an example of $G^{\rstwo}$ detection where \textcolor{black}{two different blocks (blocks $A$, $B$) are patrolled and re-patrolled in the reverse order by the target. In Fig.\,{\ref{fig:simu_detection}}(e),   {the posterior} $p(G^{\rsone}|\mathbf{z}_{1:k})$ \textcolor{black}{becomes insignificantly small} during the target's 2nd search of block $B$. This is because \textcolor{black}{the trajectory class} $G^{\rsone}$ does not allow consecutive searches of the same block. Given the target's partial trajectory (which consists of 1 search of block $A$ followed by 2 searches of block $B$), the difference between   {the posterior probabilities} $p(G^{\rstwo}|\mathbf{z}_{1:k})$ and $p(G^{\rsthree}|\mathbf{z}_{1:k})$ is \textcolor{black}{small}. As a result, the radar operator \textcolor{black}{cannot} determine the target's trajectory model. Subsequently, the target \textcolor{black}{moves} from block $B$ to block $A$ and its trajectory is classified to be $G^{\rstwo}$.}

2) $N=3,4$; normalized detection delay$\approx 30\%$

In Fig.\,{\ref{fig:simu_detection}}(f), 3 different blocks (blocks $A$, $B$, $C$) are repetitively patrolled by the target. During the target's movement from block $B$ to $C$, \textcolor{black}{$p(G^{\rstwo}|\mathbf{z}_{1:k})$ \textcolor{black}{increases monotonically} whereas $p(G^{\rsone}|\mathbf{z}_{1:k})$ and $p(G^{\rsthree}|\mathbf{z}_{1:k})$ decline to negligible values}. The reason is that \textcolor{black}{the trajectory family} $G^{\rsone}$ only permits \textcolor{black}{search} in two block   {rows; while} blocks $A$, $B$, $C$ are located at three different rows; \textcolor{black}{see Fig.\,{\ref{fig:simu_detection}}(f)}.   {Also}, the target does not make \textcolor{black}{extra} searches of block   {$B$; therefore the posterior {$p(G^{\rsthree}|\mathbf{z}_{1:k})$ decreases to zero as $k$ increases}}.

\subsection{Syntactic Tracker Improves State Estimate~Accuracy}
\textcolor{black}{Fig.\,{\ref{fig:simu,state estimate improvement}} shows that for the trajectory classes of $G^{\rsone}$, $G^{\rsthree}$ and $G^{\rstwo}$}, the CSCFG syntactic tracker improves the state estimate accuracy (defined in \textcolor{black}{\eqref{eq:state estimate improvement}}) by $19\%$ compared with the classical VSIMM tracker. The reason is that, in the VSIMM tracker, trajectories from $q_k$ to $q_{k+n}$ consist of all possible trajectories on the roadmap. By comparison, in the CSCFG syntactic tracker, trajectories from $q_k$ to $q_{k+n}$ are constrained by the meta-level trajectory estimates which is a subset of those considered by the VSIMM tracker. To illustrate this, \textcolor{black}{consider Fig.\,{\ref{fig:simu_detection}}(d) where a target follows $G^{\rsthree}$ trajectory class}. In the CSCFG syntactic tracker, once $G^{\rsthree}$ is detected (based on its posterior probability exceeding 0.9),   {$q_k^i$ predicted by the meta-level tracker for each particle equals its right-handed road at an intersection}.   {As a result, particles tend to explore  right turns at intersections to search the second region}. By comparison, particles in the VSIMM tracker try all possible trajectories  and thus have larger state estimate errors. 

\begin{figure}
\centering
\includegraphics[width=0.34\textwidth]{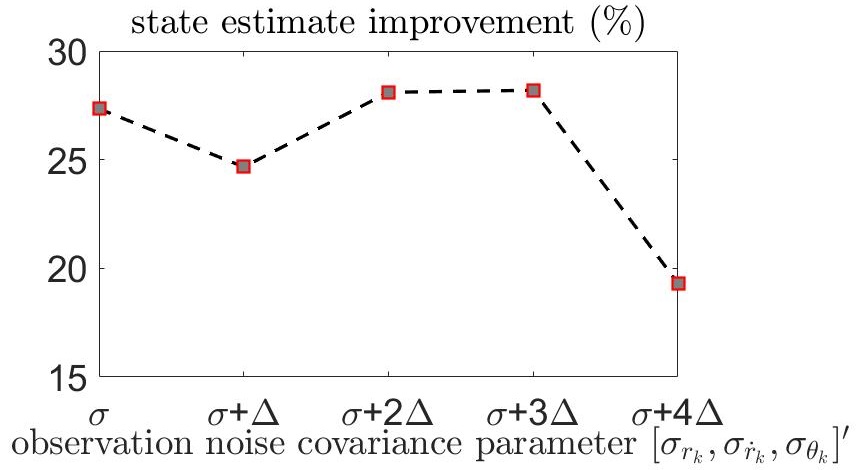}
\caption{State estimate improvement in \textcolor{black}{\eqref{eq:state estimate improvement}} of CSCFG syntactic tracker compared with the classical VSIMM tracker. \textcolor{black}{$\sigma_{r_k}, \sigma_{{\dot r}_k}$ and $\sigma_{\theta_k}$ are defined in \eqref{eq:R}. $\sigma=[5,0.3,0.5]'$ and $\Delta=[1,0.1,0.1]'$}.}
\label{fig:simu,state estimate improvement}
\end{figure}

\subsection{{Syntactic Tracker vs Template Matching}}
\label{subsec:simu,template matching}
Here  we illustrate via numerical examples the performance of  template matching for detecting anomalous trajectories. Let $M$, $N$ denote the number of block rows and columns of the roadmap; see Fig.\,{\ref{fig:simu,roadmap}}. The number of possible templates for each trajectory model ($G^{\rsone}$, $G^{\rstwo}$ and $G^{\rsthree}$) proposed in Sec.~\ref{sec:level2} grows exponentially with $M\times N$. By comparison, the number of production rules of a CSCFG only grows linearly with $M\times N$. In other words, CSCFG serves as a more efficient generative model for these  anomalous trajectories. 

We conducted 50 independent simulation trials. In each trial: (1) 5 templates for each anomalous trajectory model were stochastically generated. (2) Template matching was run on each of the simulated 60 anomalous targets in Sec.\,{\ref{subsec:simu,rsone}}, Sec.\,{\ref{subsec:simu,rsthree}} and Sec.\,{\ref{subsec:simu,rstwo}}. The target's trajectory is classified to the anomalous trajectory model
which has the template with the least edit distance. The edit distance is defined as the minimal number of replacements, insertions and deletions required to change from one string to another \cite{edit_distance}.

{\em Results.} The performance of template matching is measured by the successful classification rate (SCR):  number of successful classifications of anomalous targets divided by the number of anomalous targets. The average SCR for template matching over 50 independent trials is $45.20\%$. In comparison
using the CSCFG based syntactic tracker, the SCR is $90\%$.  The reason for the poor performance of template matching is that we only used 5  templates for each anomalous trajectory model; this cannot cover  all trajectories that belong to a specific trajectory class. (Recall from Sec.\ref{subsec:compare MC, SCFG and CSCFG} that we would need an exponential number of templates to cover all possible trajectories.)

\subsection{{Syntactic Tracker Performance with Missing Tracklets}}
\label{subsec:simu,missing tracklets}

Here we  illustrate the performance of the CSCFG syntactic tracker when  some   tracklets are missing due to missed radar observations. The performance is evaluated using  60 independent trials (20 trials for each of the anomalous trajectory models $G^{\rsone}$, $G^{\rstwo}$ and $G^{\rsthree}$) over a sweep of increasing percentage tracklets loss. 

First we consider randomly missing measurements from the entire measurement sequence. In this case the average normalized detection delay 
 of CSCFG syntactic tracker remains  virtually unchanged. The reason for this insensitivity to tracklet loss is that with high probability most of the missing tracklets are not at/around intersections - it is at these intersections where most uncertainty occurs.
 Therefore, we now focus on missing tracklets at road intersections.  We simulated  two types of radar observation processes with missing tracklets:

1) {\em Case 1.} {Missing tracklets appear at 10\% to  50\% intersections. 20 tracklets at and after  an intersection are uniformly chosen to be missing.}

2) {\em Case 2.} {Missing tracklets appear in bursts at $10\%$ to $50\%$ intersections. Intersection tracklets and 20 road tracklets after an intersection are chosen to be missing.}

\emph{Results.} 
The performance of CSCFG syntactic tracker  with missing tracklets is shown in Fig.\,{\ref{fig:simu,missingtracklets}}(a)(b). The performance metric  is  the increase in the average detection delay  defined in \eqref{eq:normalized detection delay} compared to the case without tracklet loss. For Case 1, increasing tracklets loss from 10\% to 50\% results in negligible  changes in the average detection delay; see Fig.\,{\ref{fig:simu,missingtracklets}}(a).  
For Case 2, see Fig.\,{\ref{fig:simu,missingtracklets}}(b), the average detection delay increases with increasing  tracklet loss. 
The reason for the difference 
between Case 1 and 2 is
as follows: Consider a sequence of tracklets  $a,a,a,a,b,b,b,b$. 
The detection delay is 
$\hat{t}- 4$ where 4 is the position of the first $b$ in the actual tracklet sequence and $\hat{t}$ is the position of the first $b$ in the missing tracklet sequence. In Case 1, it is very unlikely that consecutive
$b's$ starting from position 4 onwards are omitted, In comparison, Case 2 omits the first several $b$'s by construction (bursty missing tracklets).

\begin{figure*}
\centering
\begin{subfigure}{0.31\textwidth}
\includegraphics[width=\textwidth]{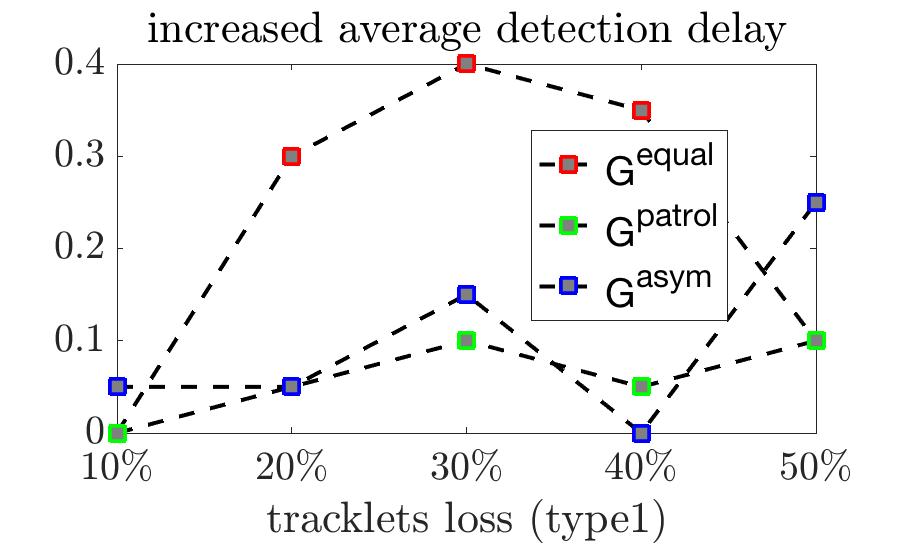}
\caption{}
\end{subfigure}%
~
\begin{subfigure}{0.3\textwidth}
\includegraphics[width=\textwidth]{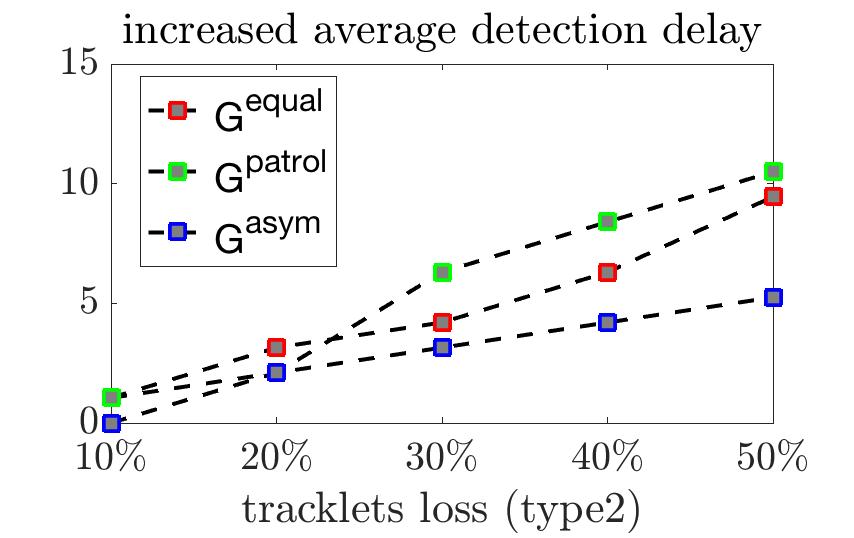}
\caption{}
\end{subfigure}%
~
\begin{subfigure}{0.3\textwidth}
\includegraphics[width=\textwidth]{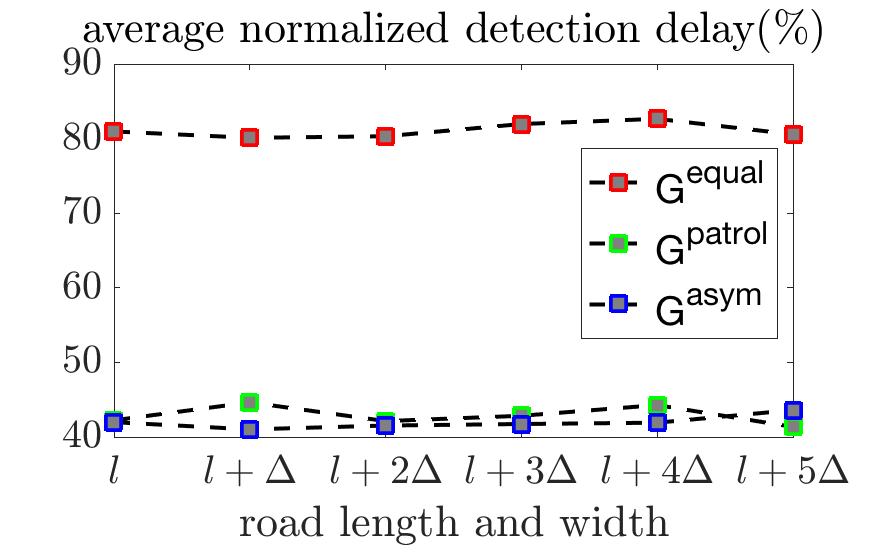}
\caption{}
\end{subfigure}
\caption{Performance of CSCFG syntactic tracker when (a) radar observations have Case 1 missing tracklets, (b) radar observations have Case 2 missing tracklets, (c) road length and width insensitivity. The increased average detection delay is computed by averaging the detection delay in \eqref{eq:normalized detection delay} over 20 independent simulation trials for each anomalous trajectory model: $G^{\rsone}$, $G^{\rstwo}$ and $G^{\rsthree}$. In (a)(b), the horizontal axis $10\%\sim 50\%$ represents the percentage of tracklets loss; see Sec.\,{\ref{subsec:simu,missing tracklets}}. {In (c), $l$=[100, 5] denote, respectively, the road length and width (in meters) and $\Delta$=[10, 1].}}
\label{fig:simu,missingtracklets}
\end{figure*}

\subsection{Syntactic Tracker with Perturbed Anomalous Trajectory}
\begin{figure}
\centering
\begin{tikzpicture}
\tikzstyle{vertex1}=[circle,fill=black!25,minimum size=10pt,inner sep=0pt]
\tikzstyle{vertex2}=[rectangle,fill=black!25,minimum size=9pt,inner sep=0pt]
\tikzset{edge/.style = {->,> = latex'}}
\node[vertex1] (c0) at  (0,0) { };
\node (d1) at (1,0){$\ldots$};
\node[vertex1] (c1) at  (2,0) { };
\node[vertex1] (c2) at  (3,0) { };
\node[vertex2] (s1) at  (3,0.8) { };
\node(d3) at  (4,0) {$\ldots$};
\node(d2) at  (4,0.8) {$\ldots$};
\node[vertex1] (c3) at  (5,0) { };
\node[vertex2] (s2) at  (5,0.8) { };
\node[vertex1] (c4) at  (6,0) { };
\node (d4) at  (7,0) {$\ldots$};
\node[vertex1] (c5) at  (8,0) { };

\draw [thick, ->,blue] (c0) edge (d1);
\draw [thick, ->,blue] (d1) edge (c1);
\draw [thick, ->,blue,dashed] (c1) edge (c2);
\draw [thick, ->,blue,dashed] (c2) edge (d3);
\draw [thick, ->,blue,dashed] (d3) edge (c3);
\draw [thick, ->,blue,dashed] (c3) edge (c4);
\draw [thick, ->,blue] (c4) edge (d4);
\draw [thick, ->,blue] (d4) edge (c5);
\draw [thick, ->,red] (c1) edge (s1);
\draw [thick, ->,red] (s1) edge (d2);
\draw [thick, ->,red] (d2) edge (s2);
\draw [thick, ->,red] (s2) edge (c4);
\end{tikzpicture}
\caption{Perturbed  anomalous trajectory. Circles and squares indicate alphabets (road names). The circle sequence connected by the blue arrows represents an anomalous trajectory that belongs to $G\in\{G^{\rsone},G^{\rstwo},G^{\rsthree}\}$. One segment (circles connected by the dashed blue arrows) of the circle sequence is replaced with a square string which makes the entire trajectory (circles and squares connected by solid arrows) slightly different from the original anomalous trajectory class $G$. Note also that the number of squares and the replaced circles may not be equal; i.e., the perturbed trajectory can have a different length.}
\label{fig:slightly different example}
\end{figure}
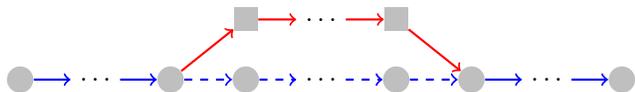

Here we evaluate the  performance of the syntactic tracker when the target's anomalous trajectory is a small perturbation   from the specified families  $G\in\{G^{\rsone},G^{\rstwo},G^{\rsthree}\}$; see Fig.\,{\ref{fig:slightly different example}}. We conducted simulations on 60 independent trials (20 trials for each anomalous trajectory model: $G^{\rsone}$, $G^{\rstwo}$, $G^{\rsthree}$). In each trial: {\em Step 1.} One unique directed sequence $\unique(q_{1:N})\in G$ was generated. Recall, $\unique(q_{1:N})$ is a sequence of alphabets (road names). {\em Step 2.} Randomly select one segment (3$\sim$ 5 alphabets) in $\unique(q_{1:N}$) and replace  this segment with a different trajectory. Note that this new inserted segment is constrained to  make the resulting  trajectory physically  feasible, e.g., jump from a road to a non adjacent road is not allowed. {\em Step 3.} The CSCFG meta-level parser was run to parse $\unique(q_{1:N})$ until the parser stopped; denote this stopping time as $t_{\text{forward}}$. {\em Step 4.} The reverse CSCFG meta-level parser\footnote{A reverse CSCFG meta-level tracker models the reverse string of the anomalous trajectory.}was run to parse the reverse string of $\unique(q_{1:N})$ until the parser stopped, denote this stopping time as $t_{\text{reverse}}$. The target is classified to the trajectory model that has maximal $t_{\text{forward}}+t_{\text{reverse}}$\footnote{The reason is that we seek to find the maximum possible trajectory length that is consistent with the model.}. We also ran template matching for comparison. 

{\em Results.}  The performance is evaluated in terms of the  SCR  defined in Sec.\,{\ref{subsec:simu,template matching}}.
SCRs for CSCFG syntactic tracker and template matching are 85.00\%, 45.07\%, respectively. This shows that the proposed CSCFG syntactic tracker makes significant improvement in classifying anomalous trajectories with small perturbations compared with template matching.

\subsection{Insensitivity to  Road Length and Width}
Fig.\,{\ref{fig:simu,missingtracklets}(c) shows that increasing road length and width results in negligible changes in the average normalized detection delay. The average normalized detection delay for $G^{\rsone}$, $G^{\rstwo}$, $G^{\rsthree}$ remains at $81\%$, $43\%$, $42\%$; see Fig.\,{\ref{fig:simu,missingtracklets}}(c). 

\section{Conclusion}
We constructed a 3-level model for the roadmap based syntactic tracking problem using a natural language model called \textcolor{black}{the} constrained stochastic context free grammar (CSCFG). At the highest level of abstraction, the roadmap was modeled as a weighted, directed graph; at the second level, trajectories were modeled via a CSCFG; finally the base level kinematics of the target (physical sensor level)  were modeled by a VSIMM. 

The key idea is that the CSCFG is a generative model for a variety of complex anomalous trajectories.
 The CSCFG model substantially generalizes earlier work 
 \cite{SCFG_intent_inference,SCFG_syntactic_tracker,SCFG_metalevel_modeling,SCFG_metalevel_estimate}
 in syntactic tracking and allows us to incorporate  realistic continuity constraints 
imposed by roadmaps.
We also presented a CSCFG-driven particle filtering algorithm to compute the posterior probability of each trajectory model -- the algorithm
combines the functionalities of IMM and the Earley Stolcke parser (from natural language processing).
In  numerical examples,  the CSCFG syntactic tracker enables anomalous trajectory detection and improves the state estimate accuracy by up to 19\% compared to the classical VSIMM tracker.


\paragraph*{Acknowledgement} We \textcolor{black}{acknowledge} Dr.\ Muralidhar Rangaswamy of Air Force Research Laboratory and Dr.\ Martie  Goulding of MacDonald Dettwiler Associates for  disucssions.

\bibliographystyle{ieeetr}
\bibliography{ref}

\begin{thebibliography}{10}

\bibitem{baselevel2}
M.~Ulmke and W.~Koch, ``Road-map assisted ground moving target tracking,'' {\em
  IEEE Transactions on Aerospace and Electronic Systems}, vol.~42,
  pp.~1264--1274, October 2006.

\bibitem{baselevel3}
D.~Streller, ``Road map assisted ground target tracking,'' in {\em 2008 11th
  International Conference on Information Fusion}, pp.~1--7, June 2008.

\bibitem{baselevel4}
Y.~Cheng and T.~Singh, ``Efficient particle filtering for road-constrained
  target tracking,'' {\em IEEE Transactions on Aerospace and Electronic
  Systems}, vol.~43, pp.~1454--1469, October 2007.

\bibitem{VSIMM1}
T.~Kirubarajan, Y.~Bar-Shalom, K.~R. Pattipati, and I.~Kadar, ``Ground target
  tracking with variable structure {IMM} estimator,'' {\em IEEE Transactions on
  Aerospace and Electronic Systems}, vol.~36, pp.~26--46, Jan 2000.

\bibitem{VSIMM2}
H.~A.~P. Blom and Y.~Bar-Shalom, ``The interacting multiple model algorithm for
  systems with markovian switching coefficients,'' {\em IEEE Transactions on
  Automatic Control}, vol.~33, pp.~780--783, Aug 1988.

\bibitem{jump_Markov}
Y.~Li, D.~Jin, Z.~Wang, P.~Hui, L.~Zeng, and S.~Chen, ``A markov jump process
  model for urban vehicular mobility: Modeling and applications,'' {\em IEEE
  Transactions on Mobile Computing}, vol.~13, pp.~1911--1926, Sept 2014.

\bibitem{ahmad2016bayesian}
B.~I. Ahmad, J.~K. Murphy, P.~M. Langdon, and S.~J. Godsill, ``Bayesian intent
  prediction in object tracking using bridging distributions,'' {\em IEEE
  transactions on cybernetics}, 2016.

\bibitem{anomalous_1}
R.~Fraile and S.~Maybank, ``Vehicle trajectory approximation and
  classification,'' in {\em British Machine Vision Conference}, 1998.

\bibitem{anomalous_2}
X.~Li, J.~Han, S.~Kim, and H.~Gonzalez, ``Roam: Rule- and motif-based anomaly
  detection in massive moving object data sets*,'' (Philadelphia),
  pp.~273--284, Society for Industrial and Applied Mathematics, 2007.

\bibitem{anomalous_3}
S.~Srivastava, K.~K. Ng, and E.~J. Delp, ``Co-ordinate mapping and analysis of
  vehicle trajectory for anomaly detection,'' in {\em 2011 IEEE International
  Conference on Multimedia and Expo}, pp.~1--6, July 2011.

\bibitem{edit_distance}
H.~Oh, S.~Kim, H.-S. Shin, A.~Tsourdos, and B.~A. White, ``Behaviour
  recognition of ground vehicle using airborne monitoring of unmanned aerial
  vehicles,'' {\em International Journal of Systems Science}, vol.~45, no.~12,
  pp.~2499--2514, 2014.

\bibitem{masterthesis}
S.~Gao, ``Roadmap enhanced improvement to the {VSIMM} tracker via a
  {Constrained Stochastic Context Free Grammar},'' Master's thesis, University
  of British Columbia, 2017.

\bibitem{constrained_grammar}
K.~E. Mark, M.~I. Miller, and U.~Grenander, ``Constrained stochastic language
  models,'' in {\em Image Models (and Their Speech Model cousins)},
  pp.~131--140, Springer, 1996.

\bibitem{constrained_grammar_parameter_estimation}
K.~Mark, M.~Miller, U.~Grenander, and S.~Abney, ``Parameter estimation for
  constrained context-free language models,'' in {\em Proceedings of the
  Workshop on Speech and Natural Language}, HLT '91, (Stroudsburg, PA, USA),
  pp.~146--149, Association for Computational Linguistics, 1992.

\bibitem{SCFG_intent_inference}
A.~Wang, V.~Krishnamurthy, and B.~Balaji, ``Intent inference and syntactic
  tracking with {GMTI} measurements,'' {\em IEEE Transactions on Aerospace and
  Electronic Systems}, vol.~47, pp.~2824--2843, Oct 2011.

\bibitem{SCFG_syntactic_tracker}
M.~Fanaswala and V.~Krishnamurthy, ``Syntactic models for trajectory
  constrained track-before-detect,'' {\em IEEE Transactions on Signal
  Processing}, vol.~62, pp.~6130--6142, Dec 2014.

\bibitem{SCFG_metalevel_modeling}
M.~Fanaswala and V.~Krishnamurthy, ``Spatiotemporal trajectory models for
  metalevel target tracking,'' {\em IEEE Aerospace and Electronic Systems
  Magazine}, vol.~30, pp.~16--31, Jan 2015.

\bibitem{SCFG_metalevel_estimate}
M.~Fanaswala and V.~Krishnamurthy, ``Detection of anomalous trajectory patterns
  in target tracking via stochastic context-free grammars and reciprocal
  process models,'' {\em IEEE Journal of Selected Topics in Signal Processing},
  vol.~7, pp.~76--90, Feb 2013.

\bibitem{balajia2012consistency}
B.~Balaji, ``Consistency of stochastic context-free grammars and application to
  stochastic parsing of {GMTI} tracker data,'' in {\em SPIE Defense, Security,
  and Sensing}, pp.~83920S--83920S, International Society for Optics and
  Photonics, 2012.

\bibitem{blasch2012high}
E.~Blasch, {\'E}.~Boss{\'e}, and D.~A. Lambert, {\em High-level information
  fusion management and systems design}.
\newblock Artech House, 2012.

\bibitem{pumping_lemma}
A.~V. Aho and J.~D. Ullman, {\em The theory of parsing, translation and
  compiling}.
\newblock Englewood Cliffs, NJ,USA: Prentice-Hall, 1972.

\bibitem{earley_stolcke_parser}
A.~Stolcke, ``An efficient probabilistic context-free parsing algorithm that
  computes prefix probabilities,'' {\em Computational linguistics}, vol.~21,
  no.~2, pp.~165--201, 1995.

\bibitem{wolfe2004attributes}
J.~M. Wolfe and T.~S. Horowitz, ``Opinion: What attributes guide the deployment
  of visual attention and how do they do it?.,'' {\em Nature Reviews
  Neuroscience}, vol.~5, no.~6, pp.~495 -- 501, 2004.

\bibitem{asymmetric1}
A.~Borji and L.~Itti, ``State-of-the-art in visual attention modeling,'' {\em
  IEEE Transactions on Pattern Analysis and Machine Intelligence}, vol.~35,
  pp.~185--207, Jan 2013.

\bibitem{asymmetric2}
M.~M. Cheng, N.~J. Mitra, X.~Huang, P.~H.~S. Torr, and S.~M. Hu, ``Global
  contrast based salient region detection,'' {\em IEEE Transactions on Pattern
  Analysis and Machine Intelligence}, vol.~37, pp.~569--582, March 2015.

\bibitem{levin1996classifying}
D.~T. Levin, ``Classifying faces by race: The structure of face categories.,''
  {\em Journal of Experimental Psychology: Learning, Memory, and Cognition},
  vol.~22, no.~6, pp.~1364 -- 1382, 1996.

\bibitem{djuric2010}
P.~M. Djuri{\'c} and M.~F. Bugallo, ``Adaptive systems of particle filters,''
  in {\em Forty Fourth Asilomar Conference on Signals, Systems and Computers
  (ASILOMAR), 2010}, pp.~59--63, IEEE, 2010.

\bibitem{intro_to_SCFG}
J.~E. Hopcroft, R.~Motwani, and J.~D. Ullman, {\em Introduction to automata
  theory, languages, and computation}.
\newblock Boston, MA;Toronto, ON;: Pearson/Addison Wesley, 3rd~ed., 2007.

\bibitem{consistency}
R.~Gecse and A.~Kovács, ``Consistency of stochastic context-free grammars,''
  {\em Mathematical and Computer Modelling}, vol.~52, no.~3, pp.~490 -- 500,
  2010.

\end{thebibliography}

\appendices
\section{\textcolor{black}{What is syntactic tracking?}}
\label{appendix:background}
\textcolor{black}{Since readers in target tracking} may not be familiar with natural language models, we present a short description of stochastic context free grammars here; see \cite{intro_to_SCFG}\cite{SCFG_metalevel_modeling} for details.

The aim of syntactic tracking is to classify a target's trajectory using  natural language processing based {\em generative models}. We view the spatial-temporal trajectory of a target as a sequence (\textcolor{black}{string}) of noisy symbols (\textcolor{black}{alphabets}). 

Both SCFG and CSCFG are \textcolor{black}{5-tuples} of the form $\{\mathcal{N},\mathcal{T},S,\mathcal{R},\mathcal{P}\}$. Here, $\mathcal{N}$ is a finite set of nonterminals and $\mathcal{T}$ is a finite set of terminals. $S\in\mathcal{N}$ denotes the starting symbol. $\mathcal{R}$ is a finite set of production rules and $\mathcal{P}$ denotes a probability function over production rules in $\mathcal{R}$. In a SCFG, the production rule is of the form
\begin{equation}
\begin{split}
&\textcolor{black}{X}\rightarrow \lambda, \textcolor{black}{X}\in\mathcal{N}\textcolor{black}{\text{ with probability } p}\label{eq:SCFG rule}\\
&\lambda: \textcolor{black}{\text{a string of nonterminals and terminals}}
\end{split}
\end{equation}
The production rule in \eqref{eq:SCFG rule} says we can replace the nonterminal \textcolor{black}{$X$} with the string $\lambda$ with probability $p$. For example, in Fig.\,{\ref{fig:SCFG example}}(a), the production rule $A\rightarrow a[0.8]$ represents we replace the nonterminal $A$ with terminal $a$ with probability 0.8. \textcolor{black}{The derivation of a SCFG is illustrated via a parse tree (see Fig.\,{\ref{fig:SCFG example}}(b)) and its output is a string of terminals. The probability of a parse tree is the product of probabilities of production rules applied to it; see Fig.\,{\ref{fig:SCFG example}}(b)}. 
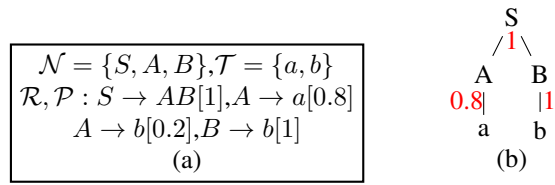
\begin{figure}
\begin{tikzpicture}
\centering
\begin{scope}
\node[draw=black,rectangle,thick,align=center,fill=white] (a) at (0,0)  
{
$\mathcal{N}=\{S,A,B\}$,$\mathcal{T}=\{a,b\}$\\
$\mathcal{R},\mathcal{P}:S\rightarrow AB[1]$,$A\rightarrow a[0.8]$\\
$A\rightarrow b[0.2]$,$B\rightarrow b[1]$\\
(a)};
\end{scope}
\begin{scope}[xshift=1.7in,yshift=0.5in,scale=0.5]
\node{S}
child{node{A}
      child{node{a}}}
child{node{B}
      child{node{b}}};
\node at (0,-3.8){(b)};
\node at (0,-0.6){\color{red}{1}};
\node at (-1.2,-2.2){\color{red}{0.8}};
\node at (1,-2.2){\color{red}{1}};
\end{scope}
\end{tikzpicture}
\caption{(a) Example of a SCFG. Numbers in brackets denote probabilities of  production rules. (b) Example of a SCFG parse tree. 
Numbers in red denote probabilities of  production rules. The probability of this parse tree is $1\times 1\times 0.8=0.8$. }
\label{fig:SCFG example}
\end{figure}
\section{A Parsing Algorithm for the constrained stochastic context free grammar (CSCFG)}
\label{appendix:parsing}
Here, we present a Bayesian parsing algorithm (with polynomial time computation cost) for the CSCFG=$\{\mathcal{N},\mathcal{T},S,\mathcal{R},\mathcal{P}\}$ defined in Sec.\,{\ref{subsec:level2}}. The aim of this parsing algorithm is to compute the one step \textcolor{black}{prediction} $p(q_{k+1} |\hat q_{1:k}, \text{CSCFG})$ and the prefix probability $p(\hat q_{1:k}|\text{CSCFG})$ for the CSCFG. $q_k\in\mathcal{T}$ denotes the clean terminal at time $k$ and \textcolor{black}{$\hat q_{1:k}=(\hat q_1,\hat q_2,\ldots,\hat q_k)$ denotes a string of noisy symbols (alphabets)}. These two probabilities are used in the particle \textcolor{black}{filtering} algorithm in Sec.\,{\ref{sec:algorithm}}. The parsing algorithm for CSCFGs is a modified version of the classic Earley Stolcke parser, and to the best of our knowledge, new.

The classic Earley Stolcke parser is used to  compute the one step prediction probability and the prefix probability for the stochastic context free grammar (SCFG)\cite{earley_stolcke_parser}. The Earley state is defined as
\begin{equation}
\label{eq:scfg earley state}
\leftidx{^{k}_{k'}}X \rightarrow \lambda.\beta\mu[\alpha,\gamma]
\end{equation}
where $X\in\mathcal{N}$ and $\lambda$, $\beta$, $\mu$ \textcolor{black}{denote strings of nonterminals and terminals}. $k$ is the current epoch and $k'$ is the back pointer to the epoch when this Earley state is generated by the prediction operation. The dot '.' marks that the portion on its left handed side \textcolor{black}{that} has been parsed or recognized by the parser. $\alpha$ and $\gamma$ are called the forward probability and the inner probability, respectively \cite{earley_stolcke_parser}. We expand the Earley state so that it can record its associated terminals 
\begin{equation}
\label{eq:CSCFG earley state}
\leftidx{^{k}_{k'}}X \rightarrow \lambda.\beta\mu[s,f][\alpha,\gamma]
\end{equation}
\textcolor{black}{In \eqref{eq:CSCFG earley state}}, $k$, $k'$, $X$, $\lambda$, $\beta$, $\mu$, $\alpha$, $\gamma$ are defined in $\eqref{eq:scfg earley state}$. $s$ is called the \emph{start symbol} and denotes the previous terminal when $X$ is rewritten by $\lambda\beta\mu$ in epoch~$k'$. $f$ is called the \emph{finish symbol} and denotes the final terminal that has been parsed before the dot.
\begin{algorithm}
\scriptsize
 \begin{algorithmic}      
\STATE{$X$, $Y$, $\Gamma\in\mathcal{N}$, $\lambda$, $\mu$, $\beta$, $\eta\in(\mathcal{N}\cup\mathcal{T})^{*}$, $\eta\notin\mathcal{N}$, $a\in\mathcal{T}$. $n$ is a general denotation for an Earley state and $u_k$ is the set of all Earley states at epoch~k. $\hat{q}_{k}$ is the hard or soft estimate at time~$k$.} 
\STATE{1. Scanning}
\FOR{$\leftidx{^{k-1}_{k'}}X \rightarrow \lambda.a\mu [s,f][\alpha,\gamma] \in u_{k-1}$}
\STATE{Add $\leftidx{^{k}_{k'}}X \rightarrow \lambda a.\mu [s',f'][\alpha',\gamma']$ to $u_k$ if $p(\hat{q_k}|a)>0$\\$\alpha'=\alpha p(\hat{q_k}|a)$\\$\gamma'=\gamma p(\hat{q_k}|a)$\\$s'=s$\\$f'=a$}\ENDFOR\\
$\zeta_k=\underset{n\in u_k}\sum \alpha(\leftidx{^{k}_{k'}}X \rightarrow \lambda a.\mu) $\\
$\forall n\in u_k$, normalize $\alpha,\gamma$ using $\zeta_k$\\
\STATE{2. Completion}
\FOR{$\leftidx{^{k}_{k'}}\Gamma\rightarrow \eta.[s,f][\alpha,\gamma]\in u_k$}\STATE{\FOR{$\leftidx{^{k'}_{k''}}X \rightarrow \lambda.Y\mu [s'',s][\alpha'',\gamma''] \in u_{k'}$}\STATE{\IF{$R_u(Y,\Gamma|s)\neq 0$}\STATE{Add $\leftidx{^{k}_{k''}}X \rightarrow \lambda Y.\mu [s',f'][\alpha',\gamma']$ to $u_k$\\$\alpha'+=\alpha''\gamma R_u(Y,\Gamma|s)$\\$\gamma'+=\gamma''\gamma R_u(Y,\Gamma|s)$\\$s'=s''$\\$f'=f$}\ENDIF}\ENDFOR}\ENDFOR\\
\STATE{3. Prediction}
\FOR{$\leftidx{^{k}_{k'}}X \rightarrow \lambda.Y\mu [s,f][\alpha,\gamma] \in u_{k}$}\STATE{Add $\leftidx{^{k}_{k}}\Gamma \rightarrow .\beta[s',f'][\alpha',\gamma']$ to $u_k$ if $R_l(Y,\Gamma|f)\neq 0 $ \\$\alpha'+=\alpha R_{l}(Y,\Gamma|f)p(\Gamma\rightarrow \beta|f)$\\$\gamma'=p(\Gamma\rightarrow \beta|f)$\\$s'=f$\\$f'=f$}\ENDFOR
\end{algorithmic}
\caption{\scriptsize Earley Stolcke Parsing Algorithm for  CSCFG).   {Compared with the Earley Stolcke Parser for the SCFG, Algorithm~\ref{algorithm:parsing} accounts for the serial constraints in scanning, completion and prediction steps}.}
\label{algorithm:parsing}   
\end{algorithm}
The probabilistic left corner relation is computed as $p(X\overset{L}\rightarrow Y|a)=\underset{X\rightarrow Y\lambda|a\in\mathcal{R}}\sum p(X\rightarrow Y\lambda|a)$ where $X$, $Y\in\mathcal{N}$, $a\in\mathcal{T}$ \textcolor{black}{and $\lambda$ is a string of nonterminals and terminals}. The probabilistic unit relation is computed as $p(X\overset{U}\rightarrow Y|a)= p(X\rightarrow Y|a)$. After computing the left corner and unit relations between nonterminals, we can get the reflexive, transitive left corner matrix $R_{l}$ and the unit production relation matrix $R_{u}$.   {See \cite{earley_stolcke_parser} for details}. 

  {The parsing algorithm} for the CSCFGs is shown in Algorithm~\ref{algorithm:parsing}. The one step \textcolor{black}{prediction} probability is computed as
\begin{equation}
\label{eq:one step predictor}
p(q_{k}|\hat q_{1:k-1},\text{CSCFG})=\frac{\underset{a=q_{k},n\in u_{k-1}}\sum\alpha(\leftidx{^{k-1}_{k'}}X\rightarrow \lambda.a\mu [s,f])}{\underset{n\in u_{k-1}}\sum\alpha(\leftidx{^{k-1}_{k'}}X\rightarrow \lambda.a\mu [s,f])}
\end{equation}
$n$ is a general denotation for the Earley state and $u_{k-1}$ is the set of all Earley states at epoch~$k-1$. The prefix probability 
is computed as
\begin{equation}
\label{eq:prefix}
p(\hat q_{1:k}|\text{CSCFG})=\prod_{t=1}^{k} \zeta_t
\end{equation}

\section{CSCFGs for meta-level modeling \textcolor{black}{and consistency}}
\label{appendix:rule and consistency}
Fig.\,{\ref{fig:production rule}} presents the CSCFG production rules that constitute generative models for the 3 anomalous trajectory classes discussed in Sec.\,{\ref{sec:level2}}, namely,  $G^{\text{\rsone}}$, $G^{\text{\rstwo}}$ and $G^{\text{\rsthree}}$. 

Next, \textcolor{black}{we show that the generative models for these trajectory classes are well posed. i.e.,  the CSCFG models generate finite length strings (trajectories).   In natural} language processing, such models are said to be consistent; mathematically, the 
 Galton Watson branching process is sub-critical.

We  now  {prove that} the production rules for $G^{\text{\rsthree}}$ in Fig.\,{\ref{fig:production rule}}   {are consistent}.   {The proofs of consistency of $G^{\rsone}$ and $G^{\rstwo}$ are similar and omitted}.   {For $G^{\rsthree}$}, the stochastic mean matrix\textcolor{black}{\cite{consistency}} is given by \eqref{eq:stochastic mean matrix} \textcolor{black}{where} $(X,B)$ denotes the expected number of variables $B$ resulting from rewriting $X$. 
It can be verified that 
the absolute value of the largest eigenvalue of the matrix \eqref{eq:stochastic mean matrix} is $\max((X,X),(D,D),(NE,NE), (NW,NW),(SE,SE),\\(SW,SW))$ and each diagonal element is less than one in magnitude. For example, given any terminal $e_{ij}$, $(X,X)$ equals the probability of the production rule $X\rightarrow BXB|e_{ij}$ which is less than one. Hence the absolute value of the largest eigenvalue of \eqref{eq:stochastic mean matrix} is less than one and therefore the CSCFG for modeling $G^{\text{\rsthree}}$ is consistent. \textcolor{black}{Proofs} of the consistency of CSCFGs for modeling $G^{\text{\rsone}}$ and $G^{\text{\rstwo}}$ \textcolor{black}{are} similar. 
\begin{equation}
\scriptsize
\label{eq:stochastic mean matrix}
\left[
\begin{smallmatrix}
0&1&2&1&0&0&0&0&0&0&0\\
0&(X,X)&(X,B)&0&0&0&0&(X,NE)&(X,NW)&(X,SE)&(X,SW)\\
0&0&0&0&1&0&0&0&0&0&0\\
0&0&1&(D,D)&0&0&0&0&0&0&0\\
0&0&0&0&0&1&0&0&0&0&0\\
0&0&0&0&0&0&1&0&0&0&0\\
0&0&0&0&0&0&0&0&0&0&0\\
0&0&0&0&0&0&0&(NE,NE)&0&0&0\\
0&0&0&0&0&0&0&0&(NW,NW)&0&0\\
0&0&0&0&0&0&0&0&0&(SE,SE)&0\\
0&0&0&0&0&0&0&0&0&0&(SW,SW)
\end{smallmatrix}\right]
\end{equation}
\begin{figure}
\centering
\fboxsep=0pt
\fbox{\scriptsize
\begin{minipage}[t][6.15cm]{8.7cm}
\begin{equation}
\nonumber
\begin{split}
&(1) S\rightarrow AXC, X\rightarrow AXC, X\rightarrow EN, X\rightarrow ES, X\rightarrow WN, X\rightarrow WS\\
&(2)e_{ij}\not\xrightarrow{\text{clockwise}}e_{jk}: A\rightarrow e_{jk}A_2\quad C\rightarrow e_{jk}C_2\\
&(3)e_{ij}\xrightarrow{\text{clockwise}}e_{jk}:
A_2(C_2)\rightarrow e_{jk}A_3(C_3), A_3(C_3)\rightarrow e_{jk}A_4(C_4)\\
&A_4\rightarrow e_{jk}E,A_4\rightarrow e_{jk},C_4\rightarrow e_{jk}W,C_4\rightarrow e_{jk}\\
&(4) \theta(e_{jk})\in\{\text{north,east}\}:TNE\rightarrow e_{jk}TNE\quad TNE\rightarrow e_{jk}\\
&e_{ij}\xrightarrow{\text{clockwise}}e_{jk}: NE\rightarrow e_{jk}NE\quad e_{ij}\not\xrightarrow{\text{clockwise}}e_{jk}: NE\rightarrow e_{jk}\\
&(5) \theta(e_{jk})\in\{\text{north,west}\}:TNW\rightarrow e_{jk}TNW\quad TNW\rightarrow e_{jk}\\
&e_{ij}\xrightarrow{\text{clockwise}}e_{jk}: NW\rightarrow e_{jk}NW\quad e_{ij}\not\xrightarrow{\text{clockwise}}e_{jk}: NW\rightarrow e_{jk}\\
&(6) \theta(e_{jk})\in\{\text{south,east}\}:TSE\rightarrow e_{jk}TSE\quad TSE\rightarrow e_{jk}\\
&e_{ij}\xrightarrow{\text{clockwise}}e_{jk}: SE\rightarrow e_{jk}SE\quad e_{ij}\not\xrightarrow{\text{clockwise}}e_{jk}: SE\rightarrow e_{jk}\\
&(7) \theta(e_{jk})\in\{\text{south,west}\}:TSW\rightarrow e_{jk}TSW\quad TSW\rightarrow e_{jk}\\
&e_{ij}\xrightarrow{\text{clockwise}}e_{jk}: SW\rightarrow e_{jk}SW\quad
e_{ij}\not\xrightarrow{\text{clockwise}}e_{jk}: SW\rightarrow e_{jk}\\
&(8)\theta(e_{jk})=\text{east}:E\rightarrow e_{jk}E,E\rightarrow e_{jk}\\
&(9)\theta(e_{jk})=\text{west}:W\rightarrow e_{jk}W,W\rightarrow e_{jk} \quad\quad\quad\quad\quad\quad\quad\quad\quad\quad G^{\rsone}
\end{split}
\end{equation}
\end{minipage}
}
\fbox{\scriptsize
\begin{minipage}[t][3.4cm]{8.7cm}
\begin{equation}
\nonumber
\begin{split}
&(1) e_{ij}\notin\operatorname{search\,block}(m,n),\exists e_{jk}\in\operatorname{search\,block}(m,n):\\
&
S\rightarrow B_{mn}XB_{mn},X\rightarrow B_{mn}XB_{mn},X\rightarrow B_{mn}B_{mn},B_{mn}\rightarrow e_{jk}W_2\\
&(2) e_{ij}\xrightarrow{\text{clockwise}}e_{jk}:
W_2\rightarrow e_{jk}W_3,W_3\rightarrow e_{jk}W_4,W_4\rightarrow e_{jk}\\
&W_4\rightarrow e_{jk}NE,W_4\rightarrow e_{jk}NW,W_4\rightarrow e_{jk}SE,W_4\rightarrow e_{jk}SW\\
& (3) \theta(e_{jk})\in\{\text{north, east}\}:
NE\rightarrow e_{jk}NE,NE\rightarrow e_{jk}\\
& (4) \theta(e_{jk})\in\{\text{north, west}\}:
NW\rightarrow e_{jk}NS,NW\rightarrow e_{jk}\\
& (5) \theta(e_{jk})\in\{\text{south, east}\}:
SE\rightarrow e_{jk}SE,SE\rightarrow e_{jk}\\
& (6) \theta(e_{jk})\in\{\text{south, west}\}:
SW\rightarrow e_{jk}SW,SW\rightarrow e_{jk} \quad\quad\quad\quad G^{\rstwo}
\end{split}
\end{equation}
\end{minipage}
}
\fboxsep=0pt
\fbox{\scriptsize
\begin{minipage}[t][2.9cm]{8.7cm}
\begin{equation}
\nonumber
\begin{split}
&(1)S\rightarrow BXBD,X\rightarrow BXB,X\rightarrow NE,X\rightarrow NW\\
&X\rightarrow SE,X\rightarrow SW,D\rightarrow BD,D\rightarrow B\\
&(2)e_{ij}\xrightarrow{\text{clockwise}}e_{jk}:
B\rightarrow e_{jk}W_2,W_2\rightarrow e_{jk}W_3,W_3\rightarrow e_{jk}W_4,W_4\rightarrow e_{jk}\\
&(3)\theta(e_{jk})\in\{\text{north, east}\}:
NE\rightarrow e_{jk}NE,NE\rightarrow e_{jk}\\
&(4)\theta(e_{jk})\in\{\text{north, west}\}:NW\rightarrow e_{jk}NW,NW\rightarrow e_{jk}\\
&(5)\theta(e_{jk})\in\{\text{south, east}\}:
SE\rightarrow e_{jk}SE,SE\rightarrow e_{jk}\\
&(6)\theta(e_{jk})\in\{\text{south, west}\}:
SW\rightarrow e_{jk}SW,SW\rightarrow e_{jk}\quad\quad\quad\quad\quad G^{\rsthree}
\end{split}
\end{equation}
\end{minipage}
}
\caption{\scriptsize CSCFG generative models for $G^{\text{\rsone}}$, $G^{\text{\rstwo}}$ and $G^{\text{\rsthree}}$. $e_{ij}\in E$ denotes the previous terminal. $\theta(.)$ is defined in \eqref{eq:graph}. $\operatorname{search\,block}(m,n)$ is described in Sec.\,{\ref{subsec:rsone}}. $e_{ij}\xrightarrow{\text{clockwise}}e_{jk}$ represents $e_{jk}$ can be reached from $e_{ij}$ by moving in the clockwise direction, e.g., $e_{2,7}\xrightarrow{\text{clockwise}}e_{7,8}$ in Fig.\,{\ref{fig:roadmap}}.}
\label{fig:production rule}
\end{figure}

\begin{IEEEbiography}[{\includegraphics[width=1in,height=1.25in,clip,keepaspectratio]{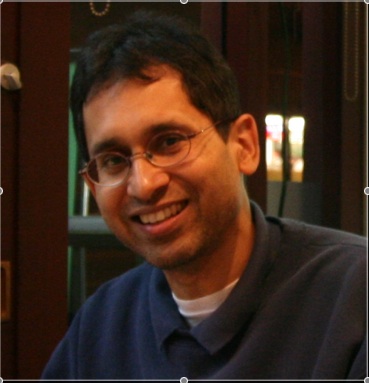}}]{Vikram Krishnamurthy}
(F'05) received the Ph.D. degree from the Australian
National University
in 1992. He is currently a professor at Cornell Tech and the
School of Electrical \& Computer Engineering,
Cornell University. From 2002-2016 he
was a Professor and Canada Research Chair
at the University of British Columbia, Canada.
His research interests include statistical signal
processing  and
stochastic control in social networks and adaptive sensing. He served
as Distinguished Lecturer for the IEEE Signal Processing Society and
Editor-in-Chief of the IEEE Journal on Selected Topics in Signal Processing.
In 2013, he was awarded an Honorary Doctorate from KTH
(Royal Institute of Technology), Sweden. He is author of the books
{\em Partially Observed Markov Decision Processes} and
{\em Dynamics of Engineered Artificial Membranes and Biosensors} published by Cambridge
University Press in 2016 and 2018, respectively.
\end{IEEEbiography}

\begin{IEEEbiography}[{\includegraphics[width=1in,height=1.25in,clip,keepaspectratio]{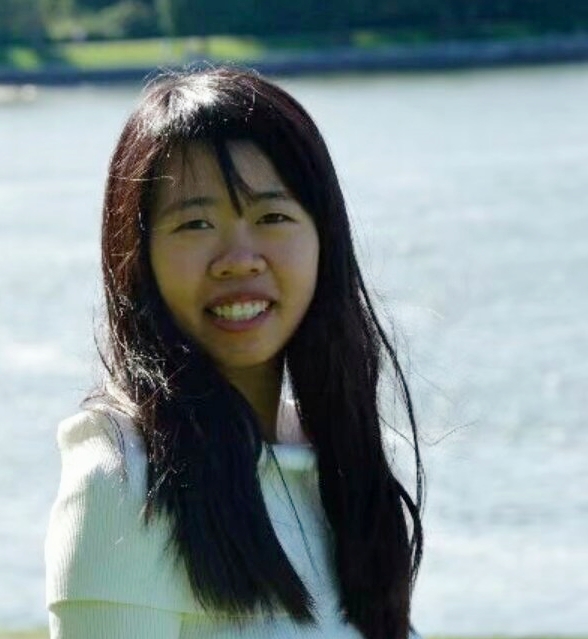}}]
{Sijia Gao} is currently pursuing a Ph.D. degree in the Department of Electrical and Computer Engineering, Cornell Tech, Cornell University. She received MAsc degree from University of British Columbia, Canada in 2017 and B.S. degree from Huazhong University of Science and Technology, China in 2014, both from Department of Electrical and Computer Engineering. Her current research interests include target tracking and applying signal processing to neuroscience.
\end{IEEEbiography}

\end{document}